\documentclass[10pt,journal,compsoc]{IEEEtran}
\ifCLASSOPTIONcompsoc
  \usepackage[nocompress]{cite}
\else
  \usepackage{cite}
\fi

\usepackage{amsmath,amsthm}
\usepackage{mathtools}
\usepackage{amssymb}
\usepackage{dutchcal}
\usepackage[ruled,linesnumbered]{algorithm2e}
\usepackage{eqparbox}
\usepackage{makecell}
\usepackage{graphicx}
\usepackage{subfigure}
\usepackage{xcolor}
\usepackage{balance}
\usepackage{booktabs}
\usepackage{multirow}
\usepackage{array}
\usepackage{mdwmath}
\usepackage{mdwtab}
\usepackage{eqparbox}

\usepackage{stfloats}

\ifCLASSOPTIONcaptionsoff
  \usepackage[nomarkers]{endfloat}
 \let\MYoriglatexcaption\caption
 \renewcommand{\caption}[2][\relax]{\MYoriglatexcaption[#2]{#2}}
\fi
\usepackage{url}
\begin{document}
%
\title{DACFL: Dynamic Average Consensus Based Federated Learning in Decentralized Topology}
\author{Zhikun~Chen,
	    Daofeng~Li,
        Jinkang~Zhu,~\IEEEmembership{Life Member,~IEEE},
        Sihai~Zhang,~\IEEEmembership{Senior Member,~IEEE}\\
\IEEEcompsocitemizethanks{
	\IEEEcompsocthanksitem Zhikun Chen and Daofeng Li are with the Department of Electronic Engineering and Information Science, School of Information Science and Technology, University of
	Science and Technology of China. Address: No. 96 Jinzhai Road, Hefei, Anhui Province, 230026, P. R. China
	.\hfil\break
	E-mail: \{zhikunch,df007\}@mail.ustc.edu.cn
	\IEEEcompsocthanksitem Jinkang Zhu is with the PCNSS laboratory, School of Information Science and Technology, University of
	Science and Technology of China. Address: No. 96 Jinzhai Road, Hefei, Anhui Province, 230026, P. R. China.
	\hfil\break
	E-mail: jkzhu@ustc.edu.cn
	\IEEEcompsocthanksitem Sihai Zhang is with the Key Laboratory of Wireless-Optical Communications, Chinese Academy of Sciences, School of Information Science and Technology, University of
Science and Technology of China. Address: No. 96 Jinzhai Road, Hefei, Anhui Province, 230026, P. R. China
	.\hfil\break
	*E-mail: shzhang@ustc.edu.cn
}
       }

\IEEEtitleabstractindextext{%
\begin{abstract}
Federated learning (FL) is a burgeoning distributed machine learning framework where a central parameter server (PS) coordinates many local users to train a globally consistent model.
Conventional federated learning inevitably relies on a centralized topology with a PS. 
As a result, it will paralyze once the PS fails. 
To alleviate such a single point failure, especially on the PS, some existing work has provided decentralized FL (DFL) implementations like CDSGD and D-PSGD to facilitate FL in a decentralized topology. 
However, there are still some problems with these methods, e.g., significant divergence between users' final models in CDSGD and a network-wide model average necessity in D-PSGD.
In order to solve these deficiency, this paper devises a new DFL implementation coined as DACFL, where each user trains its model using its own training data and exchanges the intermediate models with its neighbors through a symmetric and doubly stochastic matrix.
The DACFL treats the progress of each user's local training as a discrete-time process and employs a first order dynamic average consensus (FODAC) method to track the \textit{average model} in the absence of the PS.
In this paper, we also provide a theoretical convergence analysis of DACFL on the premise of i.i.d data to strengthen its rationality.
The experimental results on MNIST, Fashion-MNIST and CIFAR-10 validate the feasibility of our solution in both time-invariant and time-varying network topologies, and declare that DACFL outperforms D-PSGD and CDSGD in most cases.
\end{abstract}

}

\maketitle

\IEEEdisplaynontitleabstractindextext

\IEEEpeerreviewmaketitle

\ifCLASSOPTIONcompsoc
\IEEEraisesectionheading{\section{Introduction}\label{sec1}}
\else
\section{Introduction}
\fi

\IEEEPARstart{N}{umerously} increasing smart mobile devices such as phones, tablets, etc. have provided convenience for human beings and generated plentiful user-related data like images, sound, text and others.
In order to make full use of such huge amount of data, data mining and machine learning techniques \cite{wu2013data, michie1994machine,pedregosa2011scikit} have been developed.
However, these techniques usually work on a central parameter server (PS).
This makes it inevitable to collect large amounts of users' data into PS, which may lead to an excessive overhead and expose users' privacy like food preference, consuming behavior, social ties and so on.
According to the general data protection regulation (GDPR) \cite{voigt2017eu}, both industry and academia should pay more attention to the privacy protection of machine learning.
Hence, taking into consideration the privacy concerns when exploiting users' data is now becoming an imperative in the era of artificial intelligence.
To tackle this problem, Google has advocated an alternative known as federated learning (FL) \cite{konevcny2016federated,9134988,yang2019federated}. Rather than collecting users' raw data into a PS, the FL enables each user to train their respective models using their own data, and only the intermediate models are periodically synchronized by the PS to obtain a global model.

\begin{figure}[ht]
	\centering
	\includegraphics[width=0.5\textwidth]{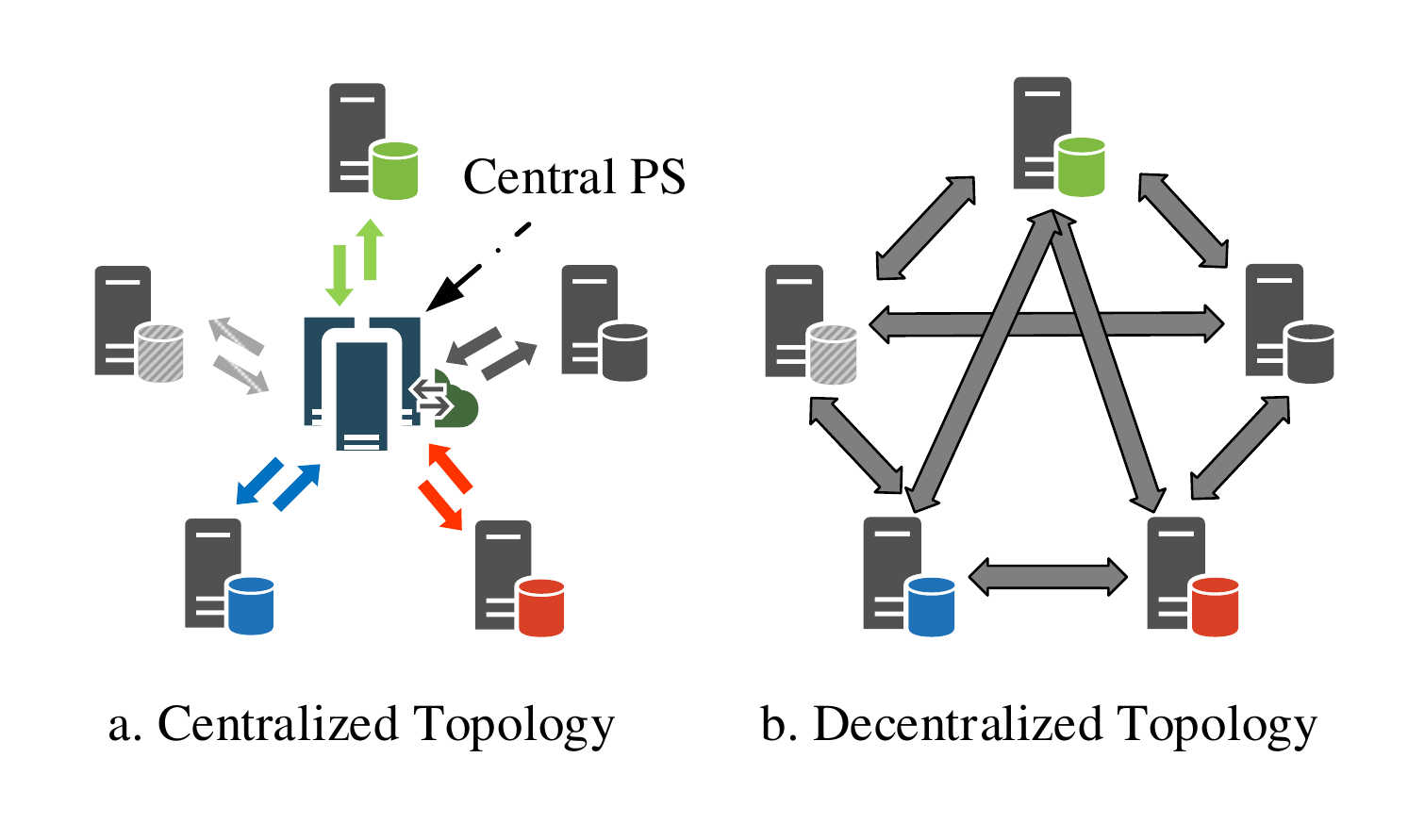}
	\caption{Centralized topology vs Decentralized topology. }
	\label{topology}
\end{figure}

However, there exist several limitations in conventional FL due to the centralized network topology (see fig. \ref{topology}). 
For example, to iteratively synchronize users' local-trained models and send back the aggregated global model to them, the communication load of the entire system is extremely unbalanced. Specifically, because all users need to communicate with the PS concurrently iteratively, the communication traffic jam is very likely to happen to the PS and the performance will be degraded when the network bandwidth is low.
What's worse, if there is a single point failure that happens to the PS by any chance, e.g., being attacked, the entire framework will be paralyzed.

To eliminate the above bottleneck caused by centralized network topology, a natural idea is that, is it possible to facilitate FL in a decentralized topology (see fig. \ref{topology}) without the support of a PS? 
Thankfully, some existing work on wireless sensor network (WSN) and device-to-device (D2D) communication \cite{hakola2010device} has confirmed the possibility of communication in a decentralized network topology.
Therefore, we believe that it is not only important but also applicable to implement FL in a decentralized topology.

In this paper, we consider the problem of facilitating federated learning in a decentralized network topology in the absence of a central parameter server. To this end, we propose a new decentralized federated learning framework based on dynamic average consensus, which is coined as DACFL. 
In DACFL, there is no central PS and all users are connected through an undirected graph, which is denoted by a symmetric doubly stochastic matrix (i.e., mixing matrix).
Each user trains its local model utilizing its own training data. 
The respective trained model intermediates of different users are treated as different discrete-time reference input sequences.
In order to aggregate different users' intermediate models\footnote{In a centralized federated learning, there is central PS to periodically aggregate different users' model into a global model and send it back to all users. However, to implement federated learning in a decentralized network topology where no PS exists, the global aggregation phase would be a critical problem.}, for each user, we employ an approximate estimation method, i.e., a first-order dynamic average consensus (FODAC), to track the \textit{average model} estimation.
Besides, to alleviate the possible local over-fitting problem, especially for non-i.i.d data, each user in DACFL takes its neighborhood weighted average model parameter to reinitialize its local model after each training round.

Actually, there has been some existing work to date devoting to implementing federated learning in a fully decentralized network topology (as is described in \ref{subsec: dfl}). 
Apart from them, references \cite{jiang2017collaborative} and \cite{lian2017can} should be the most similar work to this paper.
In \cite{jiang2017collaborative}, the authors propose a new consensus-based distributed SGD (CDSGD) algorithm for collaborative deep learning over a fixed network topology that enables data parallelization as well as decentralized computation.
In \cite{lian2017can}, the authors study a decentralized parallel SGD (D-PSGD) algorithm on a decentralized computational network. They further prove that D-PSGD achieves the same convergence rate as the centralized parallel SGD (C-PSGD) algorithm, but outperforms C-PSGD by avoiding the communication jam.
Although some existing work summarizes CDSGD and D-PSGD into the same method \cite{wang2019cooperative,haddadpour2019convergence}, note that we distinguish between CDSGD and D-PSGD by whether the algorithm requires a global average or not in this paper (D-PSGD additionally needs a network-wide model average compared with CDSGD). 
In what follows, we describe the deficiency of CDSGD and D-PSGD, and differentiate our solution from these two methods in detail.

\begin{figure*}[t]
	\centering
	\includegraphics[width=0.7\textwidth]{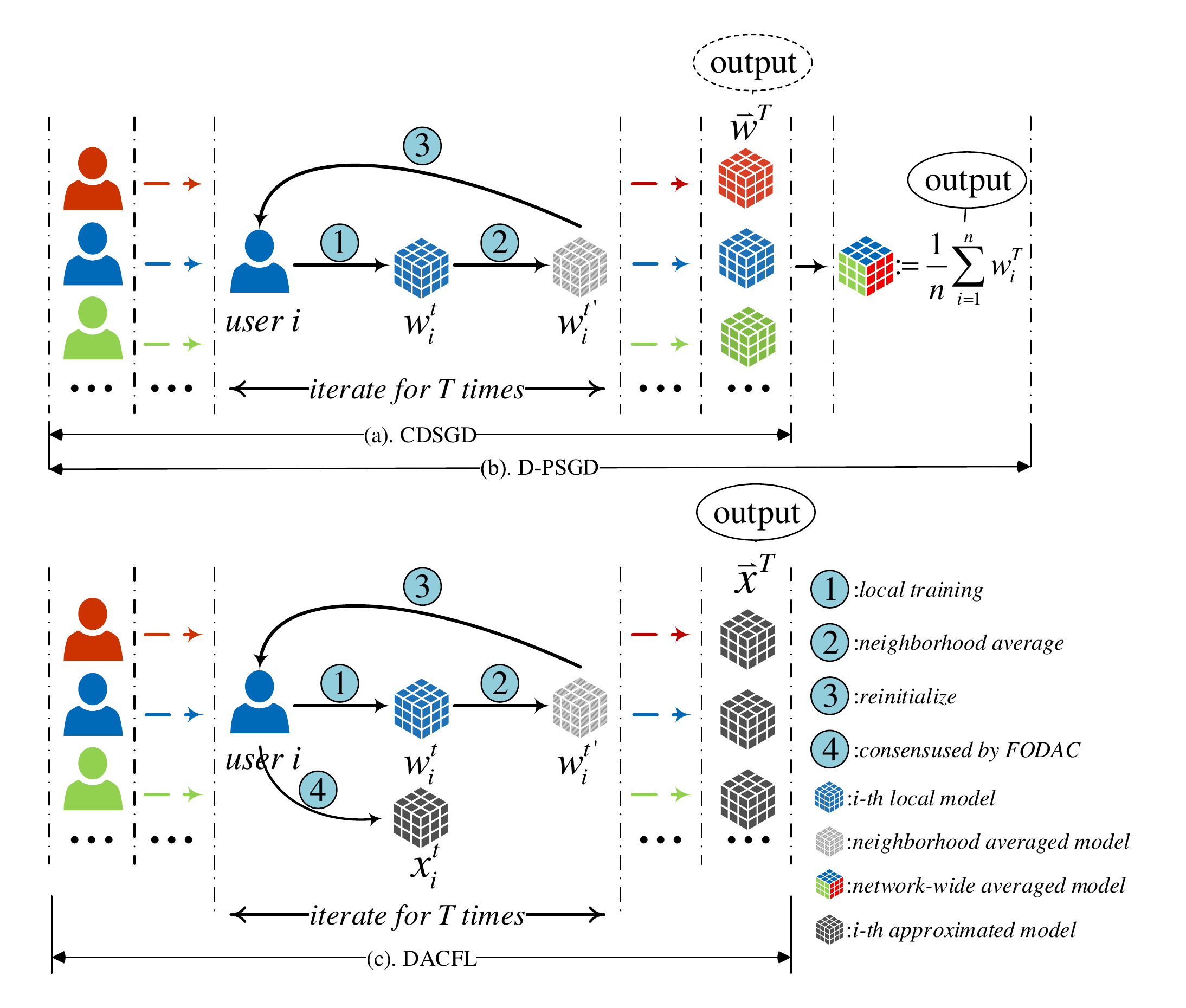}
	\caption{Comparison between CDSGD, D-PSGD and DACFL (our solution). We differ CDSGD and D-PSGD that D-PSGD additionally needs a network-wide model average before the output. Our solution differentiates CDSGD and D-PSGD by employing FODAC to approximate the ``averaged model'' over all users. An iteration of ``1$\to$2$\to$3'' for CDSGD and D-PSGD or an iteration of ``1$\to$2$\to$3$\to$4'' is called a training round (i.e., a communication round) in this paper. For more details, please refer to \textbf{Algorithm} \textbf{\ref{cdsgd}}, \textbf{\ref{dpsgd}} and \textbf{\ref{dacfl}}.}
	\label{comparison}
\end{figure*}

For CDSGD in \cite{jiang2017collaborative}, there are two main limitations. 
One is that it only considers a fixed network topology with a uniform interaction matrix (identical value of each element in the matrix). The other is that it assumes that the training data is independent and identically distributed (i.i.d) over all users. 
These two limitations make it not well suited for federated learning since mutative network topology and non-i.i.d data are common occurrences in federated learning. 
Besides, we empirically verify that when faced with a sparse topology, there is obvious variance across all users' final models. 
This is also inconsistent with the goal of FL to attain a globally consistent model for all users.
While for D-PSGD which additionally performs a network-wide model average in \cite{lian2017can}, since there is no PS in decentralized network topology, a problem that naturally arises is who should perform this network-wide model average task? And how to ensure the consistency of each user's model? Just imagine, if we make each user perform this network-wide average (although this is achievable by exchanging information through multi-hop communication), it will inevitably cause communication congestion in some users (especially those user nodes with higher degrees on the graph). 
Besides, the authors in \cite{lian2017can} consider only a fixed ring network topology.
Our solution differentiates CDSGD and D-PSGD from the following aspects. 
First, rather than taking roughly the neighborhood weighted average to replace the model aggregation like CDSGD or D-PSGD does, the DACFL employs an average consensus method, i.e., FODAC, to approximate the \textit{average model} for each user. 
Thanks to the effectiveness of FODAC, each user is able to track the \textit{average model} as the training progresses. Thus, no additional network-wide model average is needed in our DACFL.
Second, we extend more typical network topologies in this work. Apart from time-invariant (or static) and dense topology, we also consider time-varying (or dynamic) and sparse topology in this paper. 
Third, non-i.i.d data assignment in addition to i.i.d case is also considered in this paper. 
To alleviate the possible local over-fitting problem, each user in DACFL takes its neighborhood weighted average model parameter to reinitialize its local model after each training round.
Fig .\ref{comparison} briefly compares our DACFL with CDSGD and D-PSGD. 

The contributions of this paper can be summarized as follows:

\begin{itemize}
	\item This paper devises a new decentralized federated learning implementation coined as DACFL, which is more adaptable to non-ideal network topology compared with another two existing methods including CDSGD and D-PSGD w.r.t the average accuracy and the variance of accuracy. Unlike CDSGD and D-PSGD  taking roughly a neighborhood weighted average to approximate the global aggregated model, DACFL treats each user's local training as a discrete-time process and employs the FODAC to estimate the \textit{average model}, through which the users are able to obtain a near-average model in the absence of PS during the training process.

	\item We theoretically analyze the convergence of our proposed DACFL approach on the premise of some relatively ideal assumptions. The numeric result offers a convergence guarantee of our solution and reveals a positive correlation between the convergence speed and the learning rate and a negative correlation to the topology size. Specific experimental results also support our analysis.
	
	\item The DACFL is also empirically validated through a wide range of experiments including experiments on both i.i.d and non-i.i.d data, experiments on both time-invariant topology and time-varying topology with dense connectivity and sparse connectivity. Results of these experiments show that our DACFL outperforms CDSGD and D-PSGD w.r.t \textit{Average of Acc} and \textit{Var of Acc} in most cases.	
\end{itemize}

The remainder of this paper is organized as follows. 
First, some existing work about decentralized federated learning implementation and dynamic average consensus is summarized in section \ref{sec2} following a brief introduction to conventional federated learning. 
Then in section \ref{sec3}, we present the system model comprising node model and communication model, and mathematically formulate the decentralized federated learning problem as a combination of distributed machine learning and dynamic average consensus problem. 
In section \ref{sec4}, we first provide a heuristic method to construct a symmetric doubly stochastic matrix and introduce the FODAC algorithm, which is further employed to design our DACFL approach.
In section \ref{sec5}, the convergence of DACFL is theoretically analyzed on the premise of some relatively ideal assumptions. 
Then section \ref{sec6} presents the experiments and evaluates the performance of DACFL compared with CDSGD and D-PSGD. 
Finally, section \ref{sec7} concludes this paper.

\section{Related Works}
\label{sec2}

In this section, we first provide a brief introduction to federated learning, and then summarize some related work about decentralized federated learning implementations and about dynamic average consensus.

\subsection{Federated Learning}
From the technical perspective, there are two main strategies implementing federated learning: Horizontal Federated Learning (HFL) and Vertical Federated Learning\cite{yang2019federated}. 
In HFL, participating clients share the same set of features but target different populations. While in VFL, the client devices share the same population but target different sets of features. Note that throughout this paper, we focus only on the HFL. For detail introduction about VFL, please refer to \cite{hardy2017private,nock2018entity}.

The concept of federated learning was first introduced\cite{mcmahan2017communication}, where a distributed training model is executed by a number of participants, usually called clients or users, that share local model updates with a central parameter server whose role is to aggregate these updates to build a global model. 
Generally, a federated learning scenario consists of two main phases, i.e., \textit{local update} and \textit{global aggregation}. In the \textit{local update} phase, clients compute the gradients to minimize the underlying loss function using their local data. While in the \textit{global aggregation} phase, the central parameter server collects the model updates from different clients, aggregates these  model updates to form a global model and then sends back the aggregated result to the clients for their next training epoch.

Formally, suppose there are a subset of clients $C \subseteq N$ selected by the PS at training epoch $t \leq T$. Each client $c \in C$ keeps a local training dataset $D_c=\left\{X_c, Y_c\right\}$, where $X_c \in \mathbb{R}^{\left|D_c\right| \times d}$ represents the feature space of client $c$'s training data and $Y_c \in \mathbb{R}^{\left|D_c\right| \times m}$ is the associated label space of $D_c$.
Let $\ell\left(\omega;x_i,y_i\right)$ denote the loss function of data sample $x_i$, where $\omega$ denotes the parameters of neural networks, then the local loss function of client $c$ over training dataset $D_c$ can be expressed as
\begin{equation}
	\label{eq1}
	f_c\left(\omega \right)=\frac{1}{\left|D_c\right|}\sum_{i\in D_c}\ell\left(\omega;x_i,y_i\right)
\end{equation}
While for the global loss function over all clients $C$ and the whole training dataset, it can be given as
\begin{equation}
	\label{eq2}
	f\left(\omega\right)=\sum_{c=1}^{\left|C\right|}\frac{\left|D_c\right|}{\left|D\right|}f_c\left(\omega\right)
\end{equation}
where $D=\bigcup_c D_c$ represents for the whole training dataset over clients subset $C$ and $\left|D\right|=\sum_{c=1}^{\left|C\right|} \left|D_c\right|$ denotes the total number of the data samples.

To solve the above distributed optimization problem, existing work has offered several suggestions. A necessarily incomplete list of these solutions includes \textit{FedAvg} \cite{mcmahan2017communication}, \textit{FedProx} \cite{MLSYS2020_38af8613}, \textit{FedPAQ} \cite{reisizadeh2020fedpaq}, \textit{Turbo-Aggregate} \cite{so2021turbo}, \textit{FedMA} \cite{wang2020federated}, \textit{Semi-FL} \cite{chen2020semi} and \textit{Hier-FL} \cite{liu2020client}.
However, all these approaches rely on a centralized network topology where a central parameter server is necessary to execute the \textit{global aggregation} phase in federated learning \footnote{Note that in this section, we only aim at clarifying some definitions and preliminaries of federated learning. For a more comprehensive study of federated learning, please refer to \cite{yang2019federated, wahab2021federated,zhang2021survey}.}.

\subsection{Decentralized Federated Learning Implementation}
\label{subsec: dfl}
Actually, apart from references \cite{jiang2017collaborative} and \cite{lian2017can} which have been introduced in section \ref{sec1}, there has also been some existing work to date devoting to deep learning with decentralized computation without the aid of a PS.
In \cite{lalitha2018fully}, the authors present a distributed learning algorithm to address the fully decentralized federated learning problem. 
In this framework, users take a Bayesian-like approach to iterate and aggregate the beliefs of their one-hop neighbors and collaboratively estimate the global optimal parameter.
In \cite{roy2019braintorrent}, the authors propose a server-less, peer-to-peer approach for federated learning termed \textit{BrainTorrent}, particularly targeted towards medical applications. 
In this approach, all clients are assumed pair-wise connected and update their models by checking the local model version with the latest model version over the network.
In \cite{hu2019decentralized}, the authors propose a decentralized federated learning design, Combo, based on the gossip protocol. They also present a model segmentation level synchronization mechanism in order to maximize the utilization of bandwidth capacities between users. However, their design also requires a fully connected network topology and a prerequisite of randomly distributed data among the workers.
While in \cite{jiang2020bacombo}, the authors further extend the Combo into a bandwidth aware solution BACombo by greedily choosing the bandwidth-sufficient worker to reduce the transmission delay.
In \cite{hegedHus2019gossip}, the authors design an experimental study to compare federated learning with gossip learning, and find that gossip learning is comparable to federated learning in their result.
In \cite{li2020blockchain}, the authors propose a decentralized federated learning framework based on blockchain termed BFLC. The framework uses blockchain for the global model storage and the local model update exchange without a central server.
In \cite{pappas2021ipls}, the authors introduce a fully decentralized federated learning framework, termed IPLS, that is partially based on the interplanetary file system (IPFS). By using IPLS and connecting into the corresponding private IPFS network, any party can initiate the training process of a model or join an ongoing training process that has already been started by another party.
In \cite{li2021decentralized}, the authors propose the decentralized federated learning via mutual knowledge transfer (Def-KT) algorithm where local clients fuse models by transferring their knowledge to each other.

\begin{algorithm}
	\caption{CDSGD (baseline 1) \cite{jiang2017collaborative}}
	\label{cdsgd}
	\KwIn{Maximum epoch $m$; learning rate $\alpha$; number of agents $N$;}
	\KwOut{The trained models of $N$ users: $x_{m+1}^1,x_{m+1}^2,\cdots,x_{m+1}^N$;}
	\textbf{Initialize}: $x_0^j,\left(j=1,2,\cdots,N\right)$\;
	Distribute the training dataset to $N$ agents\;
	\For{each agent}{
		\For{$k=0:m$}{
			Randomly shuffle the corresponding data subset $\mathcal{D}_j$\;
			$\omega_{k+1}^j=\sum_{l \in Nb_{(j)}}\pi_{jl}x_k^l$\; \tcc{$Nb_{(j)}$ indicates the neighborhood of agent $j$; $\Pi=\left[\pi_{jl}\right]\in \mathbb{R}^{N\times N}$ indicates the connection weights between agents;} 
			$x_{k+1}^j=\omega_{k+1}^j-\alpha g_j(x_k^j)$\;
			\tcc{$g_j(x_k^j)$ is stochastic gradient of loss at $x_k^j$;}
		}	
	}
\end{algorithm}
\begin{algorithm}
	\caption{D-PSGD on the $i$th node (baseline 2) \cite{lian2017can}}
	\label{dpsgd}
	\KwIn{initial point $x_{0,i}=x_0$, step length $\gamma$, weight matrix $W$, and number of iterations $K$;}
	\For{$k=0,1,2,\cdots,K-1$}{
		Sample $\xi_{k,i}$ from local data on $i$-th node\;
		Compute a local stochastic gradient based on $\xi_{k,i}$ and current optimization variable $x_{k,i}$: $\nabla F_i\left(x_{k,i};\xi_{k,i}\right)$\;
		Compute the neighborhood weighted average by fetching optimization variables from neighbors: $x_{k+\frac{1}{2},i}=\sum_{j=1}^{n}W_{ij}x_{k,j}$\;
		Update the local optimization variable: $x_{k+1, i} \leftarrow x_{k+\frac{1}{2}, i}-\gamma \nabla F_{i}\left(x_{k, i} ; \xi_{k, i}\right)$\;
	}
	\KwOut{$\frac{1}{n}\sum_{i=1}^{n}x_{K,i}$;
	\tcc{the D-PSGD additional needs a network-wide average comparing to CDSGD;}}
\end{algorithm}

\subsection{Dynamic Average Consensus}
The dynamic average consensus problem, in opposition to the more studied static consensus \cite{tsitsiklis1984problems}, is referred to the problem in which a set of autonomous agents aims to track the average of individually measured time-varying signals by local communication with neighbors.
It has already been widely used in various fields such as formation control \cite{yang2008multi}, sensor fusion \cite{spanos2005dynamic,olfati2005consensus,olfati2005distributed}, distributed estimation \cite{martinez2007distributed} and distributed tracking \cite{yang2007distributed}.

Some existing work has studied the dynamic average consensus problem regarding the continuous-time reference inputs \cite{spanos2005dynamic, freeman2006stability,olfati2005consensus,ren2007consensus}. In \cite{spanos2005dynamic}, the authors use standard frequency-domain techniques and show that their algorithm is able to track the average of ramp reference inputs with zero steady-state error. 
In the context of input-to-state stability, the authors of \cite{freeman2006stability} show that proportional dynamic average consensus algorithm can track with bounded steady-state error the average of bounded reference inputs with bounded derivatives. Besides, a proportional-integral dynamic average consensus algorithm is also designed to track the average of constant reference inputs with sufficiently small steady-state error. 
In \cite{olfati2005consensus}, the authors propose a dynamic consensus algorithm and apply it to design consensus filters. Their algorithm can track with some bounded steady-state error the average of a common reference input with a bounded derivative.
In \cite{ren2007consensus}, the authors study a problem similar to that in \cite{olfati2005distributed} but further assume that agents know the nonlinear model which generates the time-varying reference function.

While for the dynamic average consensus problem regarding the discrete-time reference inputs, the authors of \cite{zhu2010discrete} have proposed a class of discrete-time dynamic average consensus algorithms and analyzed their convergence properties. Their algorithms are able to track a class of time-varying reference inputs including polynomials, logarithmic-type functions, periodic functions and other functions whose $n$-th-order differences are bounded, for $n \geq 1$, with zero or sufficiently small steady-state error.

For our decentralized federated learning implementation in this paper, we employ the first-order dynamic average consensus (FODAC) algorithm  \cite{zhu2010discrete} (see \textbf{Algorithm \ref{fodac}}) for each user to track the \textit{average model} in the absence of central PS.

\section{System Model and Problem Formulation}
\label{sec3} 
For federated learning in a centralized topology, the most critical stage is that the central PS aggregates the model updates from different users and sends back the global model to them iteratively.
However, when it comes to a decentralized topology where there is no PS, one of the most difficult problem arises, i.e., how can we execute the \textit{global aggregation} phase in a fully decentralized way in the absence of a central parameter server?

In this section, we first formulate a user as a node in an undirected weighted graph in \ref{subsec3.1}. Then, how the users communicate is formulated as a communication model based on a symmetric and doubly stochastic symmetric matrix (also referred to as a mixing matrix) in \ref{subsec3.2}. Finally, we formally constitute the decentralized federated learning implementation as a minimization optimization problem in \ref{subsec3.3}.

\subsection{Node Model}
\label{subsec3.1}
A node model refers to a user containing its own local dataset and training its local model using the local training data.

Assume that there are $N$ nodes in a decentralized network topology, where each node is labeled by $i \in V=\{1, 2, \cdots, N\}$. 
We denote the local dataset on the $i$-th node ${D}_i$, then the whole dataset formed by all local dataset is ${D} = {D}_1 \cup \cdots \cup {D}_N$, where ${D}_i \cap {D}_j = \emptyset$  if $i\neq j$. 

The machine learning model of node $i$ at round $t\left(1\leq t\leq T\right)$ is represented by $\omega_i^t$. 
Generally, all nodes' machine learning models are required to be structured identical (i.e., the depth of the neural network, each layer's type and width, etc.) and to be initialized by the same parameters, i.e., $\omega_1^0 = \omega_2^0 = \cdots = \omega_N^0 = \omega^0$. 
Here the $\omega^0$ denotes model initialization, which usually follows Glorot-initialization \cite{glorot2010understanding} or He-initialization \cite{he2015delving} in Pytorch. 
For the \textit{local update} phase, each node trains its own $\omega_i^t$ using its local training dataset.
Hence, the intermediate model parameters generated by each node $i$ during the \textit{local update} phase can be regarded as a discrete-time reference signal $\boldsymbol{\omega}_i = \{\omega_i^0, \omega_i^1,\cdots, \omega_i^{T}\}$.
We also define $\bar{\omega}^t  = \frac{1}{N}\sum_{i=1}^{N}\omega_i^t $ the \textit{average model} over $N$ nodes at round $t$. 
However, to implement federated learning, a \textit{global aggregation} phase is still needed in addition to \textit{local update}. 
So, we are now faced with the problem of how to make nodes periodically synchronize the \textit{average model} in the absence of a central PS?
 
\subsection{Communication Model}
\label{subsec3.2}
A communication model in this section refers to two rules that govern the information exchange between all nodes. 
One is a connectivity rule ensuring that the information of each node influences the information of any other nodes. 
The other is a rule on connection weights that a node uses when combing its information with the information received from its neighbors.

We represent a decentralized network topology formed by $N$ nodes at round $t$ as an undirected graph $\mathcal{G}\left(t\right) \coloneqq \left(V,E\left(t\right)\right)$, where the $E(t)\subset V\times V$ is an edge set. Besides, node $i$ and $j$ are called neighbor nodes to each other if and only if  $(i,j) \in E(t)$, which further means that node $i$ and node $j$ can directly communicate with each other at round $t$.
For the connectivity rule, the $\mathcal{G}(t)$ is required to be a connected graph \cite{de2007old}, such that the information of node $i$ can influence the information of any other nodes directly or indirectly.
While for the connection weights, we further define a mixing matrix $\mathbf{W}(t) = \left[\mathbf{w}_{ij}(t)\right] \in \mathbb{R}^{N \times N} $ to denote the connection weights between nodes in $\mathcal{G}(t)$, where 
\begin{equation}
	\left\{ {\begin{array}{*{20}{c}}
				{0 < \mathbf{w}_{ij}(t) < 1, \text{if } (i,j) \in E(t)}\\
				{\mathbf{w}_{ij}(t) = 0,\text{if } (i,j) \notin E(t)}
				\end{array}.}\right.
\end{equation}
For the sake of convergence, the $\mathbf{W}(t)$ is required to be a doubly stochastic symmetric matrix, i.e., it satisfies $\mathbf{W}(t)\mathbf{1} = \mathbf{1}$ and $\mathbf{1}^\mathrm{T}\mathbf{W}(t) = \mathbf{1}^\mathrm{T}$. Here $\mathbf{1} \in \mathbb{R}^N $ is the column vector whose entries are all ones, $\mathrm{T}$ indicates the matrix transpose operator.
The doubly stochastic property ensures a weighted average where the weight sum is 1 when a node combines other nodes' information. 
 
\subsection{Problem Formulation}
\label{subsec3.3}
On the basis of node model and communication model, a remaining problem is how to effectively execute \textit{global aggregation} phase without the help of a central PS? In other words, how to enable each node to synchronize the \textit{average model} in the absence of the PS?
In what follows, we show that implementing \textit{global aggregation} phase in a decentralized topology can be converted into a dynamic average consensus problem and hence a dynamic consensus algorithm can be employed to effectively solve it.  

Specifically, by treating the \textit{local update} phase as a discrete-time process and taking the $\boldsymbol{\omega}_i = \{\omega_i^0, \omega_i^1,\cdots, \omega_i^{T}\}$ as the discrete-time reference input on node $i$, to synchronize the \textit{average model} is to track the average of all nodes' reference inputs, which can be also referred to as a dynamic average consensus problem, whose 
objective can be formulated as:
\begin{equation}
	\label{eq4}
	\min_{\boldsymbol{x}^t}\left\|\boldsymbol{x}^{t}-\bar{\omega}^{t-1}\mathbf{1}\right\|_2^2
\end{equation}
Here $x_i^t$ denotes the estimation of node $i$ at round $t$, $\bar{\omega}^t$ denotes the \textit{average model} at round $t$, and $\boldsymbol{x}^t$ is a vector denoting all the estimations on $N$ nodes at round $t$   $\boldsymbol{x}^t=\left[x_1^t;x_2^t;\cdots;x_N^t\right]$. 
In this paper, we advocate the first-order dynamic average consensus (FODAC) algorithm \cite{zhu2010discrete} (see \textbf{Algorithm \ref{fodac}}) to solve problem \eqref{eq4} during the decentralized federated learning process.

In addition to the consensus problem \eqref{eq4}, another objective for federated learning is to minimize the global loss function shown by \eqref{eq2}. 
So, we can summarize the objective of the decentralized federated learning as follows:
\begin{equation}
	\label{eq5}
	\left\{\begin{array}{l}
		\min \limits_{\omega} f(\omega) \coloneqq \sum_{i=1}^{N} \frac{\left|D_i\right|}{\left|D\right|} f_{i}(\omega) \\
		\min\limits_{\boldsymbol{x}^t}\left\|\boldsymbol{x}^{t}-\bar{\omega}^{t-1}\mathbf{1}\right\|_2^2
	\end{array}\right.
\end{equation}
Moreover, if each node holds the same number of training samples, \eqref{eq5} can be rewritten as:
\begin{equation}
	\label{eq6}
	\left\{\begin{array}{l}
		\min \limits_{\omega} f(\omega) \coloneqq \frac{1}{N} \sum_{i=1}^{N} f_{i}(\omega) \\
		\min\limits_{\boldsymbol{x}^t}\left\|\boldsymbol{x}^{t}-\bar{\omega}^{t-1}\mathbf{1}\right\|_2^2
	\end{array}\right.
\end{equation}
In section \ref{sec4}, we design a DACFL algorithm to solve this problem.

\section{Algorithm Design}
\label{sec4}
To facilitate our proposed solution, we need to first construct a symmetric doubly stochastic matrix. 
Therefore, we first demonstrate how we construct the symmetric doubly stochastic matrix. 
Then, a brief introduction to the first-order dynamic average consensus algorithm is given.
And last, we propose the DACFL training algorithm for our decentralized FL implementation.

\subsection{Construct a Symmetric Doubly Stochastic Matrix}
A symmetric doubly stochastic matrix is defined as a square matrix $\mathbf{W}$ which meets $\mathbf{W}\mathbf{1} = \mathbf{1}$, $\mathbf{1}^\mathrm{T}\mathbf{W} = \mathbf{1}^\mathrm{T}$, $\forall \mathbf{w}_{ij} \in \mathbf{W}\geq 0$ \cite{tsitsiklis1984problems,chen2012fast}.
Intuitively, a very simple matrix that meets these conditions can be $\mathbf{W} \coloneqq \left[\frac{1}{n}\right]^{n \times n}$. 
To generate a more random matrix $\mathbf{W}$, we take the following heuristic method to construct the mixing matrix used in this paper: (i) generate a doubly stochastic matrix $\mathbf{A}$; (ii) transpose the matrix $\mathbf{A}$ and get  $\mathbf{A}^{\mathrm{T}}$; (iii) symmetrization: $\mathbf{W}=\frac{1}{2}\left(\mathbf{A}+\mathbf{A}^{\mathrm{T}}\right)$. 
The heuristic construction method is shown in \textbf{Algorithm \ref{construction}}\footnote{For the specific ``sparse matrix'' in section \ref{sec6}, it is generated by the Sinkhorn-Knopp algorithm \cite{knight2008sinkhorn}.}.

\begin{algorithm}
	\caption{A heuristic method to construct a symmetric doubly stochastic matrix}
	\label{construction}
	\KwIn{Dimension: $n$;}
	\KwOut{Matrix: $\mathbf{W}$;}
	$\mathbf{A}(1,1)=\rm{\textbf{rand}}$\;
	\For{$i=2:n-1$}{
		$d=1-\rm{\textbf{sum}}\left(\mathbf{A}\left(1,1:i-1\right)\right)$\;
		$\mathbf{A}(1,i)=d\times \rm{\textbf{rand}}$\;
	}
	
	\For{$i=2:n-1$}{
		$d=1-\rm{\textbf{sum}}\left(\mathbf{A}\left(1:i-1,1\right)\right)$\;
		$\mathbf{A}(i,1)=d\times \rm{\textbf{rand}}$\;
	}
	\For{$i=2:n-1$}{
		\For{$j=2:n-1$}{
			$d_1=1-\rm{\textbf{sum}}\left(\mathbf{A}(i,1:j-1)\right)$\;
			$d_2=1-\rm{\textbf{sum}}\left(\mathbf{A}(1:i-1,j)\right)$\;
			$d=\rm{\textbf{min}(d_1,d_2)}$\;
			$\mathbf{A}(i,j)= d \times \rm{\textbf{rand}}$\;
		}
	}
	\For{$i=1:n-1$}{
		$\mathbf{A}(n,i)=1-\rm{\textbf{sum}}(\mathbf{A}(1:n-1,i))$\;
	}
	\For{$i=1:n$}{
		$\mathbf{A}(i,n)=1-\rm{\textbf{sum}}(\mathbf{A}(i,1:n-1))$\;
	}
	\If{$\mathbf{A}(n,n)<0$}{
		\rm{\textbf{Repeat} line 1-22}\;
	}
	$\mathbf{W}=\frac{1}{2}\left(\mathbf{A}+\mathbf{A}^{\mathrm{T}}\right)$\;
	\tcc{\rm{\textbf{rand}}: a function that randomly generates a float number in $(0,1)$;\\
	\rm{\textbf{sum}}: a sum function;\\
	\rm{\textbf{min}: a minimum function;}:
	}
\end{algorithm}

\subsection{First-Order Dynamic Average Consensus}
As is demonstrated in section \ref{subsec3.3}, by treating the intermediate models generated during the \textit{local update} phase as the discrete-time reference inputs, to synchronize the \textit{average model} can be also referred to as a dynamic average consensus problem \eqref{eq4}. 
To solve this problem, \cite{zhu2010discrete} has proposed a first-order dynamic average consensus algorithm (FODAC) and has proved that the FODAC tracks the average state with either a zero steady-state error or an upper-bounded steady-state error (cf. \cite{zhu2010discrete}). 
\textbf{Algorithm \ref{fodac}} briefly shows the pseudocode of FODAC.

\begin{algorithm}
	\caption{First-order dynamic average consensus \cite{zhu2010discrete}}
	\label{fodac}
	\KwIn{Reference inputs of $N$ nodes: $r_i(t),(i=1,2,\cdots,N)$;}
	\KwOut{Consensus states of $N$ nodes: $x_i(T),(i=1,2,\cdots,N)$;}
	\textbf{Initialize:} mixing matrix $\mathbf{W}=\left[\mathbf{w}_{ij}\right]\in \mathbb{R}^{N \times N}$; initialized consensus state: $x_i(0)=r_i(0)$; number of iterations: $T$\;
	\For{$t=0,1,2,\cdots,T$}{
		$x_i(t+1)=x_i(t)+\sum_{j\neq i}\mathbf{w}_{ij}\left(x_j(t)-x_i(t)\right)+\Delta r_i(t)$\;
		\tcc{$\Delta r_i(t)$ indicates the first-order difference of $r_i(t)$: $\Delta r_i(t)=r_i(t)-r_i(t-1)$;}
	}
\end{algorithm}

\subsection{DACFL Algorithm}
As is formulated in \eqref{eq5} (or \eqref{eq6}), the objective of the decentralized federated learning implementation is to simultaneously minimize a global loss function $f(\omega)$ and solve a dynamic average consensus problem. 
For the former sub-problem, a distributed stochastic gradient descent can be used here-in, while for the latter sub-problem, we employ the FODAC to solve it. By combining the distributed gradient descent and the FODAC method, we devise an algorithm called \textit{Dynamic Average Consensus based Federated Learning} (DACFL).

Three main steps constitute our DACFL algorithm: 
(i) each node trains its own model using its local data; 
(ii) each node computes a neighborhood weighted average model $\omega_i^{t'}$ by exchanging its intermediate model with its neighbors;
(iii) each node performs the FODAC to track the \textit{average model} $\bar{\omega}^t$.
More specifically, step (i) can be also referred to as the \textit{local update} phase in federated learning. 
In step (ii), we further use the neighborhood weighted average model $\omega_i^{t'}$ to reinitialize the user's local model (line 6, \textbf{Algorithm \ref{dacfl}}), which also differentiates CDSGD and D-PSGD. As is empirically demonstrated in our experimental results, such re-initialization is more robust to sparse topology and non-i.i.d data since it to some extent avoids the local overfitting problem.
In step (iii), we employ the FODAC to estimate the \textit{average model} of all nodes in a fully decentralized manner, which helps handle the model fusion in \textit{global aggregation} phase without a central PS.
The complete DACFL training algorithm is shown in \textbf{Algorithm \ref{dacfl}}. 

\begin{algorithm}
	\caption{DACFL (our solution)}
	\label{dacfl}
	\KwIn{mixing matrix: $\mathbf{W}(t)=\left[\mathbf{w}_{ij}(t)\right]\in \mathbb{R}^{N \times N}$;
	      initialized models of $N$ nodes: $\omega_1^0=\omega_2^0=\cdots=\omega_N^0$;
          initialized consensus states of $N$ nodes: $x_1^0=x_2^0=\cdots=x_N^0$; number of communication rounds $T$; number of nodes $N$; learning rate $\lambda$;}
     \KwOut{consensus states of $N$ nodes: $x_1^T,x_2^T,\cdots,x_N^T$;}
     \textbf{Initialize:} $\boldsymbol{x}^0\coloneqq\left[x_1^0,x_2^0,\cdots,x_N^0\right]; \boldsymbol{\omega}^0\coloneqq\left[\omega_1^0,\omega_2^0,\cdots,\omega_N^0\right]$; $\boldsymbol{x}^0=\boldsymbol{\omega}^0$\;
     \For{each node $i\in 1,2,\cdots,N$ parallel}{
     	\For{$t=0,1,2,\cdots,T-1$}{
     		Compute the neighborhood weighted average model: $\omega_i^{t'}=\sum_{j=1}^{N}\mathbf{w}_{ij}(t)\omega_j^t$\;
     		Randomly sample a batch training data $\zeta_i^t$ from local data on $i$-th node\;
     		Use neighborhood weighted average model to reinitialize and  update the local model: $\omega_i^{t+1} \leftarrow \omega_i^{t'}-\lambda \nabla f_i\left(\omega_i^{t'};\zeta_i^t\right)$\; 
     		\tcc{$f_i(\cdot)$ indicates the loss function on node $i$; $\nabla$ indicates the gradient operator;}
     		Perform the average consensus by FODAC: $\Delta \omega_i^t=\omega_i^t-\omega_i^{t-1}$; \tcc{$\omega_i^{-1}=\omega_i^0$} $x_i^{t+1}=\sum_{j=1}^{N}\mathbf{w}_{ij}(t)x_j^t + \Delta \omega_i^t$\;
     	}
     }
\end{algorithm}

\section{Convergence Analysis}
\label{sec5}
Without loss of generality, each user is assumed to hold the same number of training samples in this paper such that we can denote the loss function $f(x)$ as

\begin{equation}
f(\omega) \coloneqq \frac{1}{N} \sum_{i=1}^{N} f_i(\omega) 
\end{equation}
To complete the analysis, we make the following assumptions on the loss functions.

\newtheorem{assumption}{Assumption}
\begin{assumption}
	\label{ass1}
	\textbf{(L-Smooth)} Each loss function $f_i(\omega)$ is L-smooth such that
	\begin{equation}
	f_i(v) \leq f_i(w)+(v-w)^{\mathrm{T}} \nabla f_i(w)+\frac{L}{2}\|v-w\|_{2}^{2}
	\end{equation}
\end{assumption}

In a conventional centralized learning way, the gradient $\nabla f(\omega)$ is computed based on the whole dataset. 
However, in a decentralized federated learning, each user holds a local data shard for its own training and gradient computation $\nabla f_i(\omega)$. 
To declare the user-wise gradients $\nabla f_i(\omega)$, we have \textbf{Assumption \ref{ass2}} and \textbf{\ref{ass3}} on the premise of i.i.d data.

\begin{assumption}
	\label{ass2}
	\textbf{(Bounded Gradients)} For each user $i$ and a randomly sampled batch data $\zeta_{i}$, there exists $G > 0$ such that
	\begin{equation}
	\mathbb{E}_{\zeta_{i} \sim \mathcal{D}_{i}}\left\|\nabla f_{i}\left(\omega ; \zeta_{i}\right)\right\|^{2} \leq G^{2}
	\end{equation}
\end{assumption}

\begin{assumption}
	\label{ass3}
	\textbf{(Uniform Gradient First-order Difference)} At any round $t$, define 
	\begin{align*}
		\Delta \mathcal{g}_i^t = \nabla f_i(\omega^{t'};\zeta_i) - \nabla f_i(\omega^{t-1'};\zeta_i)=\mathcal{g}_i^t-\mathcal{g}_i^{t-1}
	\end{align*}
	as the first-order difference of gradient, this paper assumes an i.i.d data distribution over all users for the convergence analysis, such that
	\begin{equation}
		\mathbb{E}\left[\Delta \mathcal{g}_i^t\right] = \mathbb{E}\left[\Delta \mathcal{g}_j^t\right] = \Delta \mathcal{g}^t, \forall i,j,  \, 1\le i,j \le N
	\end{equation}
\end{assumption}

Note that the mixing matrix $\mathbf{W}(t)$ of the communication network topology $\mathcal{G}(t)$ plays a pivotal role enabling our algorithm, which follows the assumptions.
\begin{assumption}
	\label{as1}
	\textbf{(Symmetric Doubly Stochastic Matrix)} At any round $t$, the mixing matrix $\mathbf{W}(t)$ is symmetric and doubly stochastic, such that
	\begin{equation}
		\mathbf{W}(t) \mathbf{1}=\mathbf{1}, \mathbf{1}^{\mathrm{T}} \mathbf{W}(t)=\mathbf{1}^{\mathrm{T}}, \mathbf{W}(t)=\mathbf{W}(t)^{\mathrm{T}}
	\end{equation}
\end{assumption}

Additionally, we make \textbf{Assumption \ref{ass4}} to ensure the average model parameter can be well tracked.
\begin{assumption}
	\label{ass4}
	\textbf{(Bounded First-order Differences of Model Parameter)} At any round $t$, there exists a constant $\theta > 0$ ensures an upper bound of each user's model parameters such that 
	\begin{equation}
	\left\|\omega_i^{t}-\omega_i^{t-1}\right\|^2 \leq \theta^2
	\end{equation}
\end{assumption}

Actually, the above statement can be guaranteed by choosing a sufficiently small  learning step size $\lambda$ and a proper activation function. 

Following the \textbf{Assumption \ref{ass4}}, we can easily draw a \textit{relatively bounded first-order difference} similar to Eq.(1) in \cite{zhu2010discrete} such that
\begin{align}
	\label{eq10}
	\Delta \omega_{\rm{max}}^t - \Delta \omega_{\rm{min}}^t \le \kappa
\end{align}
where $\kappa$ is the upper bound related to $\theta$ and 
\begin{align*}
	&\Delta \omega_i^t:= \omega_i^t - \omega_i^{t-1} \\
	&\Delta \omega_{\rm{max}}^t := \max_{i=1,2,\dots,N} \Delta\omega_i^t \\
	&\Delta \omega_{\rm{min}}^t := \min_{i=1,2,\dots,N} \Delta\omega_i^t
\end{align*}

The above \eqref{eq10} ensures the FODAC to track the average of the time-varying reference inputs, w.r.t.,  model parameters $\omega_i^t$, with a sufficiently small steady-state error.  For detail proof of the conclusion, please refer to \cite{zhu2010discrete}. Note, to track the average of matrix-form reference inputs can be reformulated as tracking the average of scalar-form reference inputs by considering the matrix as a set of element-wise scalars.

Following the above assumptions, we present the convergence rate of our proposed algorithm. In our result, we follow the convention in literature \cite{lian2017can} to use the average expected squared gradient norm to characterize the convergence rate.

\newtheorem{theorem}{Theorem}
\begin{theorem}
	\label{theo1}
	Following the aforementioned assumptions, we have the average expected squared gradient norm following
	\begin{equation}
		\label{eq11}
		\begin{small}
			\begin{aligned}
				\frac{1}{T}\sum_{t=0}^{T-1}\left\|\nabla f(\bar{\omega}^t)\right\|^2
				&\le \frac{2}{\lambda T}\mathbb{E}\left(f(\bar{\omega}^0) - f(\bar{\omega}^T)\right) + G^2 + \frac{\theta^2}{\lambda^2} + \frac{L\theta^2}{\lambda} \\
				&\le \underbrace{\frac{2}{\lambda T}\left(f(\bar{\omega}^0) - f^*\right)}_{:\rm{C}_0} + \underbrace{G^2 + \frac{\theta^2}{\lambda^2} + \frac{L\theta^2}{\lambda}}_{:\rm{C}_1}
			\end{aligned}
		\end{small}
	\end{equation}
	where $f^*$ denotes the minimum loss value of $f(x)$.
\end{theorem}

Note that in \eqref{eq11}, the average squared gradient norm is bounded by $\rm{C}_0+\rm{C}_1$, where the $\rm{C}_0$ gradually tends to $0$ when training iteration $T$ increases. In other words, the average squared gradient norm is bounded by a learning rate related term $\rm{C}_1$ when $T \to +\infty$. 
Until now, we have declared the convergence taking the average model parameter $\bar{\omega}^t$ as the input of loss function $f$. 
Moreover, the final output of DACFL algorithm $\boldsymbol{x}^T = [x_1^T, x_2^T, \dots, x_N^T]^{1 \times N}$ can track the $\bar{\omega}^T$ with a small steady-state error under the above assumptions. (Please refer to \cite{zhu2010discrete} for the convergence guarantee of FODAC.) 
This further offers a convergence guarantee for our DACFL solution.

The detailed proof of \textbf{Theorem \ref{theo1}} is deferred to the appendix of this paper.

\section{Experiments and Performance Evaluation}
\label{sec6}
In this section, we first declare the experimental setup and then evaluate the performance of DACFL on different datasets with vary topologies and data allocations. 
Note that, the curves of following experimental result are smoothed by Savitzky-Golay filter \cite{schafer2011savitzky}.

\subsection{Experimental Setup}
\subsubsection{Dataset}
Three publicly accessible datasets including MNIST \cite{deng2012mnist}, Fashion-MNIST (FMNIST) \cite{xiao2017fashion} and CIFAR-10 \cite{krizhevsky2009learning} widely used in image classification field are used for our performance validation in this paper.
(i) MNIST: a dataset includes 70,000 images of hand-written digits, totally 10 classes, with a training set consisting 60,000 examples and a test set consisting 10,000 examples, respectively. 
(ii) Fashion-MNIST: a new dataset comprises 28$\times$28 grayscale images of 70,000 fashion products from 10 categories. The training set has 60,000 images and the test set has 10,000 images.
(iii) CIFAR-10: a dataset consists totally 60,000 color images with three RGB channels which can be classified into 10 classes. The training set has 50,000 examples and the test set has 10,000 examples.

\subsubsection{Dataset Allocation}
We design two ways for data allocation in our experiments, i.e., i.i.d and non-i.i.d allocations. 
(i) i.i.d: each user is assigned the same number of training samples with a uniformly random distribution over 10 classes.
(ii) non-i.i.d: the training set is sorted by labels first and then divided into multiple shards with the same number of training samples. Finally, each user samples only 2 shards of data without replacement.
Note that in this setup, the non-i.i.d allocation only considers the \textit{class imbalance} of data \cite{wahab2021federated}. 

\subsubsection{Communication Network Topology}
As is demonstrated in section \ref{sec3}, we can represent the communication network topology with the mixing matrix. 
In our experiments, we design the topology from different perspectives.
(i) time-varying and time-invariant: for the time-invariant topology, we initialize the mixing matrix before training and keep it unchanged during the training process; while for the time-varying topology, we reconstruct the mixing matrix every 10 training rounds.
(ii) sparse and dense connectivity: for the sparse ($\psi$=0.5) topology, half entries of the mixing matrix are 0; while for the dense ($\psi$=1.0) topology, all entries in the mixing matrix are non-zeros. \footnote{Note here we call a matrix with `sparse' just for simplicity, which is not strictly equivalent to a ``sparse matrix". The ``sparse matrix'' here is generated by Sinkhorn-Knopp algorithm \cite{knight2008sinkhorn}.} 
All the above considered mixing matrices are symmetric and doubly stochastic.

\subsubsection{Neural Network Structure}
For the training models used to perform the image classification tasks, we use the same convolutional neural network (CNN) for MNIST and Fashion-MNIST which contains two 5$\times$5 convolutional layers (each layer is followed with a batch normalization and 2$\times$2 max pooling), a fully connected layer with ReLu activation and a final softmax output layer. 
For CIFAR-10 dataset, we use a CNN consisting two convolutional layers (each layer is followed with batch normalization, ReLu activation and 2$ \times$2 max pooling), two fully connected layers with ReLu activation and a final softmax output layer. 
Both CNN models are mended from \cite{mcmahan2017communication}.
Unless otherwise specified, some important hyper-parameters in our experiments are set as TABLE \ref{para-set}.

\subsubsection{Baselines and Performance Metric}
In this work, we compare our DACFL with a conventional centralized federated learning method, i.e., FedAvg and another two decentralized federated learning implementations called CDSGD \cite{jiang2017collaborative} and D-PSGD \cite{lian2017can}. 
For the FedAvg implementation in this paper, a centralized topology with the same number of users to the decentralized topology is considered, where all users are ensured to participate in each training round. 
For CDSGD and D-PSGD, the mixing matrix and other hyper-parameters are consistent with DACFL.
Additionally, we assume that there still exists a ``god node'' for D-PSGD to perform the network-wide model average, hence generating a global model used for the performance test.
To make the result more clear, we design two metrics including \textit{Average of Acc} and \textit{Var of Acc} to indicate the performances of different methods. 
Specifically, we respectively test each user's trained model (or estimated model in DACFL) and get the test accuracy of all users. Then the \textit{Average of Acc} is computed through averaging all users' test accuracy and \textit{Var of Acc} is the variance over all users' test accuracy. 
Generally, a superior decentralized federated learning method is expected to obtain a higher \textit{Average of Acc} and a smaller \textit{Var of Acc}.
Note that for FedAvg and D-PSGD where the final output is the only global model, the \textit{Average of Acc} is actually same with the test accuracy and the \textit{Var of Acc} is 0.

\begin{table}[t]
	\caption{Experimental parameters setting}
	\centering 
	\begin{tabular}{lc} 
		\hline 
		Parameter & Numerical value\\ 
		\hline
		Number of nodes: & 10\\
		Training rounds: & 100\\
		Local batch size: & 20\\
		Local epoch: & 1\\
		Decaying for learning rate: & 0.995\\  
		Loss function: & Cross Entropy\\  
		Learning rate: & MNIST/FMNIST: 0.001, CIFAR: 0.005\\
		\hline
	\end{tabular}
	\label{para-set}
\end{table}

\subsection{Why Choose FODAC? A Numerically Empirical Perspective}
In order to clarify how FODAC benefits our solution, some specific numerical experiments are designed in this section. More specifically, we separately apply FODAC, CDSGD and D-PSGD to track the average of two types of discrete-time inputs under three different mixing matrices. 

\textbf{Inputs I}: A class of discrete-time inputs with relatively large variance between each user, 
\[R_i(t) = \sin(t)+\left(\frac{1}{t}\right)^i+t+i\]

\textbf{Inputs II}: A class of discrete-time inputs with relatively small variance between each user, 
\[R_i(t) = \sin(t)+\left(\frac{1}{t}\right)^i+t\]
where $R_i(t)$ denotes the input of $i$-th user at time $t$, with $i\in \left\{1,2,\cdots,10\right\}$,$t \in \left\{1,2,\cdots,20\right\}$.

\begin{figure*}
	\centering
	\subfigure[Large variance inputs (\textbf{Inputs I})]{
		\label{large var}
		\begin{minipage}[b]{0.33\textwidth}
			\includegraphics[width=1\textwidth]{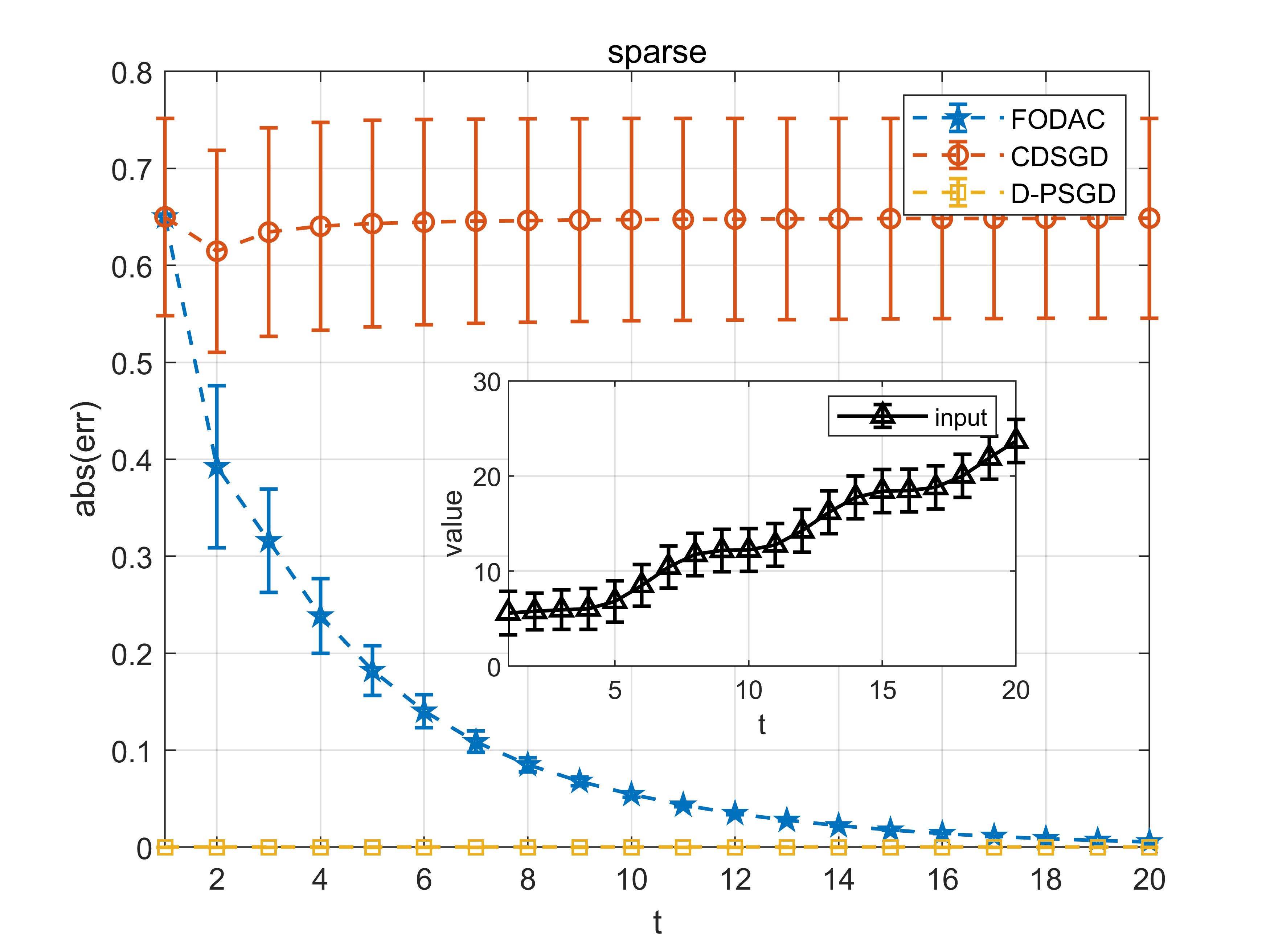}
		\end{minipage}
		\begin{minipage}[b]{0.33\textwidth}
			\includegraphics[width=1\textwidth]{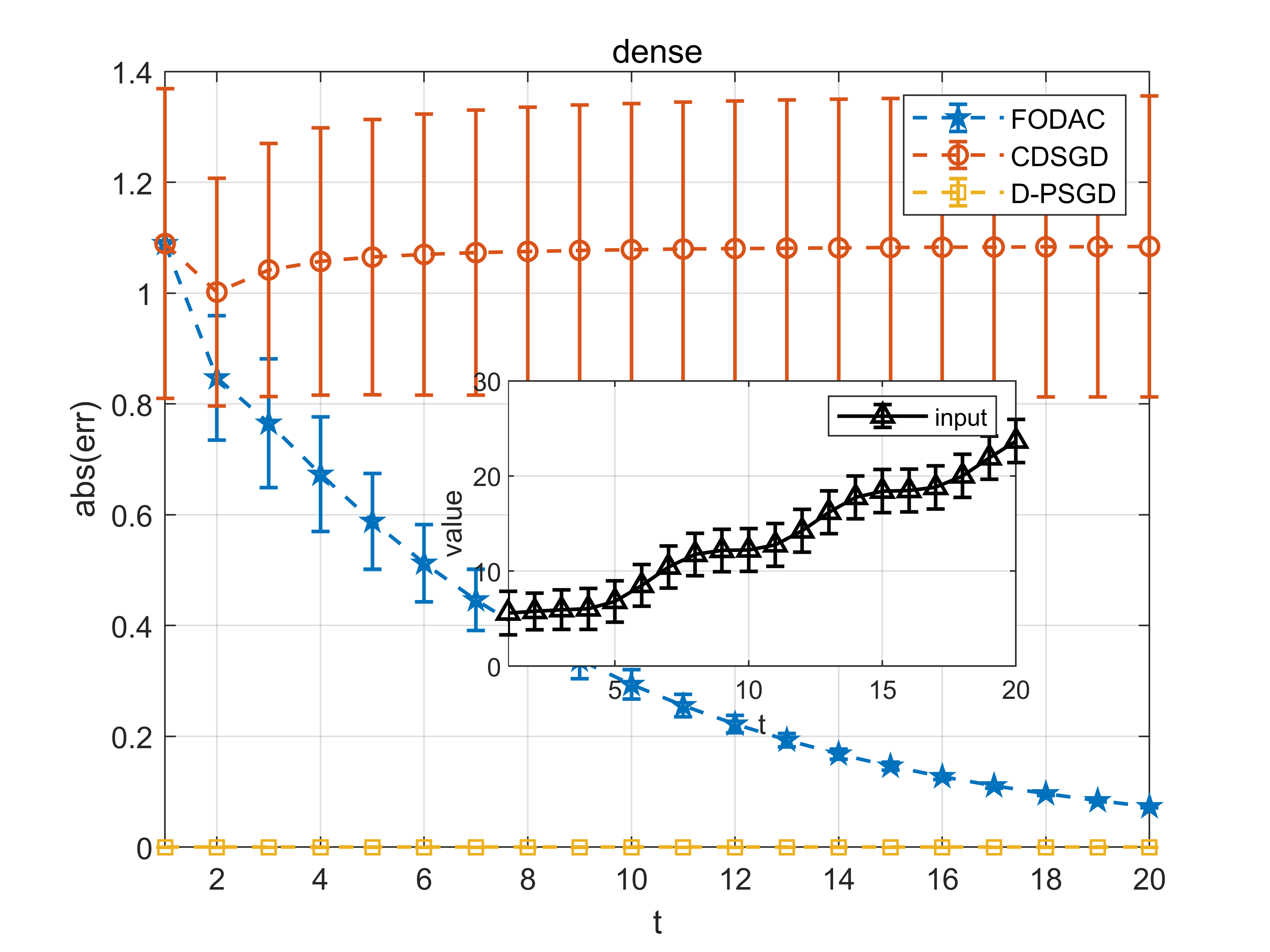}
		\end{minipage}
		\begin{minipage}[b]{0.33\textwidth}
			\includegraphics[width=1\textwidth]{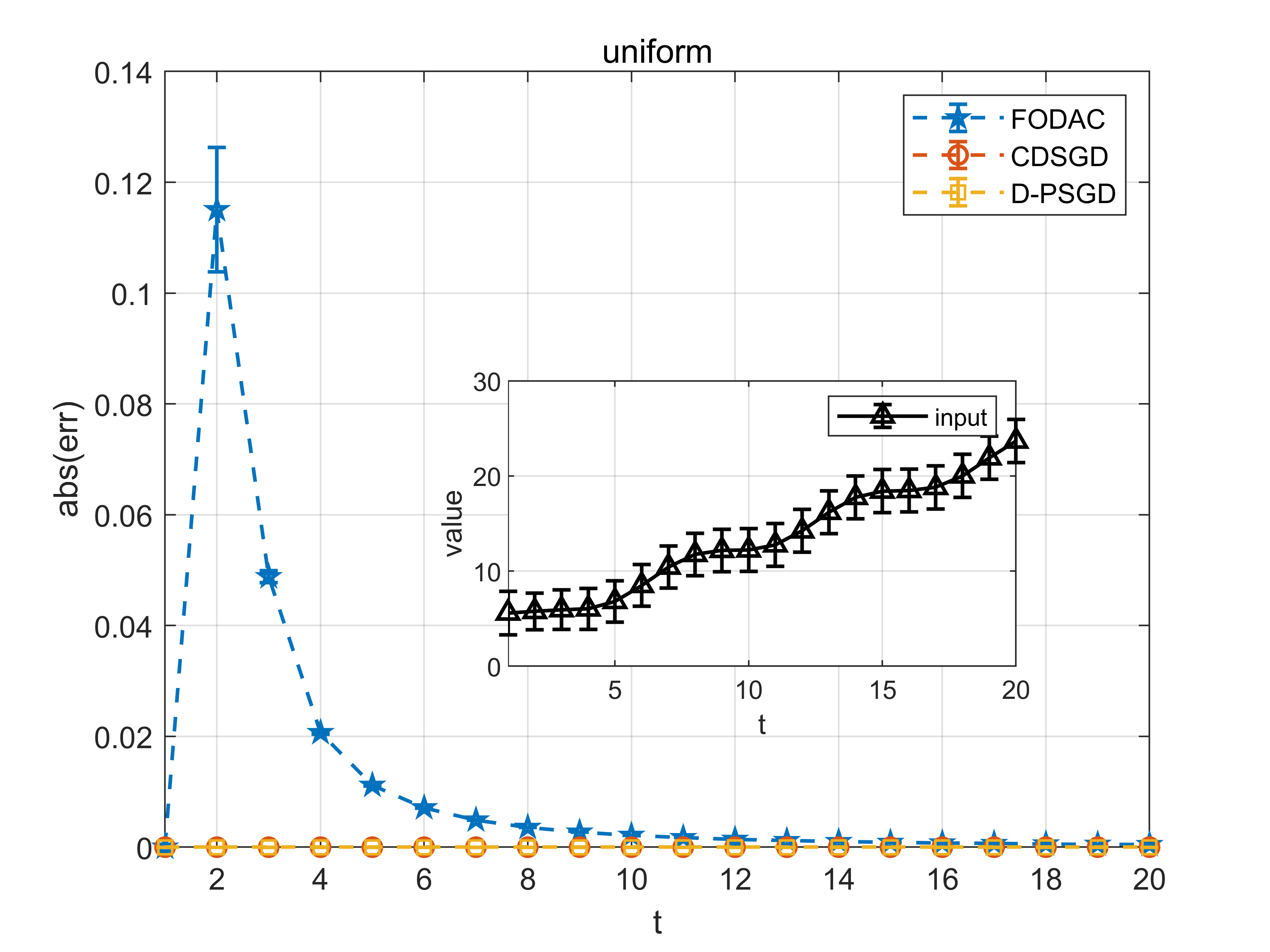}
		\end{minipage}
	}
	\subfigure[Small variance inputs (\textbf{Inputs II})]{
		\label{small var}
		\begin{minipage}[b]{0.33\textwidth}
			\includegraphics[width=1\textwidth]{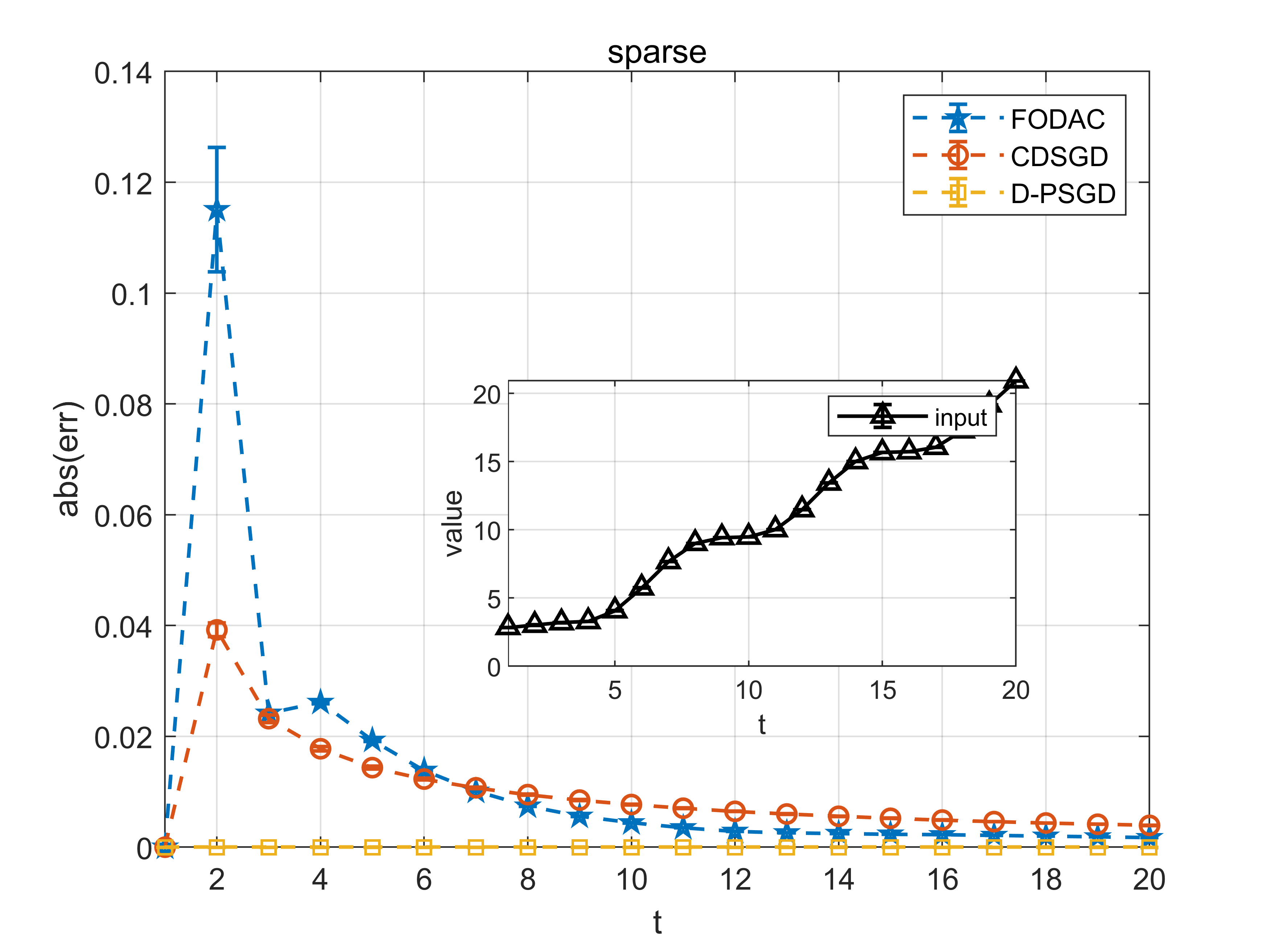}
		\end{minipage}
		\begin{minipage}[b]{0.33\textwidth}
			\includegraphics[width=1\textwidth]{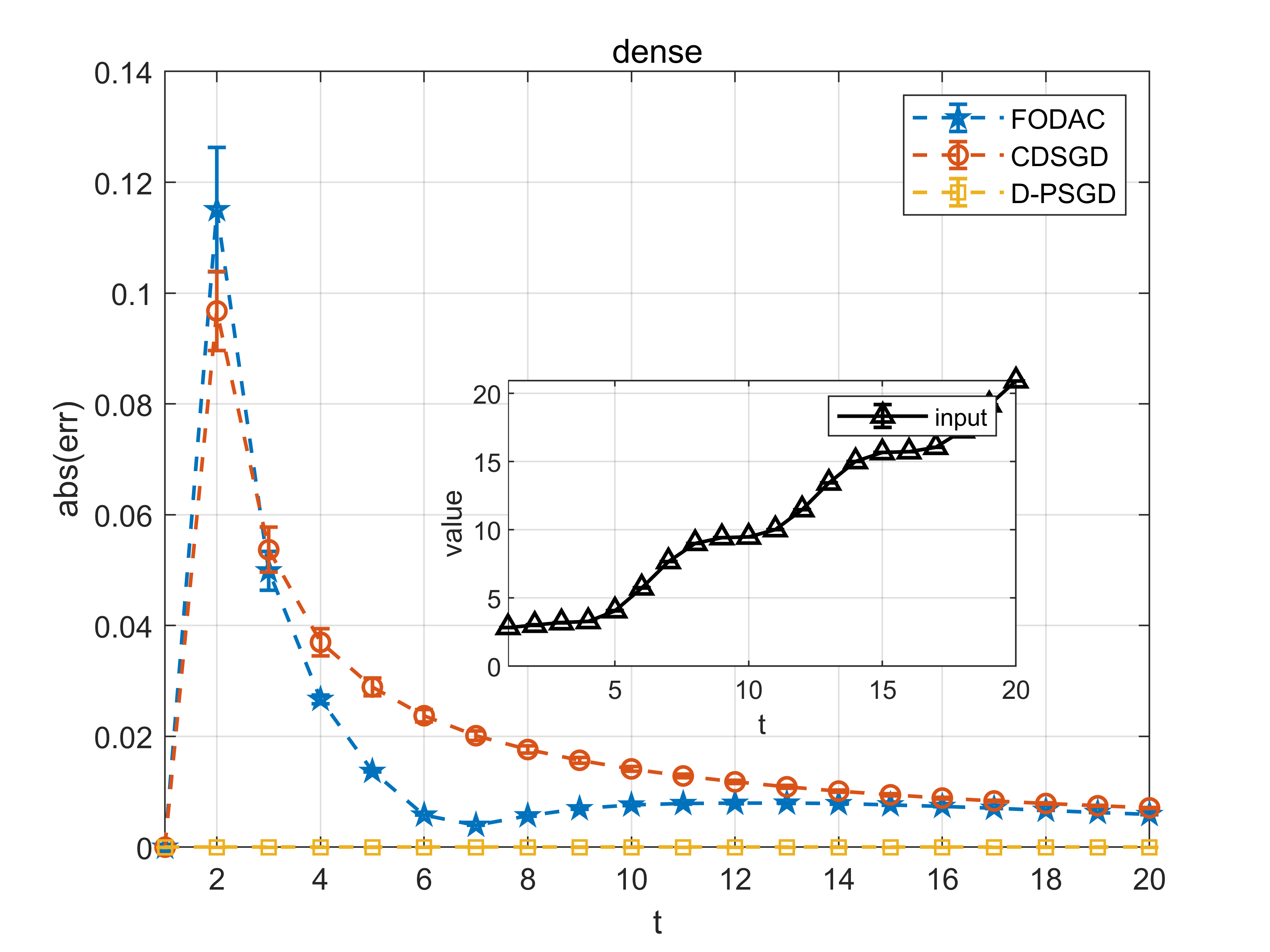}
		\end{minipage}
		\begin{minipage}[b]{0.33\textwidth}
			\includegraphics[width=1\textwidth]{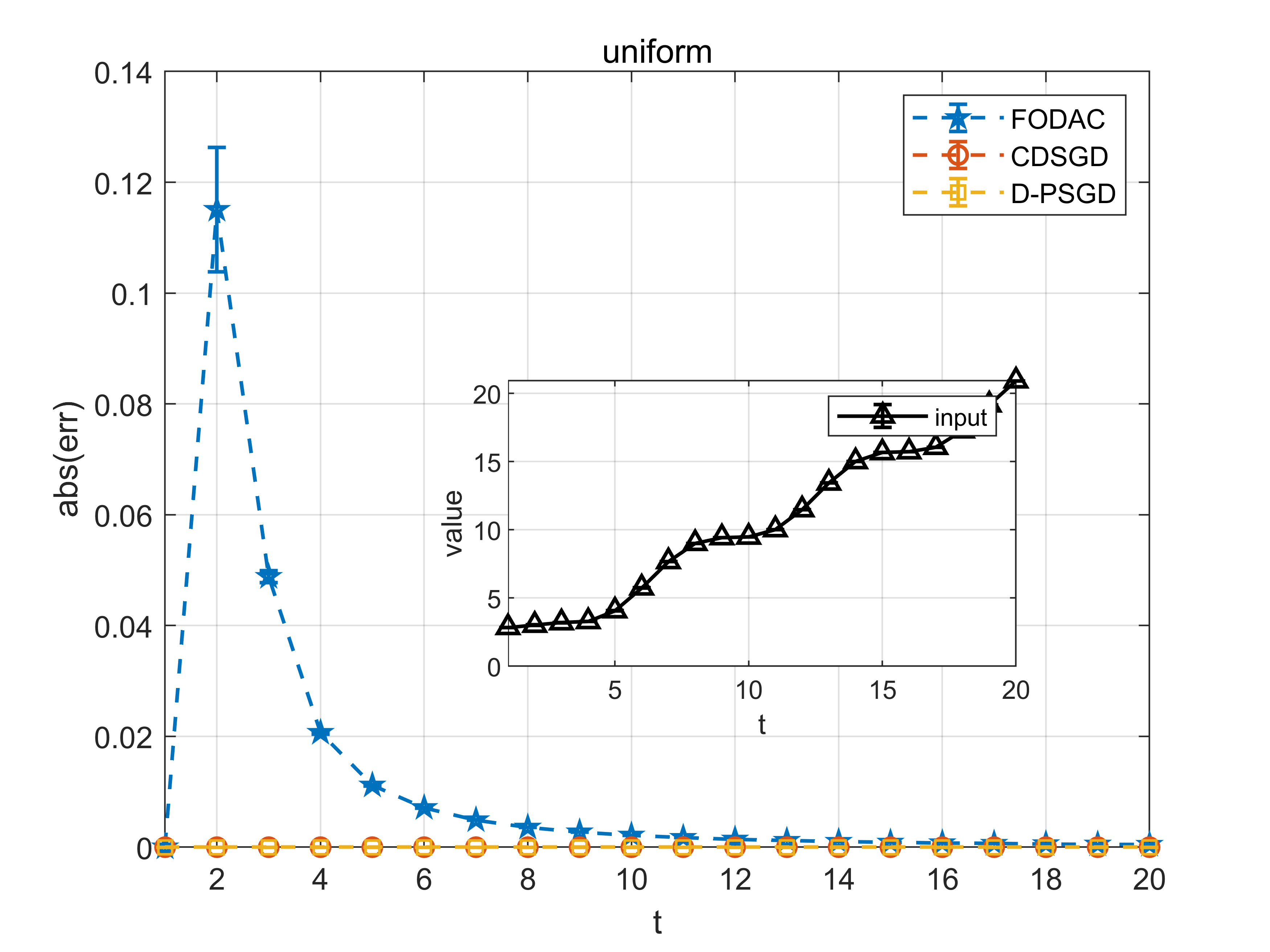}
		\end{minipage}
	}
	\caption{The result of approximating the average by different algorithms}
	\label{impact}
\end{figure*}

For the mixing matrices, (i) \textbf{sparse}: a 10$\times$10 mixing matrix with $\psi$=0.5, i.e., half entries are 0; (ii) \textbf{dense}: a 10$\times$10 mixing matrix with $\psi$=1.0, i.e., all entries are non-zeros; (iii) \textbf{uniform}: a 10$\times$10 mixing matrix with all entries being 0.1. 
Note all kinds of matrices considered here are symmetric doubly stochastic. 
For CDSGD, we take roughly the neighborhood weighted average as the estimated value (see line 6 in \textbf{Algorithm \ref{cdsgd}}); for D-PSGD, the estimated value used here is the result of executing an additional network-wide average on the estimation by CDSGD. While for FODAC, we take the consensus state (see line 3 in \textbf{Algorithm \ref{fodac}}) as the estimated result. Then, the absolute error can be computed by \[\mathrm{abs(err)} = \left| \bar{R}_i(t)-\hat{R}_i(t)\right|\]
where the $\hat{R}_i(t)$ denotes the estimated value and $\bar{R}_i(t)$ denotes the average of inputs with $\bar{R}_i(t)=\frac{1}{10}\sum_{i=1}^{10}R_i(t)$.
The results are shown in Fig. \ref{impact}.

As can be seen in fig. \ref{large var}, the FODAC is superior to CDSGD when users' reference inputs are with large variance under both sparse and dense mixing matrices. It can be concluded that if employing CDSGD to approximate the average of inputs, an estimating error and a large deviation between users are not negligible. 
While the CDSGD becomes feasible only when the inputs turn to be with small variance or when the mixing matrix is uniform. Nonetheless, the FODAC still slightly outperforms CDSGD from the perspective of convergence speed when users' inputs are with small variance under both sparse and dense mixing matrices, which are shown in Fig. \ref{small var}.
Note that D-PSGD outperforms the other two methods because it additionally executes a network-wide average which enables it to accurately compute the average value of the inputs. However, such a network-wide average is unacceptable when it comes to a fully decentralized topology where no PS helps to do this.

From this empirical result, we conclude that the FODAC method is more adaptable than CDSGD and D-PSGD on approximating the average in a fully decentralized topology. Actually, this also motivates us to employ FODAC to decentralized federated learning since (i) the common non-i.i.d data usually leads to a large variance between users' local models; (ii) a non-ideal communication topology (a not uniform mixing matrix) often arises in practical; (iii) there is no central PS to perform a network-wide model average.

\subsection{Performance on i.i.d Data}
\label{sec:iid}
\begin{figure*}[ht]
	
	\centering
	\subfigure[Average of Acc on MNIST]{
		\label{s iid1}
		\begin{minipage}[b]{0.31\textwidth}
			\includegraphics[width=1\textwidth]{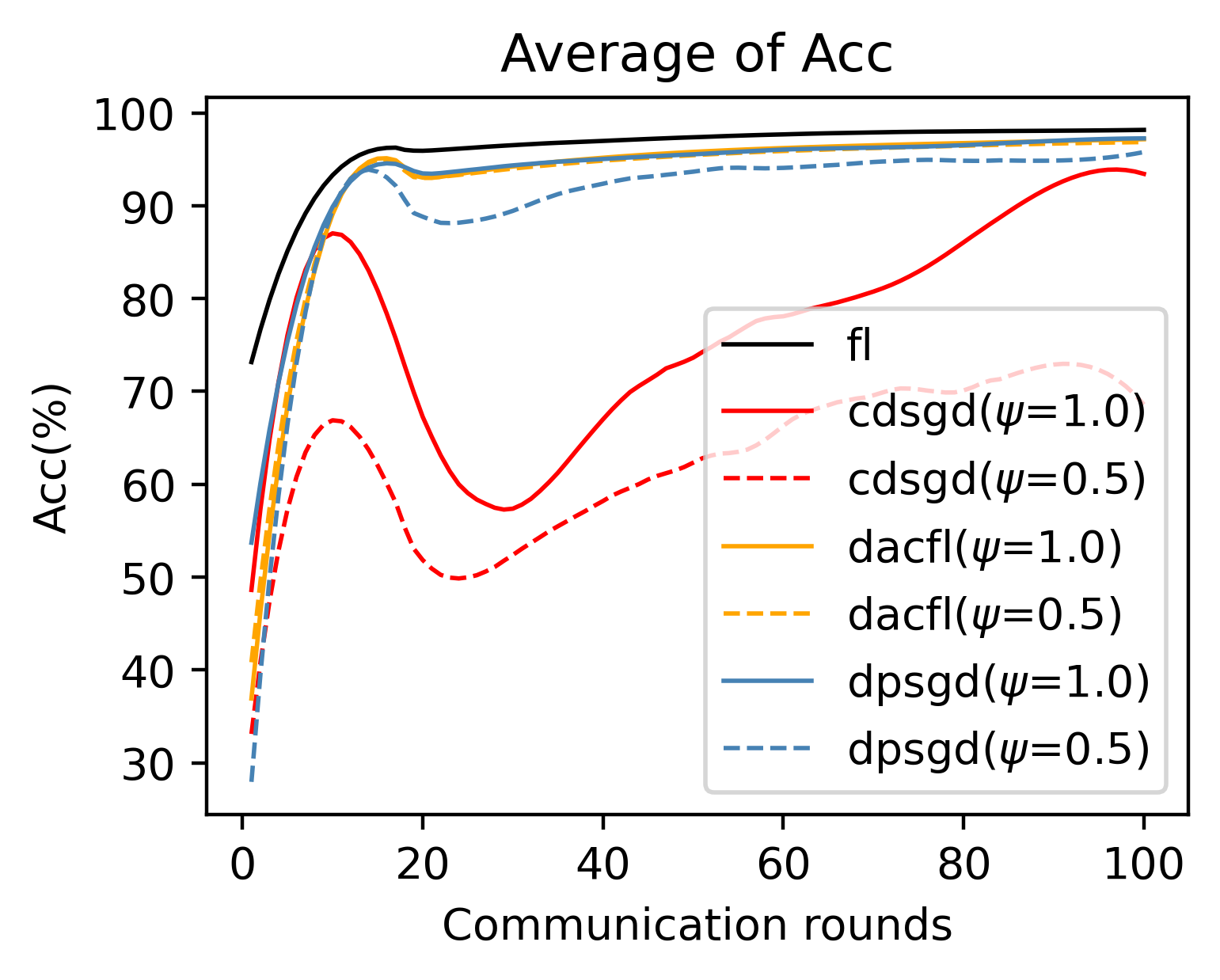}
		\end{minipage}
	}
	\subfigure[Average of Acc on FMNIST]{
		\label{s iid2}
		\begin{minipage}[b]{0.31\textwidth}
			\includegraphics[width=1\textwidth]{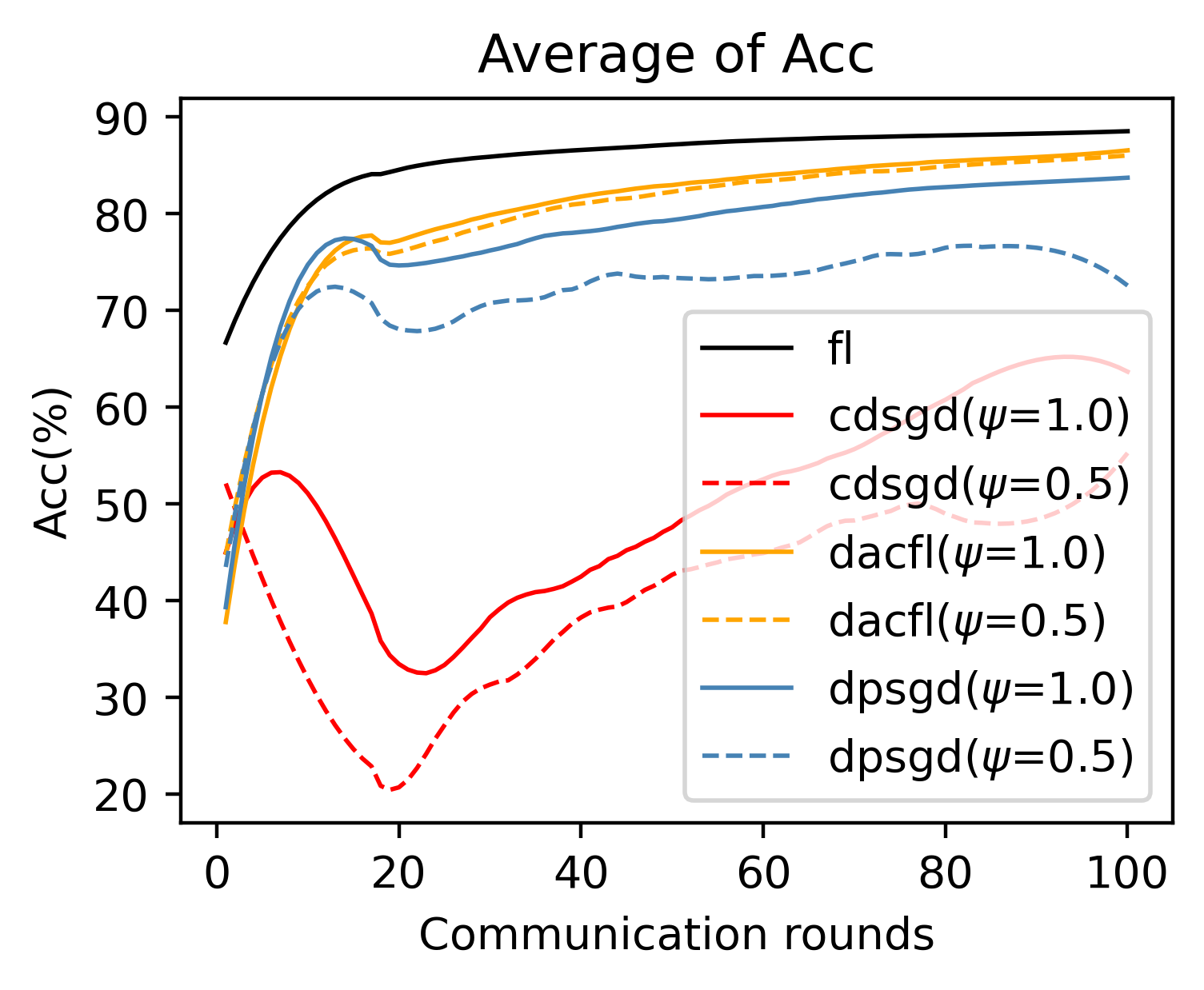}
		\end{minipage}
	}
	\subfigure[Average of Acc on CIFAR-10]{
		\label{s iid3}
		\begin{minipage}[b]{0.31\textwidth}
			\includegraphics[width=1\textwidth]{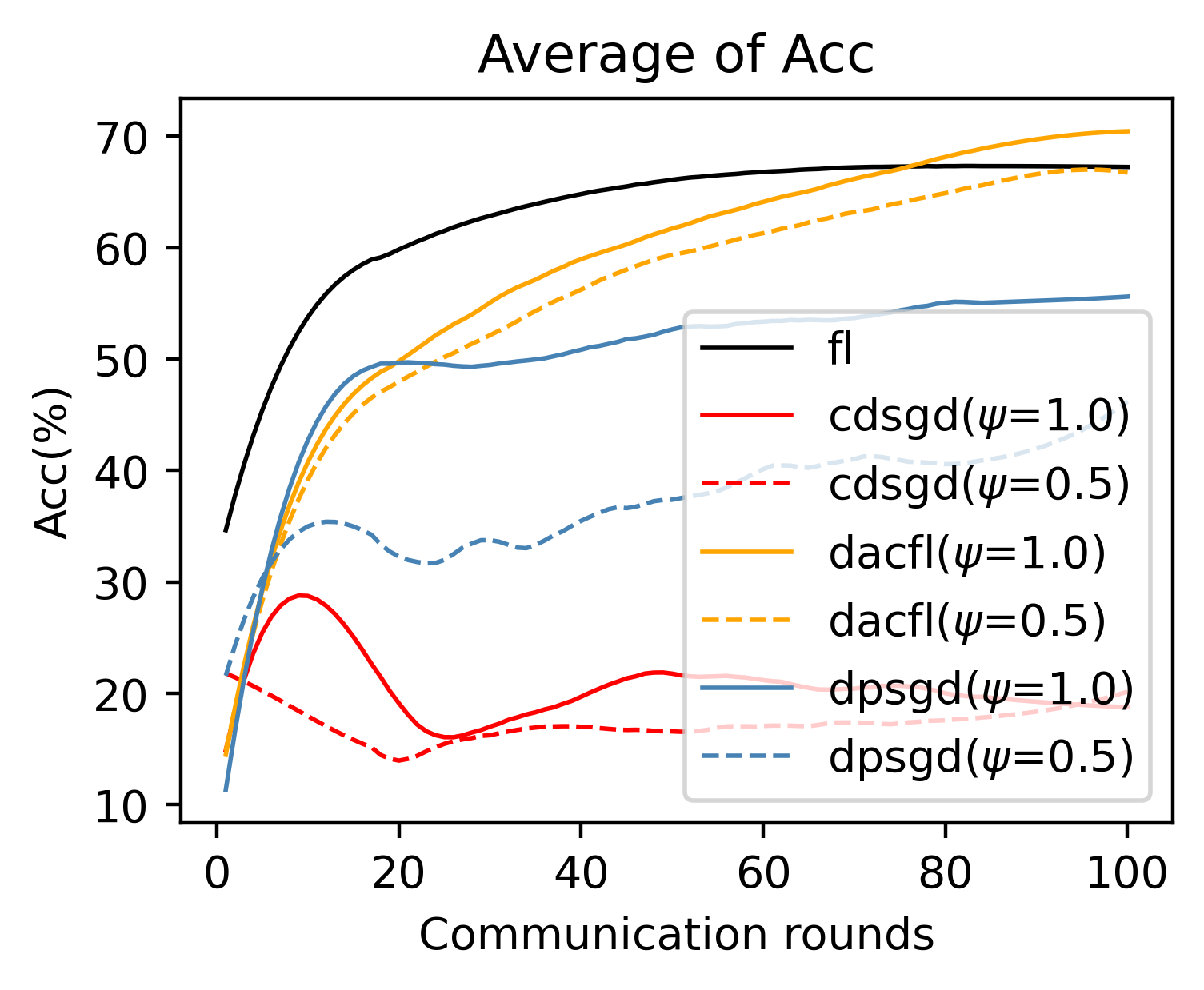}
		\end{minipage}
	}
	\subfigure[Var of Acc on MNIST]{
		\label{s iid4}
		\begin{minipage}[b]{0.31\textwidth}
			\includegraphics[width=1\textwidth]{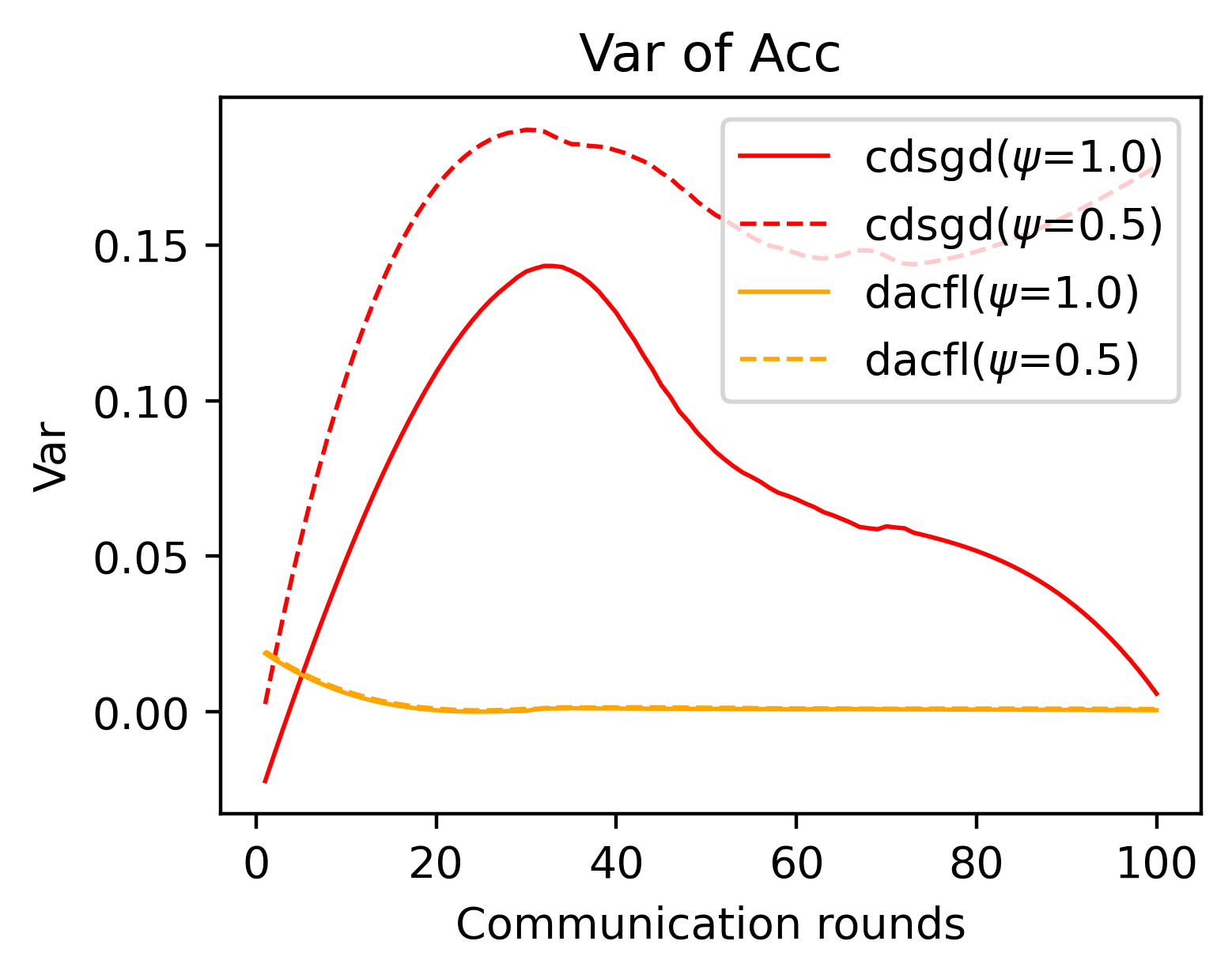}
		\end{minipage}
	}
	\subfigure[Var of Acc on FMNIST]{
		\label{s iid5}
		\begin{minipage}[b]{0.31\textwidth}
			\includegraphics[width=1\textwidth]{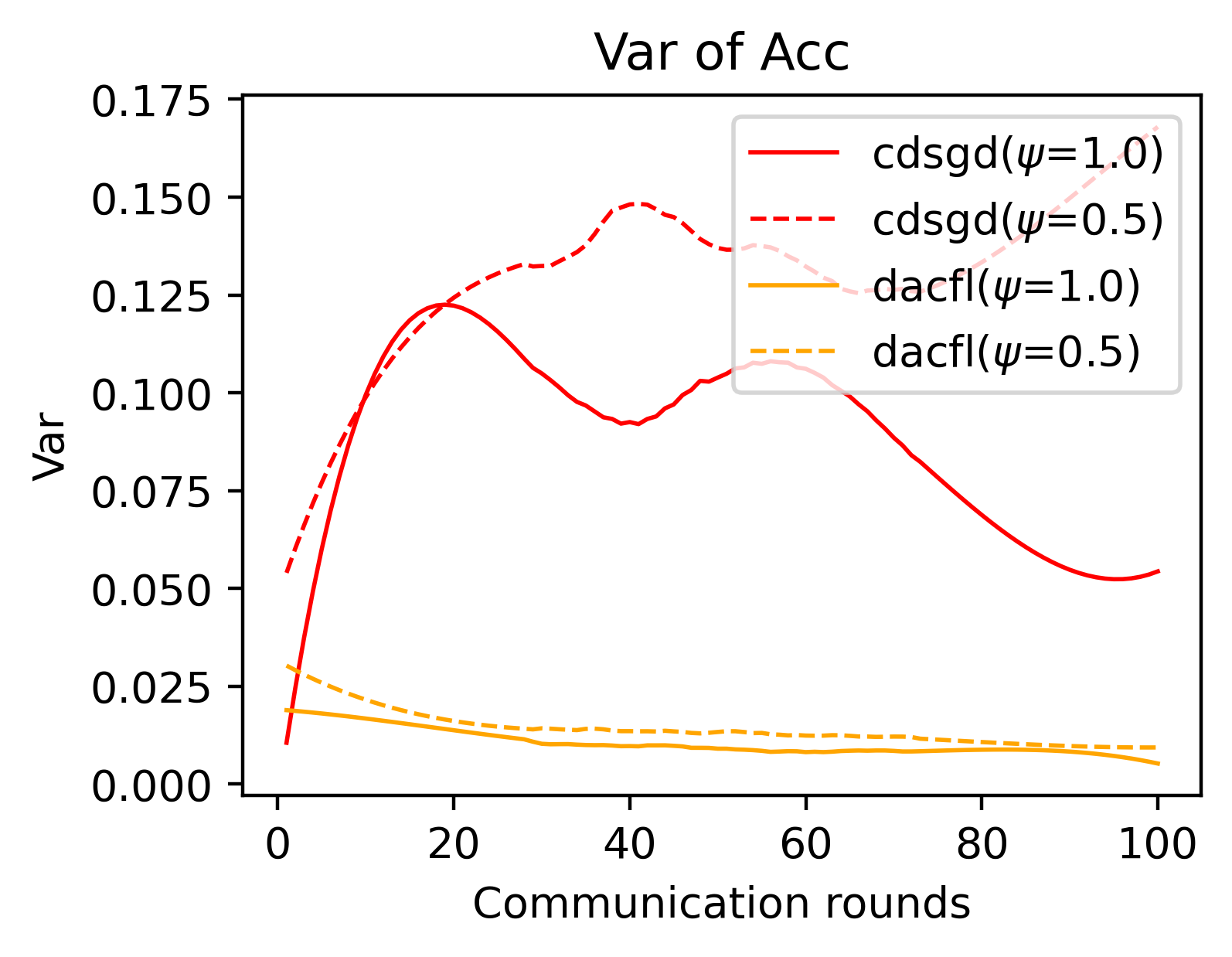}
		\end{minipage}
	}
	\subfigure[Var of Acc on CIFAR-10]{
		\label{s iid6}
		\begin{minipage}[b]{0.31\textwidth}
			\includegraphics[width=1\textwidth]{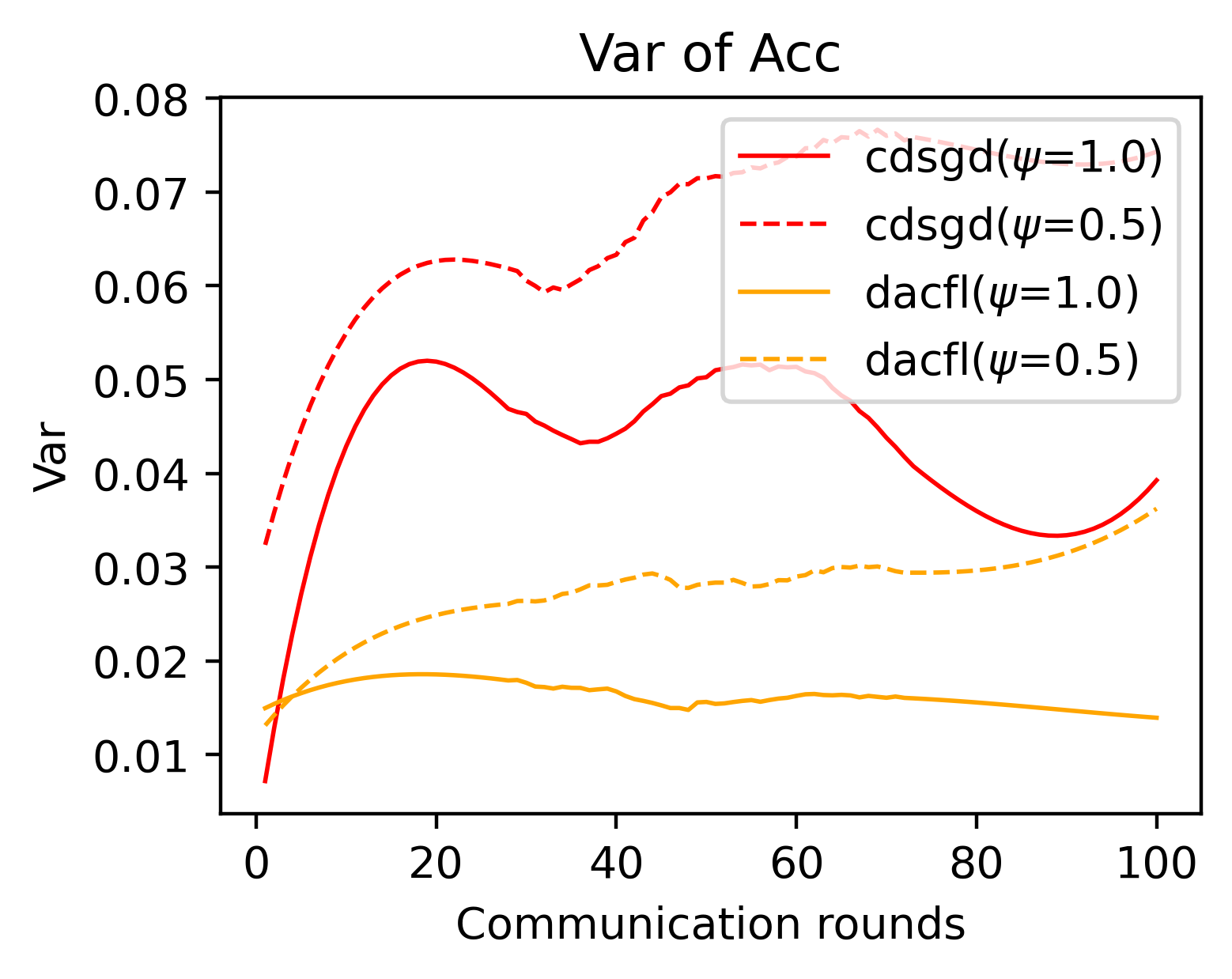}
		\end{minipage}
	}
	\caption{Performance comparison with i.i.d data and time-invariant topology (a communication round also corresponds to a training round in our experiments).} 	
	\label{fig:static i.i.d}
\end{figure*}

\begin{figure*}[htbp]
	
	\centering
	\subfigure[Average of Acc on MNIST]{
		\label{d iid1}
		\begin{minipage}[b]{0.31\textwidth}
			\includegraphics[width=1\textwidth]{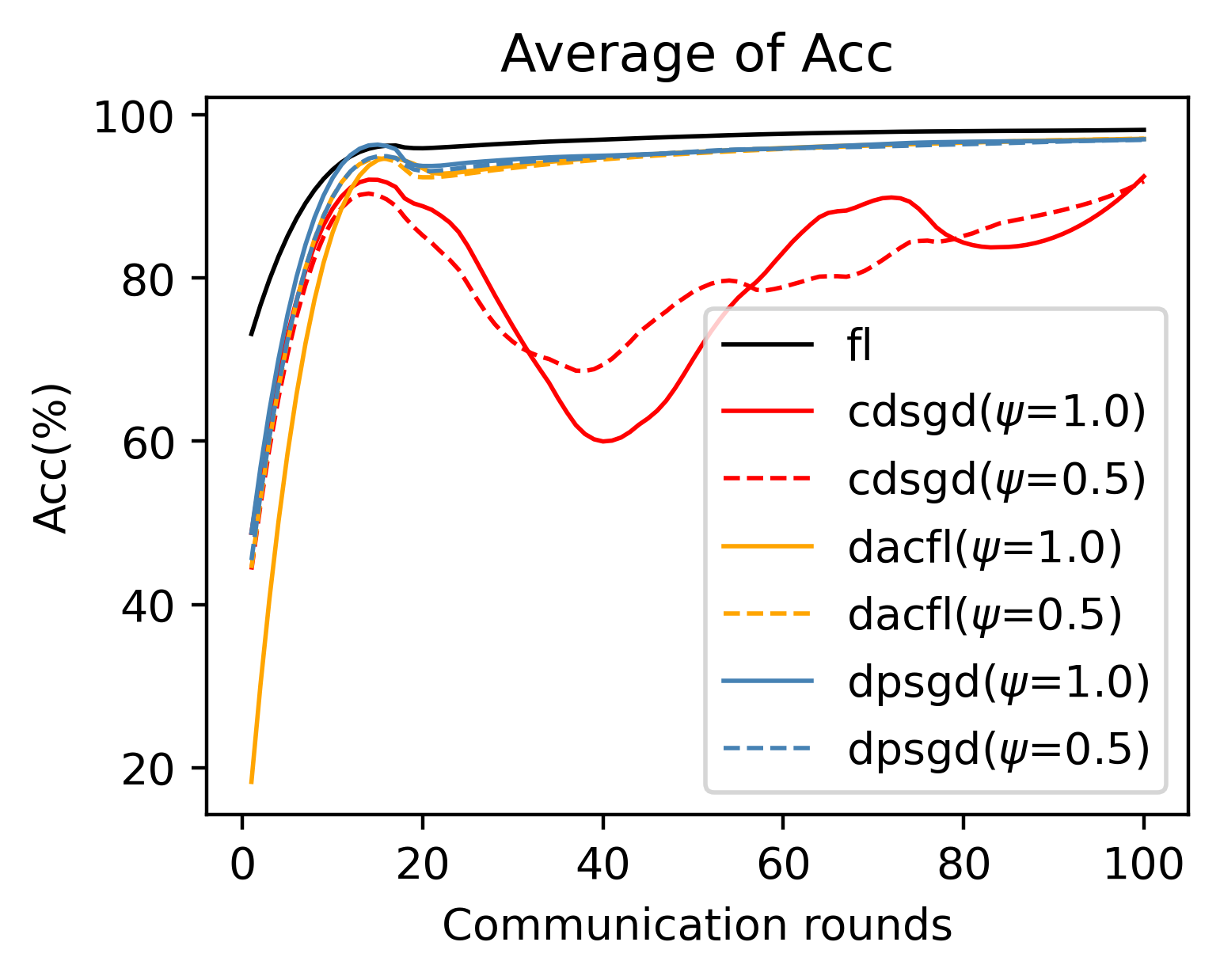}
		\end{minipage}
	}
	\subfigure[Average of Acc on FMNIST]{
		\label{d iid2}
		\begin{minipage}[b]{0.31\textwidth}
			\includegraphics[width=1\textwidth]{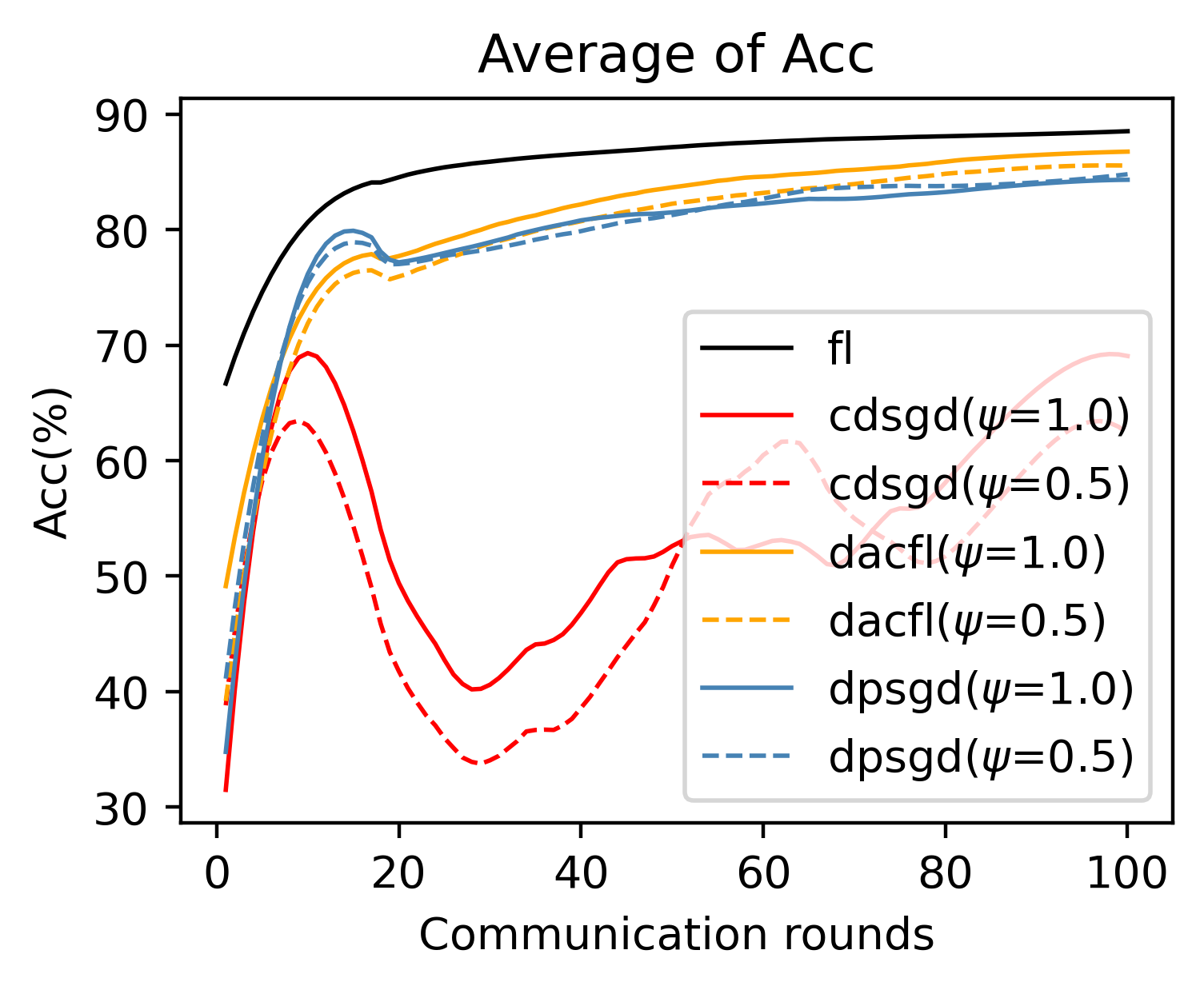}
		\end{minipage}
	}
	\subfigure[Average of Acc on CIFAR-10]{
		\label{d iid3}
		\begin{minipage}[b]{0.31\textwidth}
			\includegraphics[width=1\textwidth]{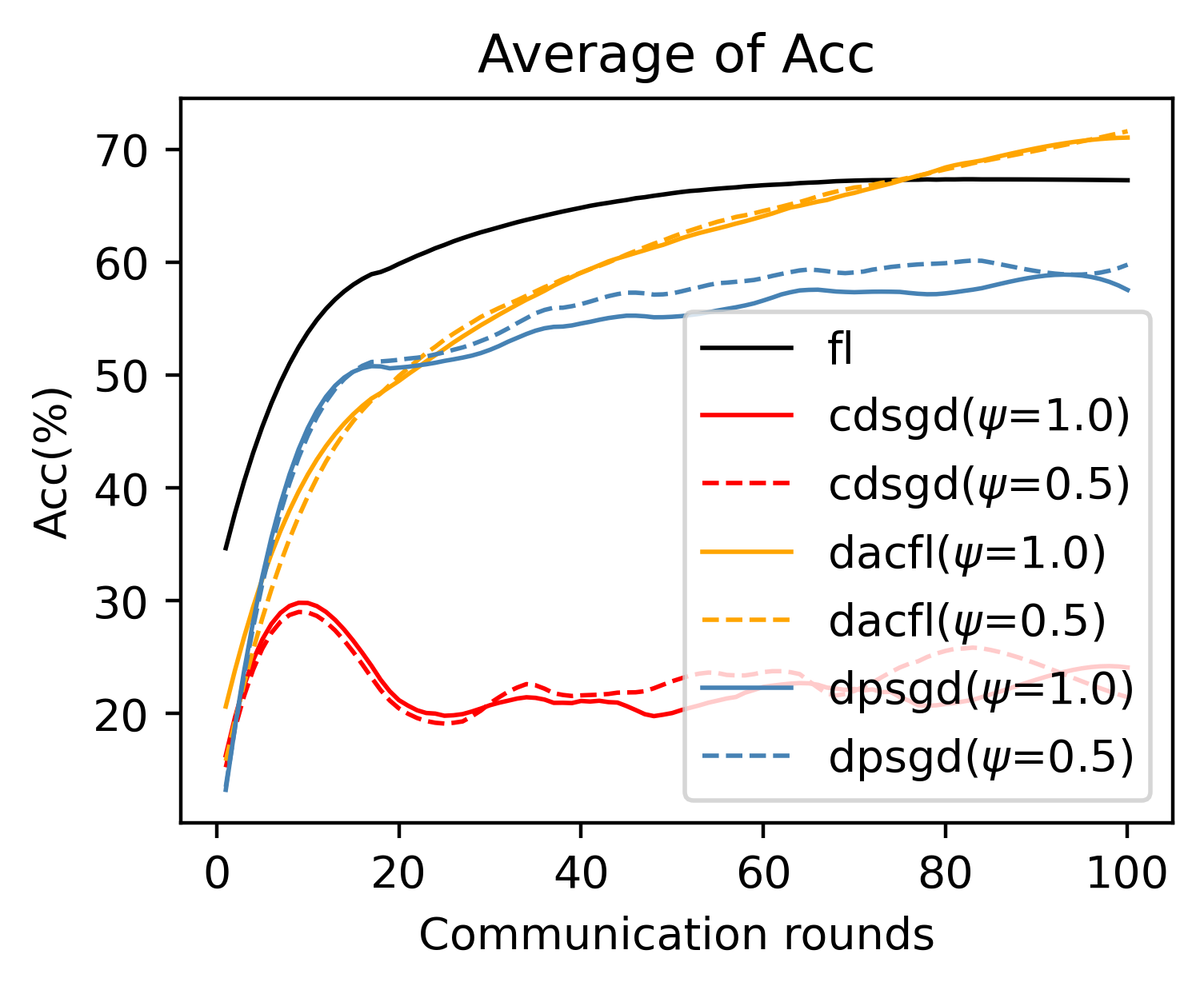}
		\end{minipage}
	}
	\subfigure[Var of Acc on MNIST]{
		\label{d iid4}
		\begin{minipage}[b]{0.31\textwidth}
			\includegraphics[width=1\textwidth]{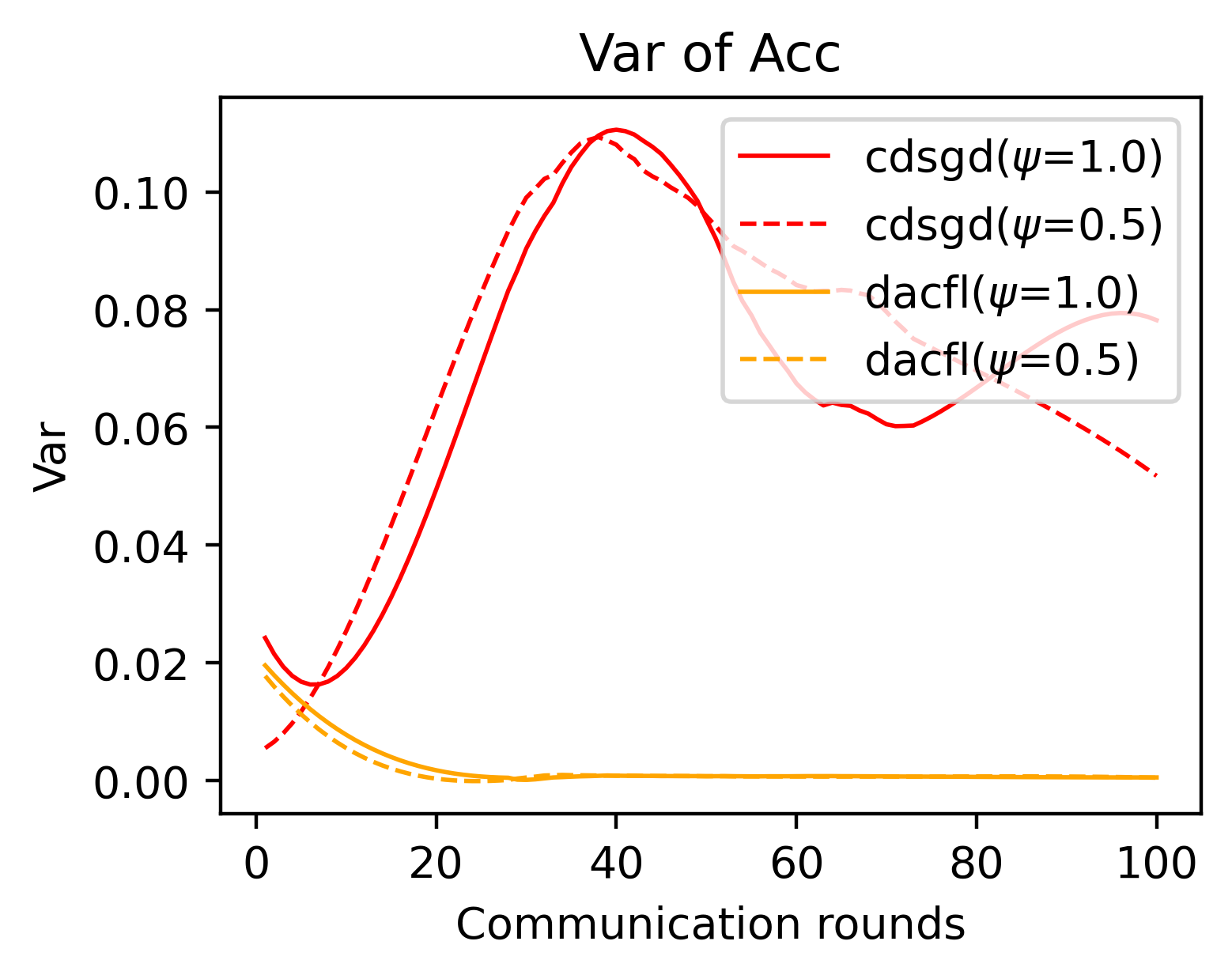}
		\end{minipage}
	}
	\subfigure[Var of Acc on FMNIST]{
		\label{d iid5}
		\begin{minipage}[b]{0.31\textwidth}
			\includegraphics[width=1\textwidth]{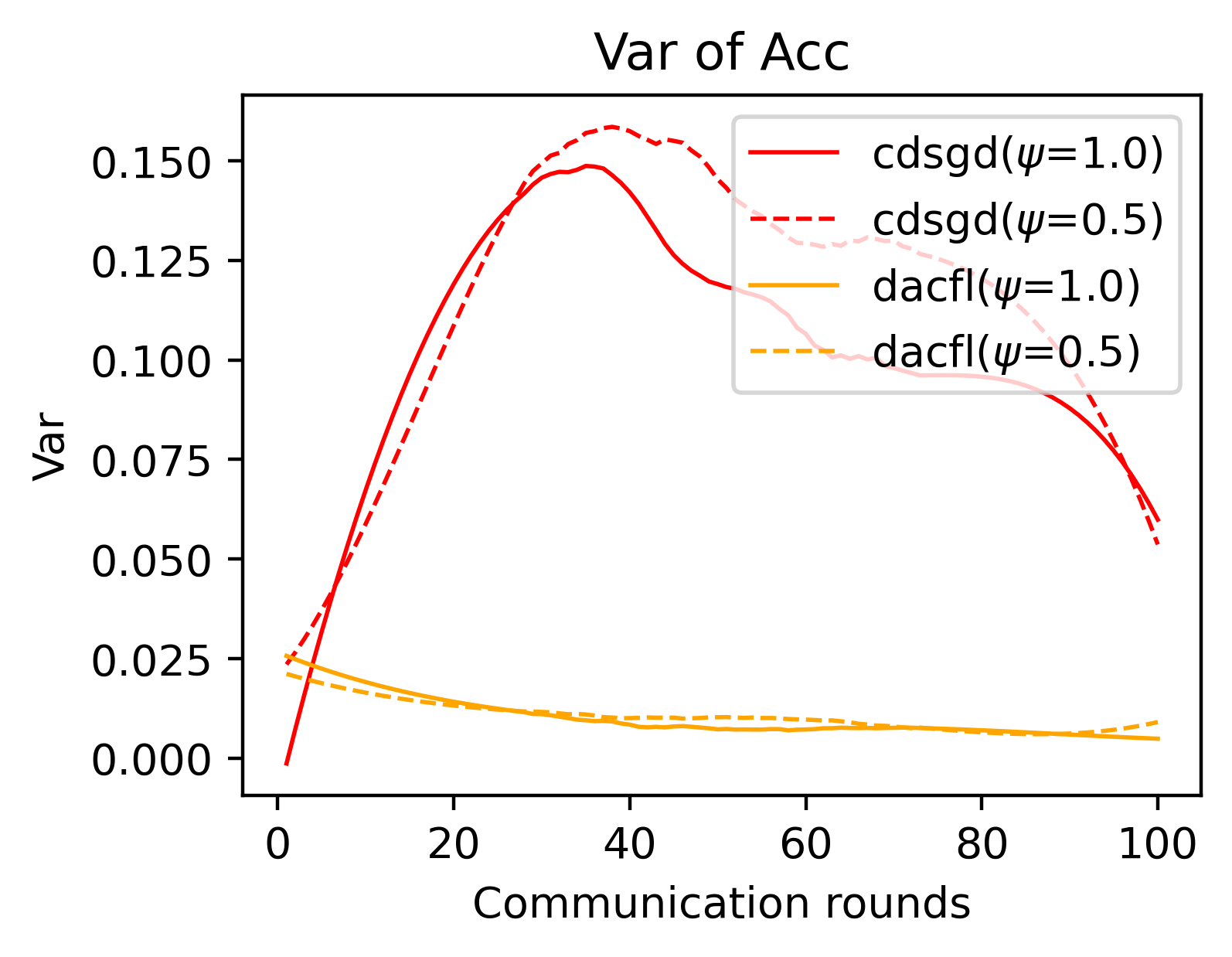}
		\end{minipage}
	}
	\subfigure[Var of Acc on CIFAR-10]{
		\label{d iid6}
		\begin{minipage}[b]{0.31\textwidth}
			\includegraphics[width=1\textwidth]{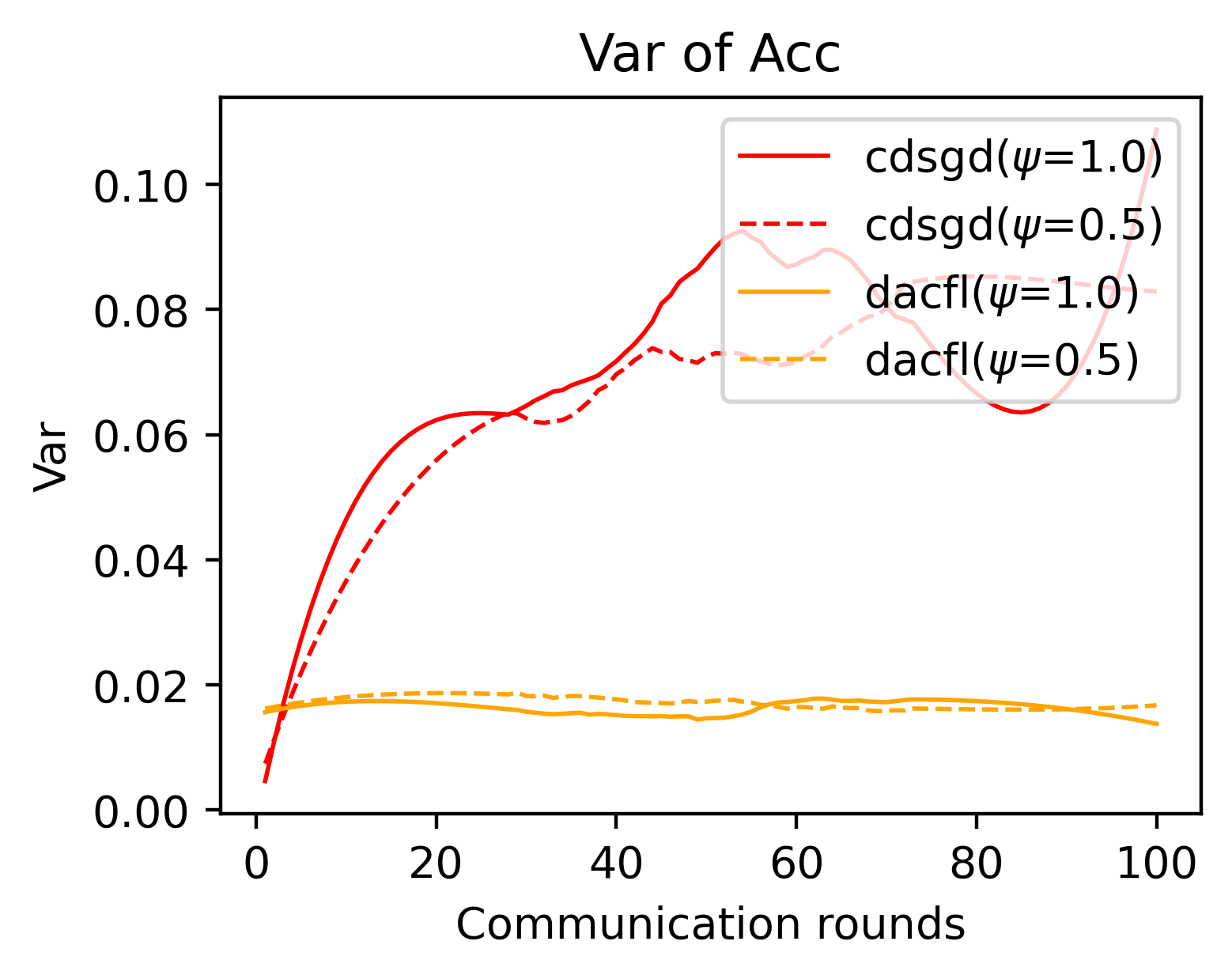}
		\end{minipage}
	}
	\caption{Performance comparison with i.i.d data and time-varying topology} 	
	\label{fig:dynamic i.i.d}
\end{figure*}

Fig. \ref{fig:static i.i.d} and \ref{fig:dynamic i.i.d} respectively shows the performances of different algorithms with i.i.d data allocation under time-invariant and time-varying topologies. 
The hyper-parameters in this experiment follow Table \ref{para-set}.

\subsubsection{Time-invariant Topology}
Fig. \ref{fig:static i.i.d} shows the result with i.i.d data and time-invariant  topology.
From the following three perspectives, we can draw different conclusions.

First, the proposed DACFL outperforms D-PSGD and CDSGD in terms of \textit{Average of Acc}, albeit slightly inferior to the conventional FedAvg. 
Specifically, from fig. \ref{s iid1}, \ref{s iid2} and \ref{s iid3}, the DACFL finally harvests $97\%$, $86\%$, $70\%$ accuracy under a dense topology and $96\%$, $85\%$, $67\%$ accuracy under a sparse topology, respectively on MNIST, FMNIST and CIFAR-10 datasets. 
This is superior to the result of CDSGD with $93\%$, $64\%$, $18\%$ accuracy under a dense topology and $68\%$, $51\%$, $20\%$ accuracy under a sparse topology, and the result of D-PSGD with $97\%$, $83\%$, $55\%$ accuracy under a dense topology and $95\%$, $75\%$, $45\%$ accuracy under a sparse topology.
Since D-PSGD additionally performs a network-wide model average over all users, it holds a better \textit{Average of Acc} than CDSGD. 
Contrarily, there exists different levels of deviation between user's intermediate training models, which may severely rely on the unevenness of the decentralized communication network topology.
By means of FODAC, each user in DACFL is able to well approximate the ``near average'' model. 
This is exactly why DACFL outperforms CDSGD and D-PSGD in this experiment. 

Second, the DACFL is less sensitive to the sparsity of communication topology. That is, the DACFL has minimal degradation in the \textit{Average of Acc} when a sparse topology arises.
Specifically, the DACFL has $1\%$, $1\%$, $3\%$ of accuracy reduction on three datasets.
While the D-PSGD has reduction of $2\%$, $8\%$, $10\%$ on three datasets and the CDSGD has reduction of $25\%$, $13\%$ on MNIST and FMNIST.\footnote{Since CDSGD dose not converge on CIFAR-10 after 100 rounds, it is not counted here.}
This result indicates that both CDSGD and D-PSGD ask for tighter topology requirements than DACFL for convergence guarantees.

Third, each user has closer performance in DACFL than that in CDSGD. 
From fig. \ref{s iid4}, \ref{s iid5} and \ref{s iid6}, it can be seen that, the variance of accuracy of DACFL is smaller and more stable when comparing to CDSGD, and gradually tends to around $0$ as the training progresses, especially on MNIST.
This result also supports that the average model can be well tracked by DACFL through the FODAC consensus method. 

In summary, the DACFL approach outperforms D-PSGD and CDSGD on i.i.d data under time-invariant topology. 

\subsubsection{Time-varying Topology}
In this section, we investigate how the time-varying topology affects the performance of DACFL and compare it with CDSGD and D-PSGD.
Fig. \ref{fig:dynamic i.i.d} presents the result on i.i.d data under time-varying topology.

From the perspective of \textit{Average of Acc}, the DACFL still outperforms D-PSGD and CDSGD. 
Take the result on FMNIST dataset as an example (fig. \ref{d iid2}), the DACFL finally reaches $87\%$ accuracy, which is better than the $84\%$ accuracy of D-PSGD and $68\%$ of CDSGD.
Intuitively, the FedAvg with a centralized topology performs better than other decentralized implementations as it has a central parameter server to do the \textit{global aggregation} phase. This would not be affected by the varying decentralized topology considered in this section. 

Besides, a time-varying topology has greater randomness than a time-invariant topology.
Due to the randomness, the accuracy degradation caused by the topology sparsity becomes smaller for all decentralized implementations considered in this section.
Specially, for D-PSGD on FMNIST (fig. \ref{d iid2}) and CIFAR-10 (fig. \ref{d iid3}), the average accuracy under a sparse topology is even greater than that under a dense topology.
This might because that the randomness introduced by time-varying topology reduces the possibility of early over-fitting that may be caused by a sparse topology.

Finally, for the \textit{Var of Acc} shown by Fig. \ref{d iid4}, \ref{d iid5} and \ref{d iid6}, result similar to that of a static topology arises. 
As is shown in the figures, the DACFL holds smaller and more stable variance of accuracy than CDSGD on both dense and sparse topology over all the datasets. 
This confirms the effectiveness of our approach again.

In summary, the DACFL still has higher feasibility in the case of time-varying topology.

\subsection{Performance on non-i.i.d Data}
\label{sec:noniid}
\begin{figure*}[htbp]
	
	\centering
	\subfigure[Average of Acc on MNIST]{
		\label{d noniid1}
		\begin{minipage}[b]{0.31\textwidth}
			\includegraphics[width=1\textwidth]{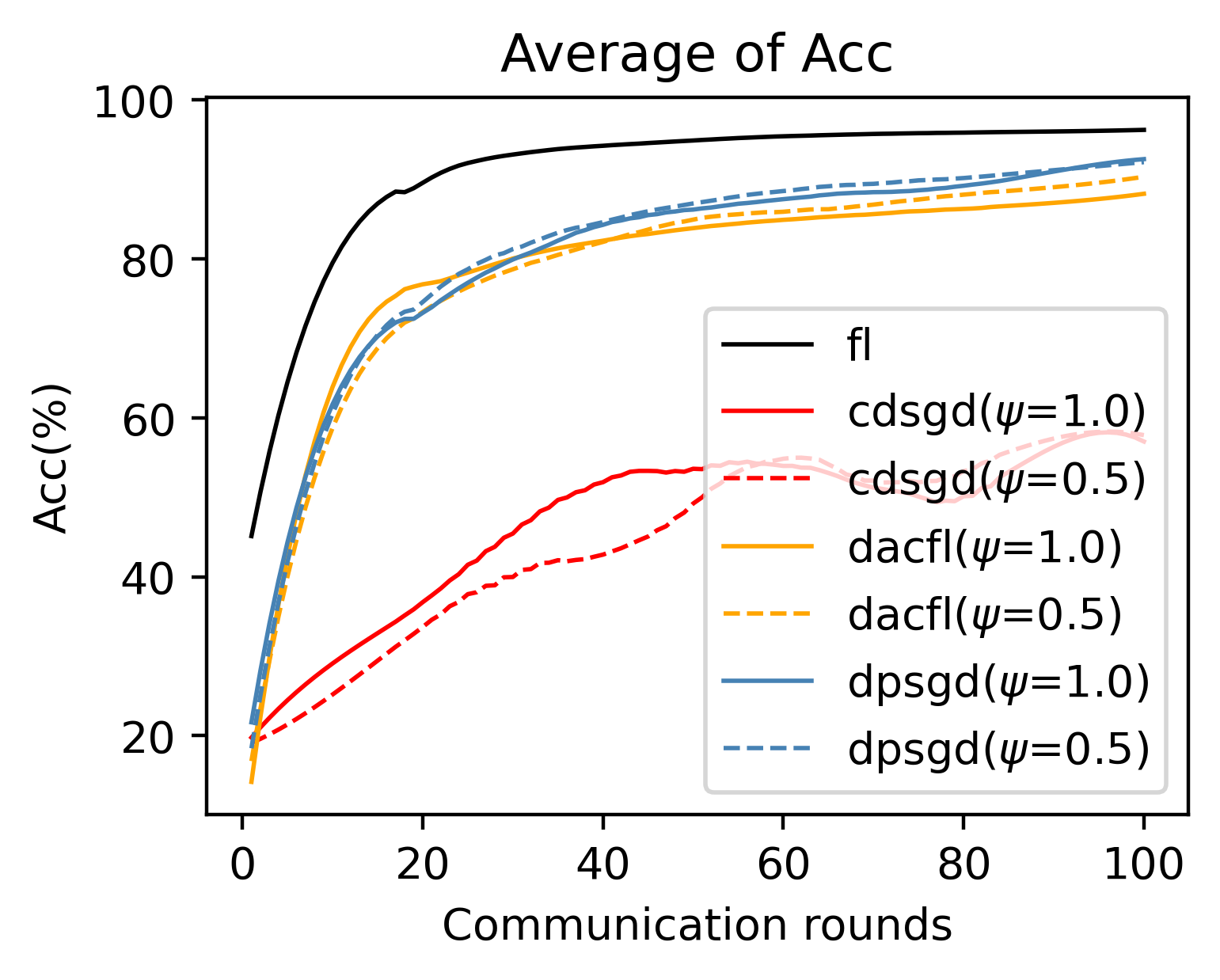}
		\end{minipage}
	}
	\subfigure[Average of Acc on FMNIST]{
		\label{d noniid2}
		\begin{minipage}[b]{0.31\textwidth}
			\includegraphics[width=1\textwidth]{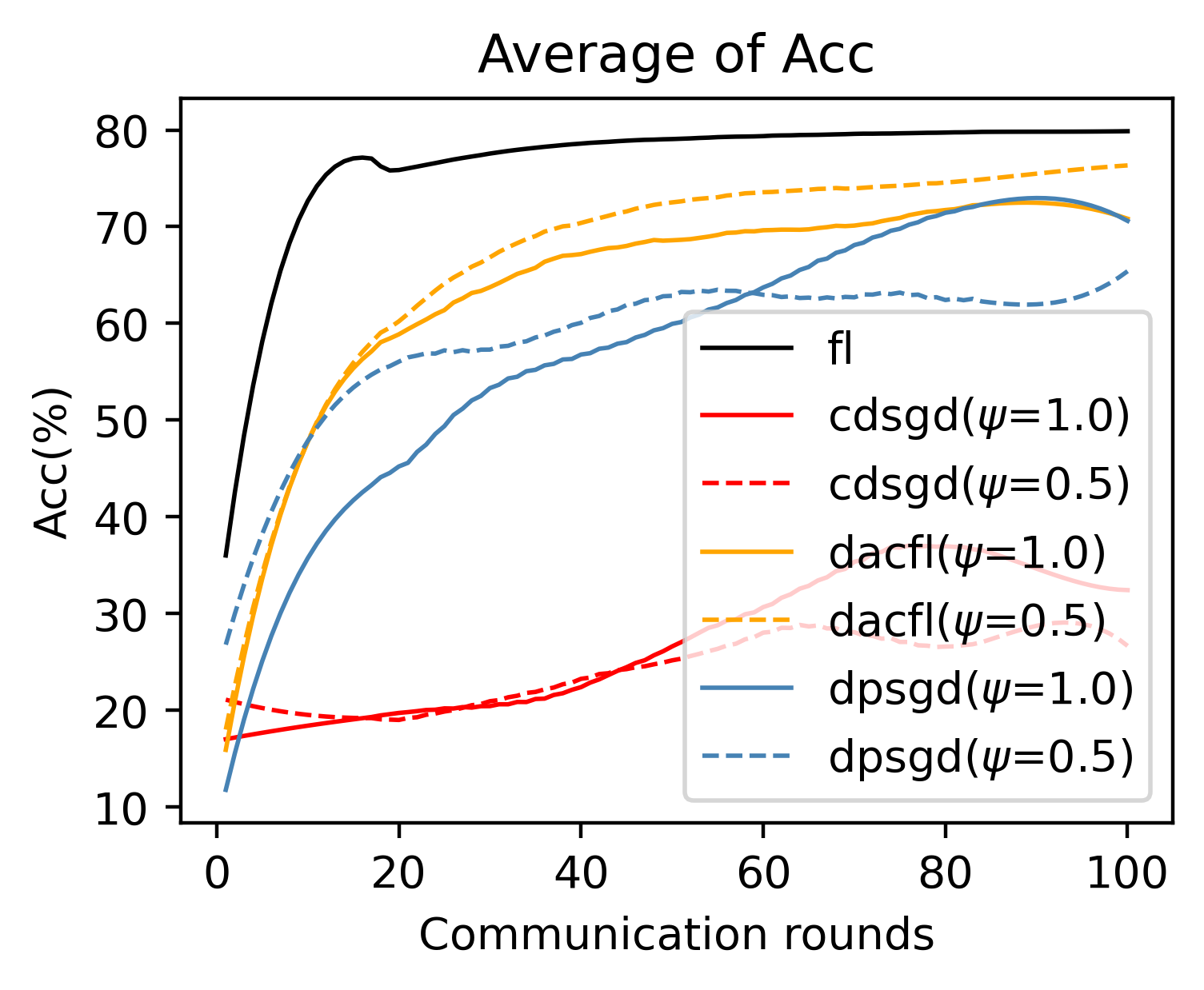}
		\end{minipage}
	}
	\subfigure[Average of Acc on CIFAR-10]{
		\label{d noniid3}
		\begin{minipage}[b]{0.31\textwidth}
			\includegraphics[width=1\textwidth]{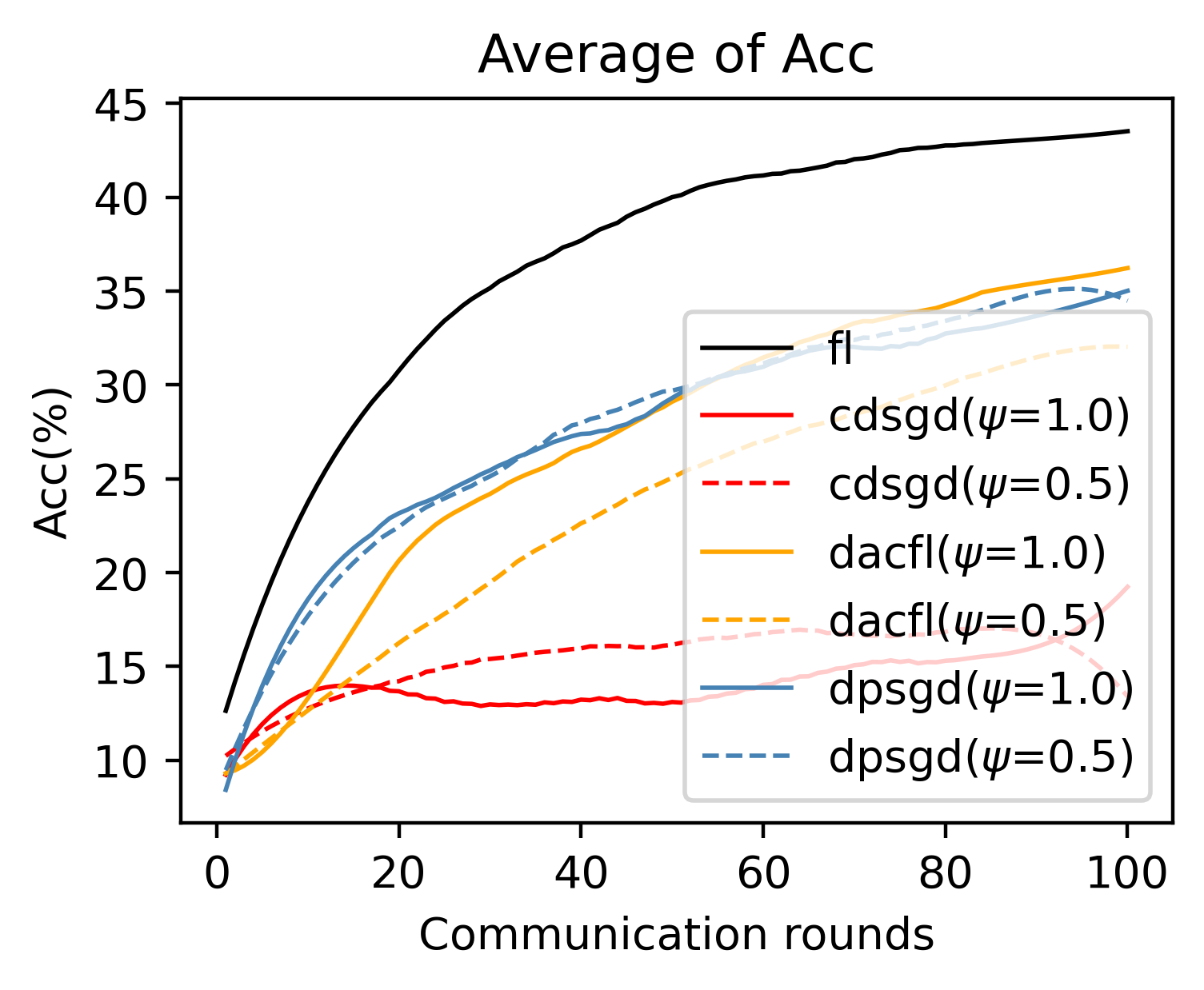}
		\end{minipage}
	}
	\subfigure[Var of Acc on MNIST]{
		\label{d noniid4}
		\begin{minipage}[b]{0.31\textwidth}
			\includegraphics[width=1\textwidth]{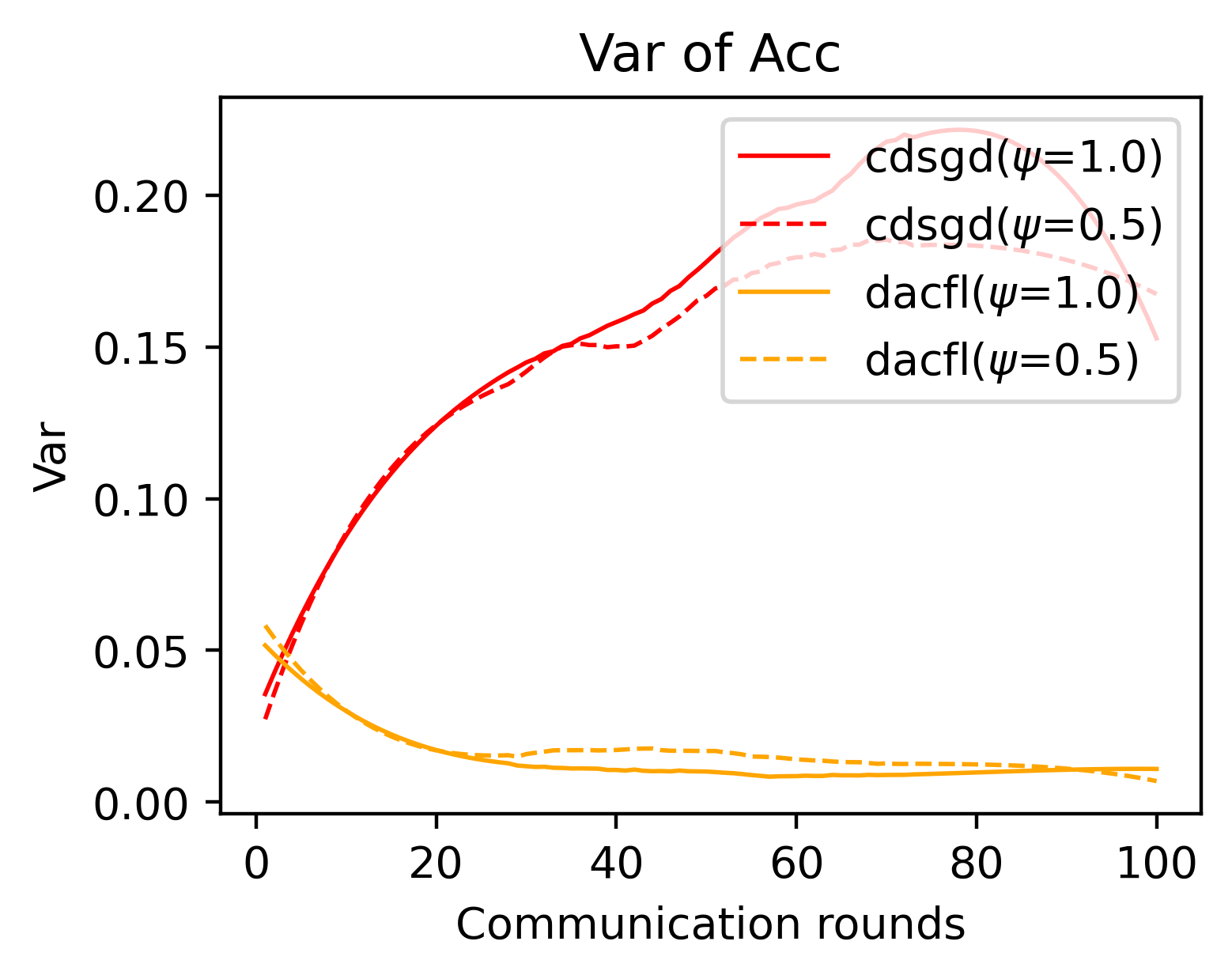}
		\end{minipage}
	}
	\subfigure[Var of Acc on FMNIST]{
		\label{d noniid5}
		\begin{minipage}[b]{0.31\textwidth}
			\includegraphics[width=1\textwidth]{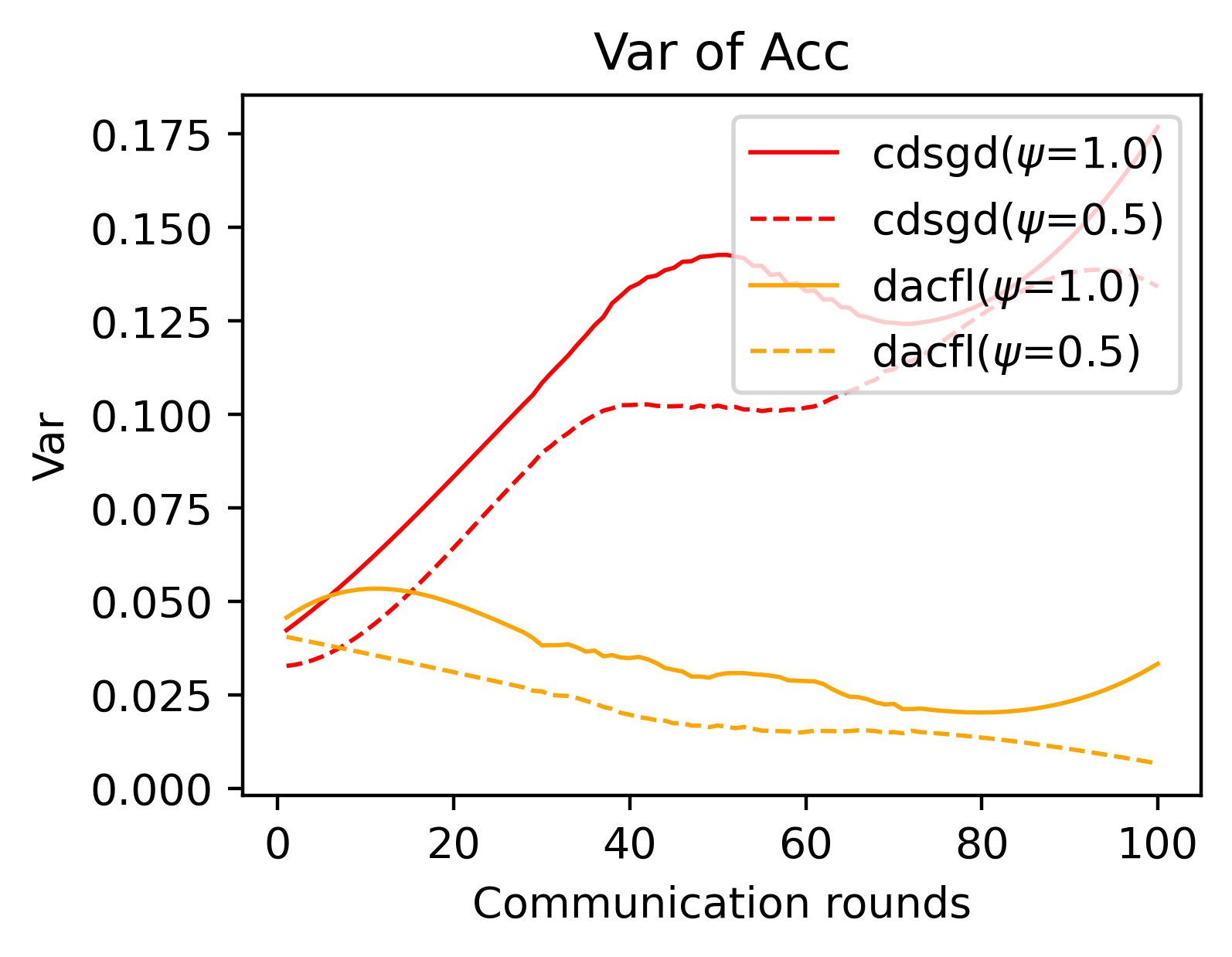}
		\end{minipage}
	}
	\subfigure[Var of Acc on CIFAR-10]{
		\label{d noniid6}
		\begin{minipage}[b]{0.31\textwidth}
			\includegraphics[width=1\textwidth]{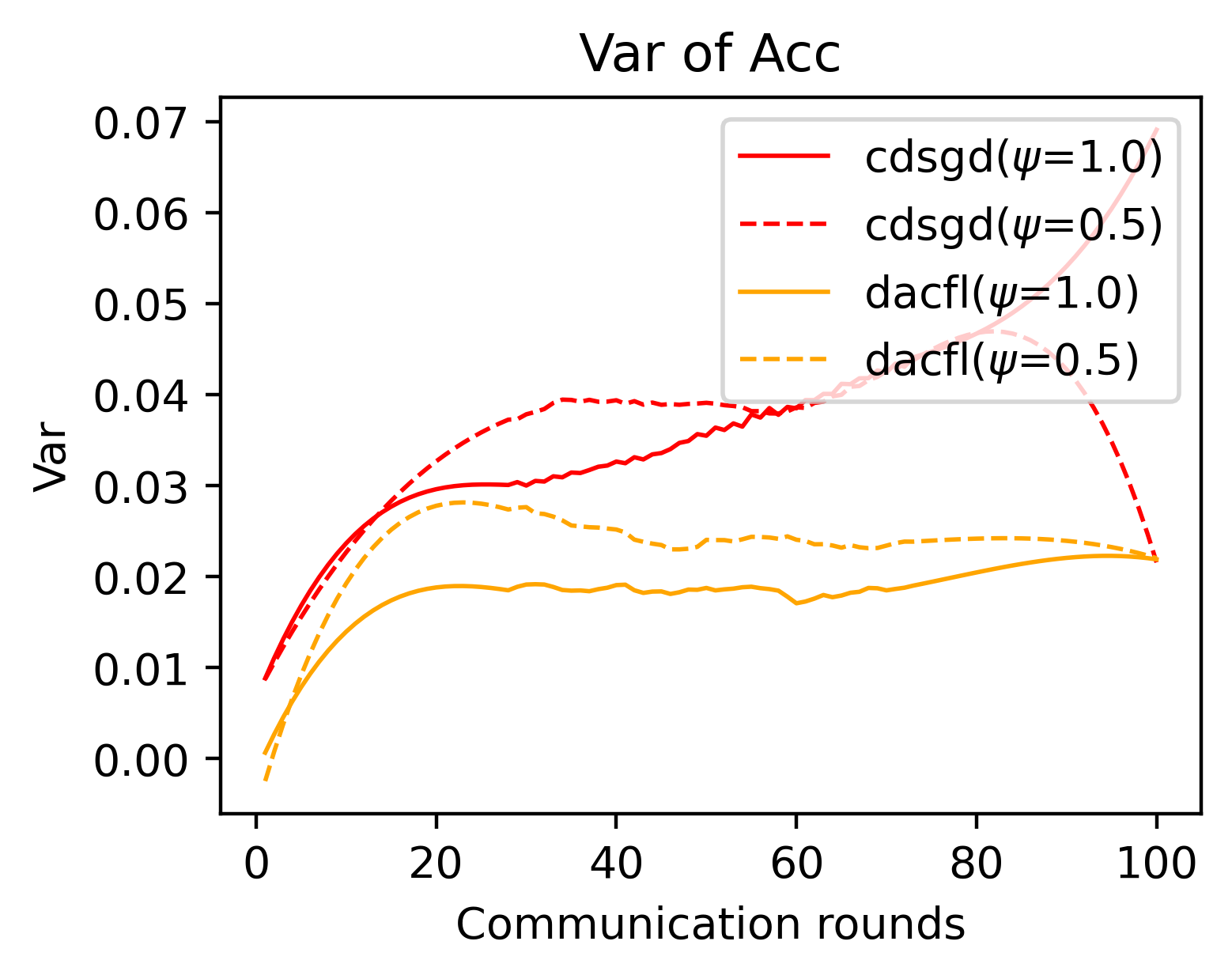}
		\end{minipage}
		
	}
	\caption{Performance comparison with non-i.i.d data and time-varying topology} 	
	\label{fig:dynamic non-i.i.d}
\end{figure*}

The section \ref{sec:iid} has shown the result of DACFL on i.i.d data and declared its practicality under both time-invariant and time-varying topology.
In this section, we test the performance of DACFL on non-i.i.d data and show the result in Fig. \ref{fig:static non-i.i.d} and Fig. \ref{fig:dynamic non-i.i.d}.
The hyper-parameters in this experiment follow Table \ref{para-set}.

\subsubsection{Time-invariant Topology}
\begin{figure*}[ht]
	
	\centering
	\subfigure[Average of Acc on MNIST]{
		\label{s noniid1}
		\begin{minipage}[b]{0.31\textwidth}
			\includegraphics[width=1\textwidth]{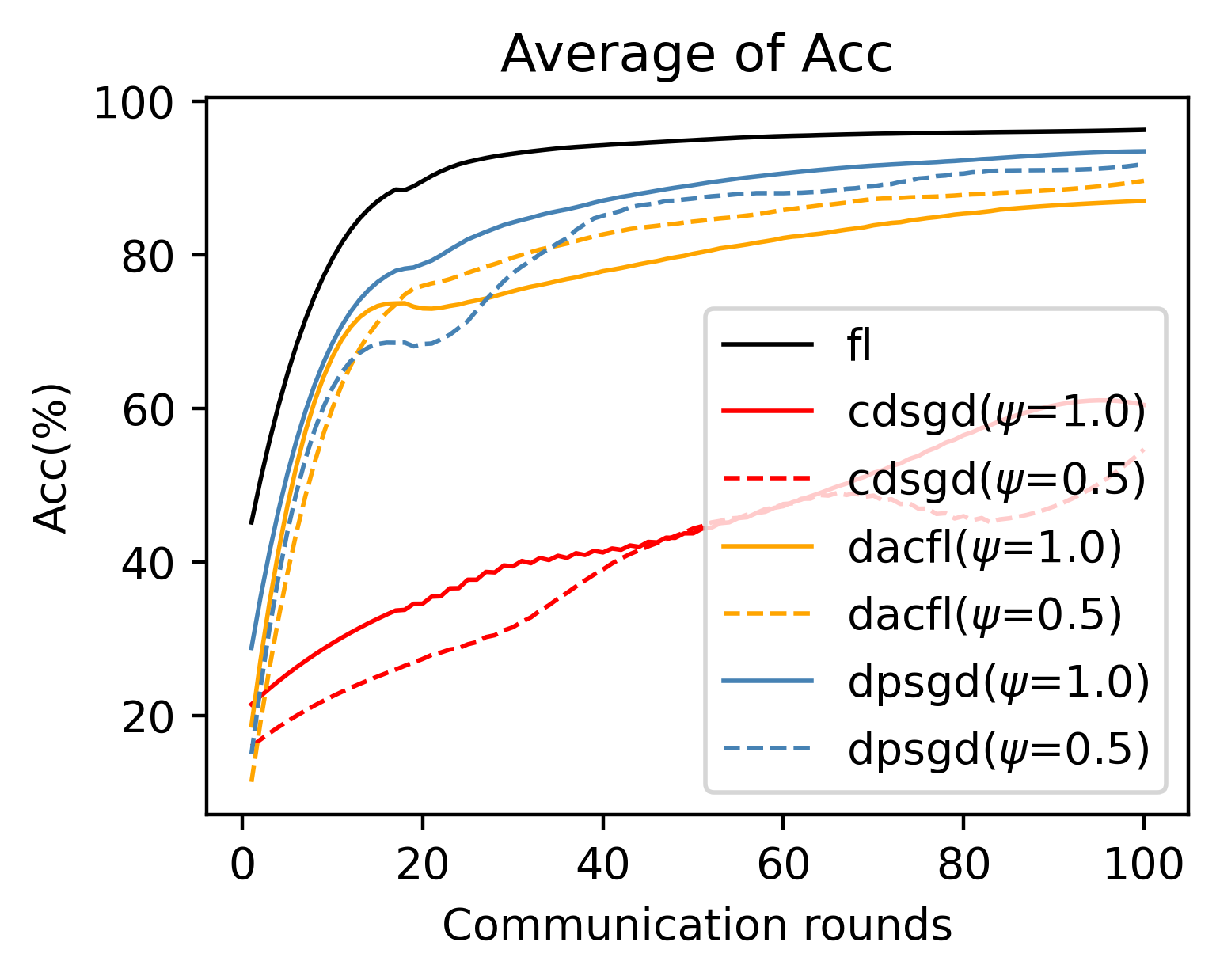}
		\end{minipage}
	}
	\subfigure[Average of Acc on FMNIST]{
		\label{s noniid2}
		\begin{minipage}[b]{0.31\textwidth}
			\includegraphics[width=1\textwidth]{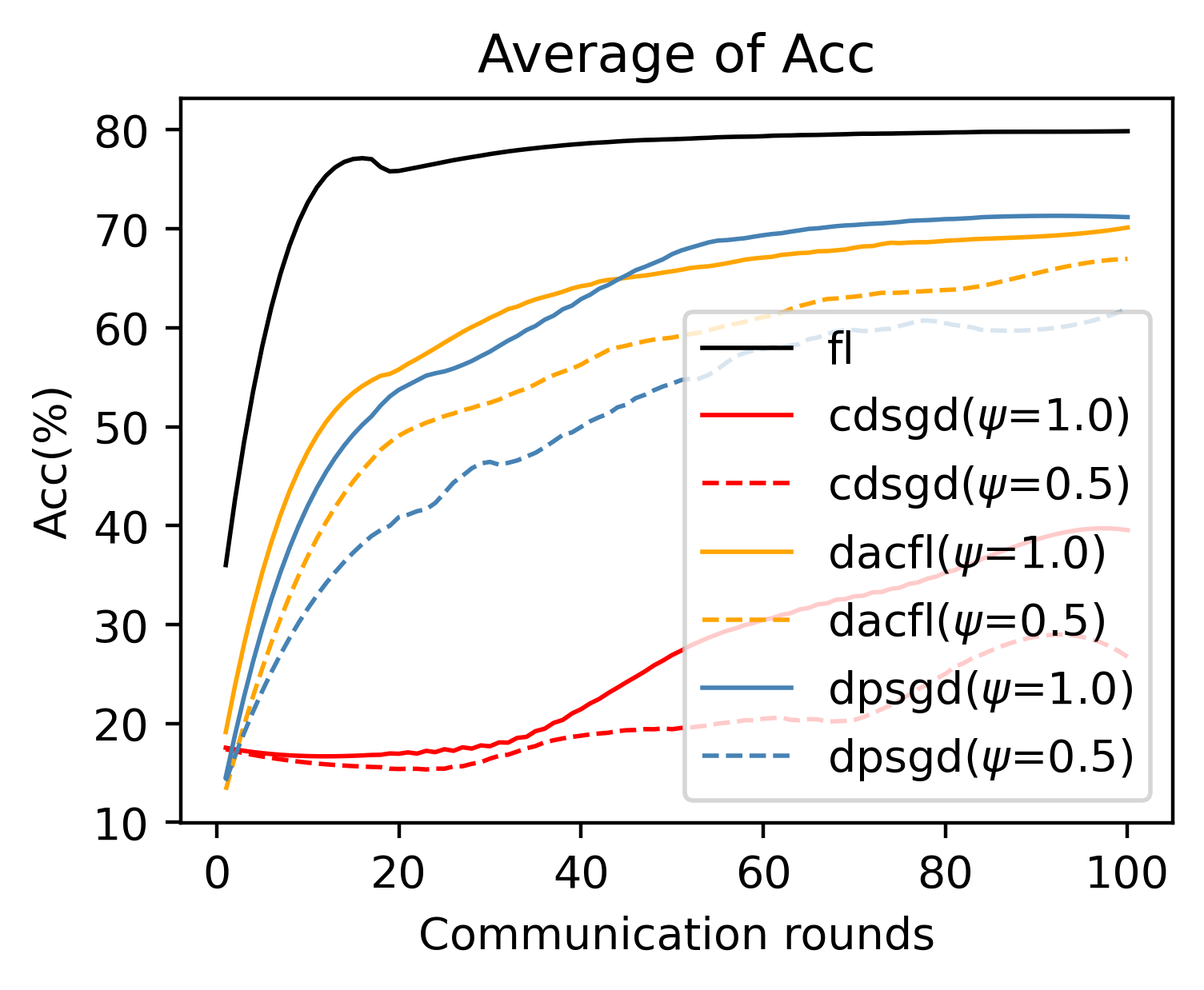}
		\end{minipage}
	}
	\subfigure[Average of Acc on CIFAR-10]{
		\label{s noniid3}
		\begin{minipage}[b]{0.31\textwidth}
			\includegraphics[width=1\textwidth]{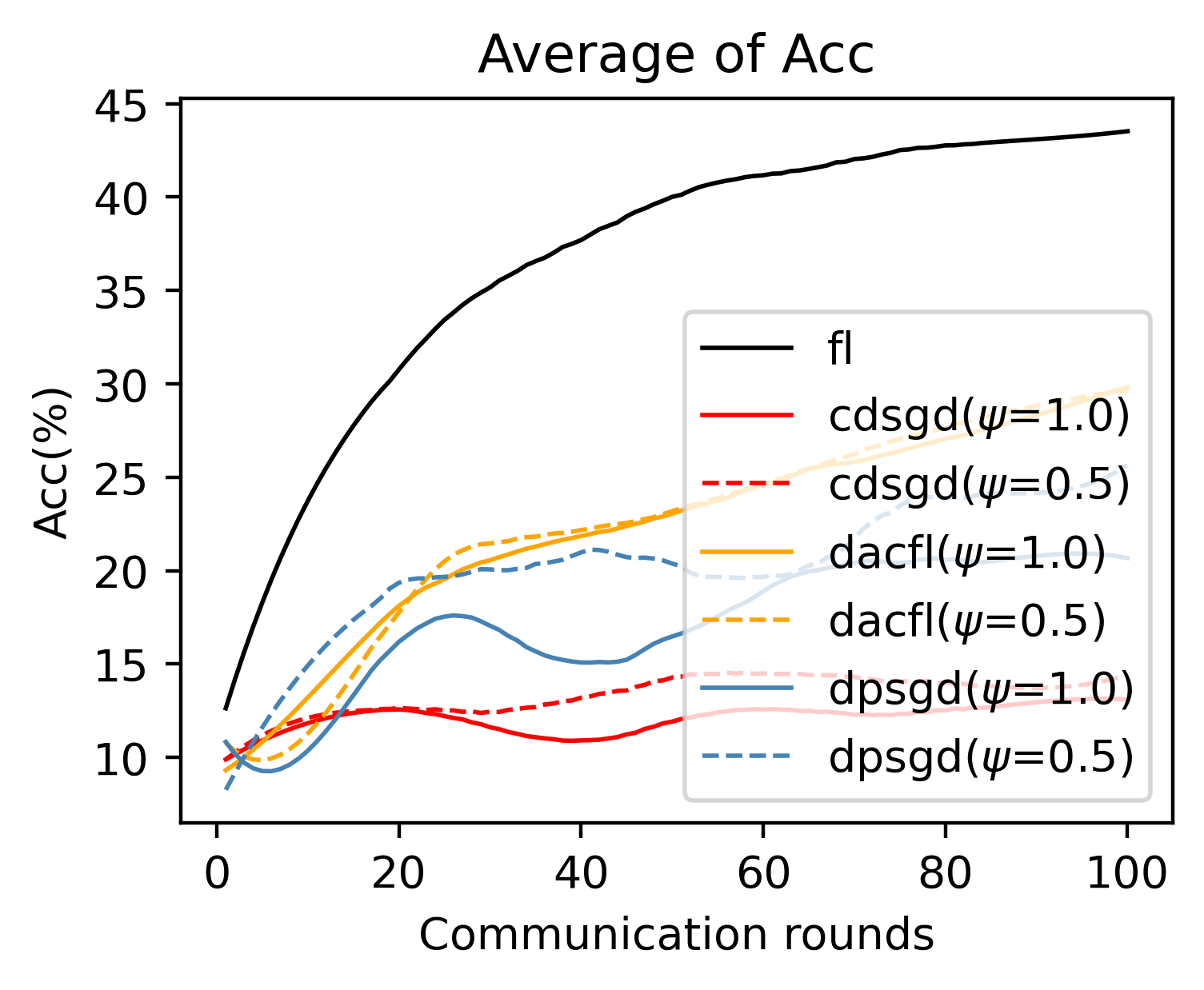}
		\end{minipage}
	}
	\subfigure[Var of Acc on MNIST]{
		\label{s noniid4}
		\begin{minipage}[b]{0.31\textwidth}
			\includegraphics[width=1\textwidth]{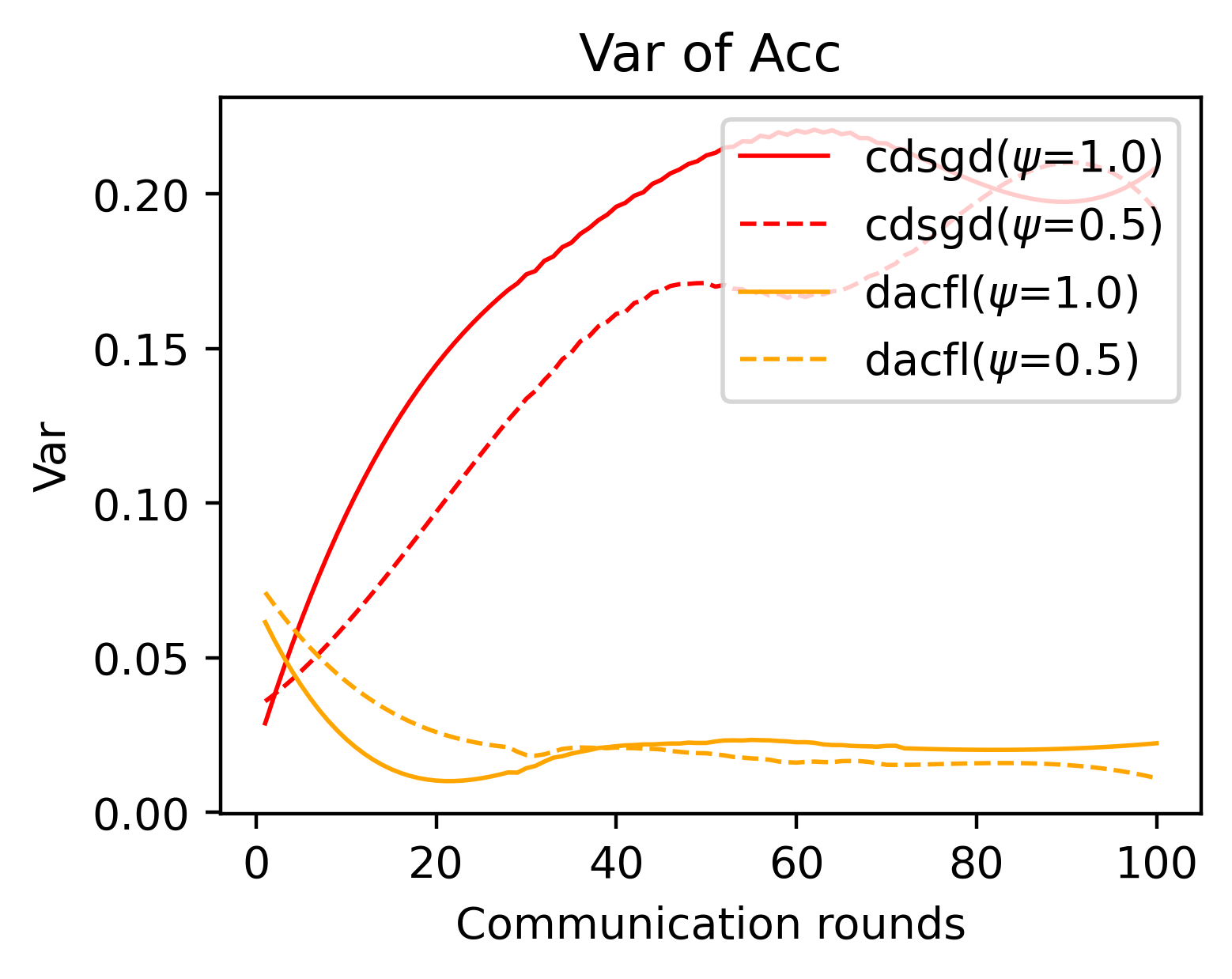}
		\end{minipage}
	}
	\subfigure[Var of Acc on FMNIST]{
		\label{s noniid5}
		\begin{minipage}[b]{0.31\textwidth}
			\includegraphics[width=1\textwidth]{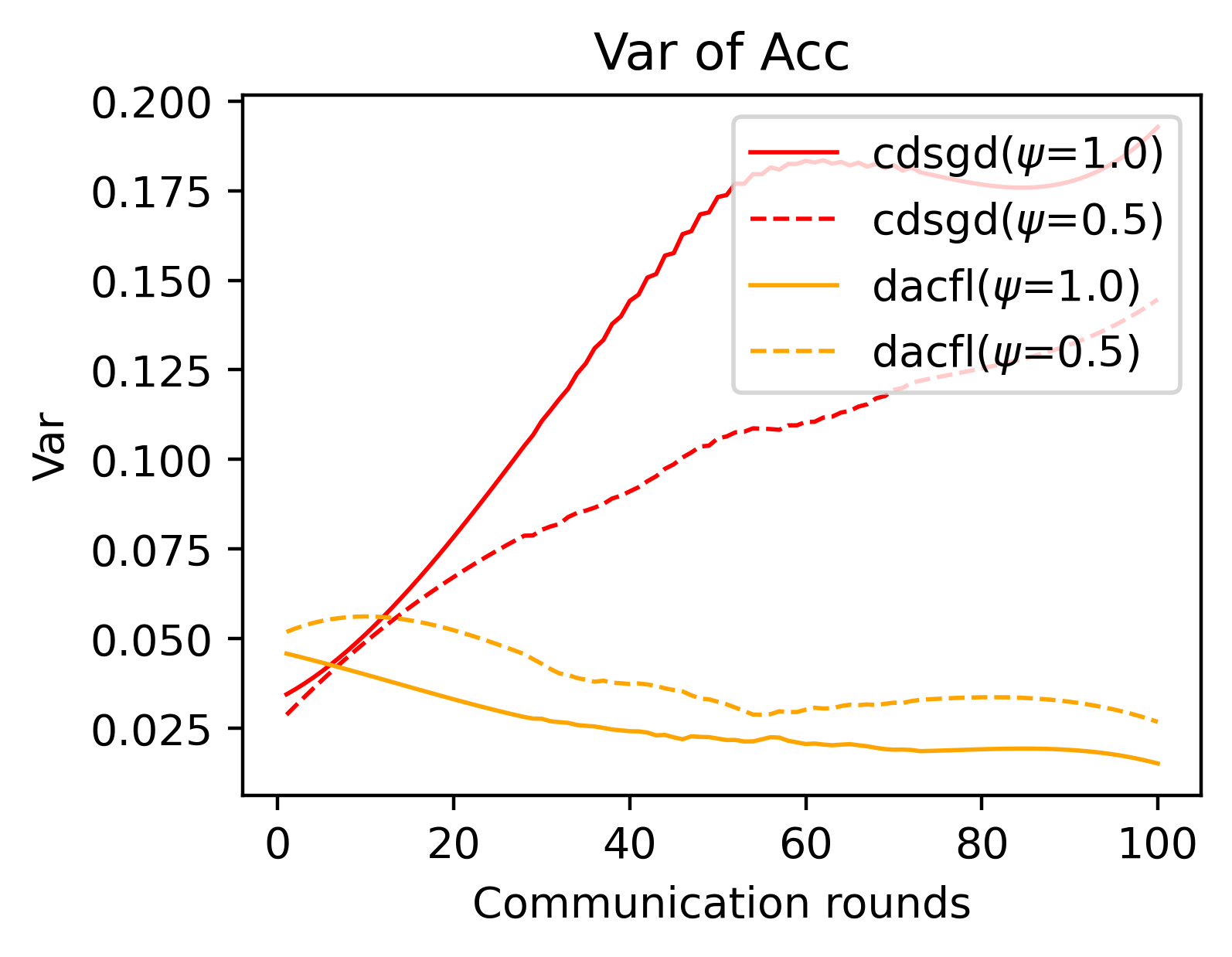}
		\end{minipage}
	}
	\subfigure[Var of Acc on CIFAR-10]{
		\label{s noniid6}
		\begin{minipage}[b]{0.31\textwidth}
			\includegraphics[width=1\textwidth]{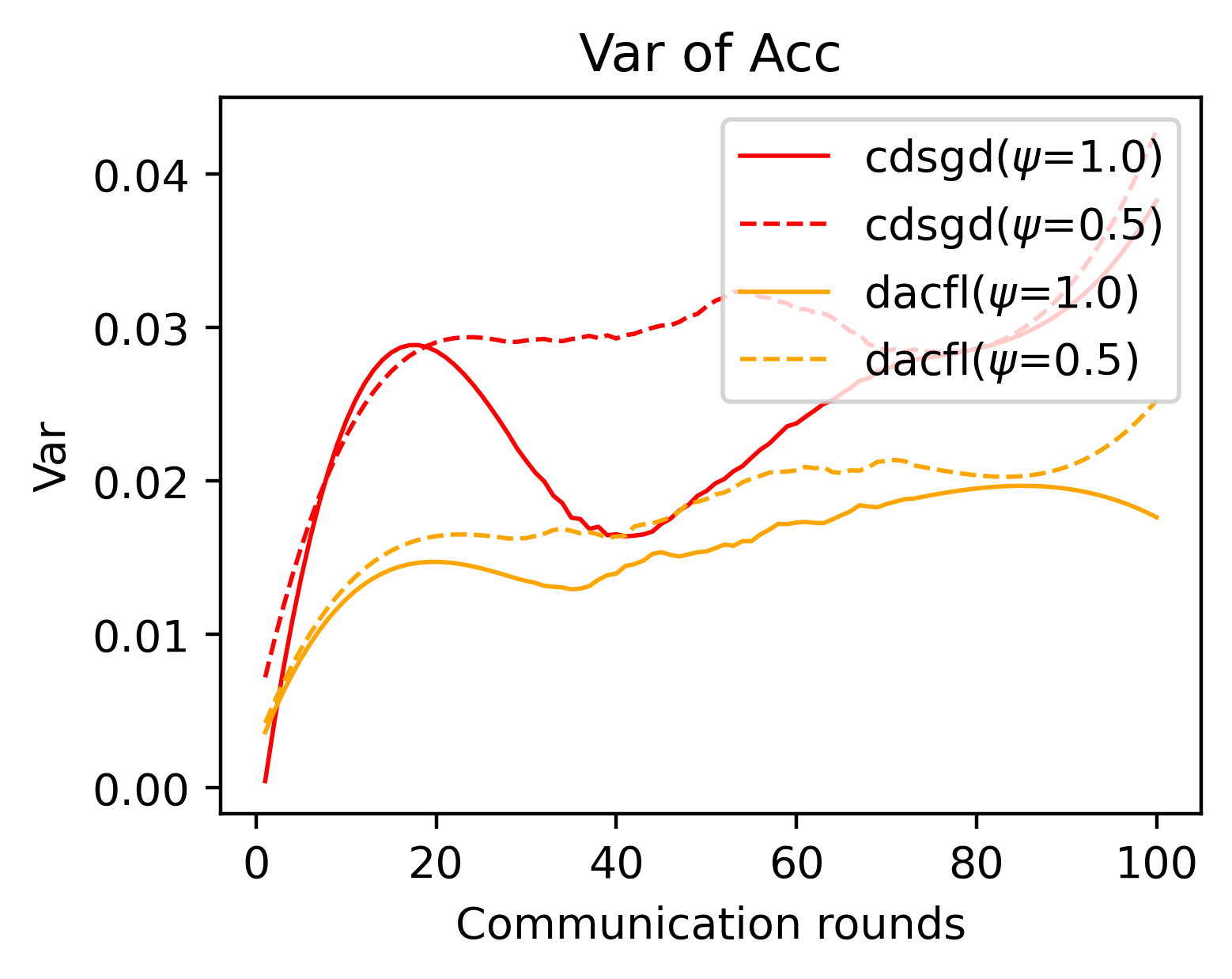}
		\end{minipage}
	}
	\caption{Performance comparison with non-i.i.d data and time-invariant topology} 	
	\label{fig:static non-i.i.d}
\end{figure*}

Fig. \ref{fig:static non-i.i.d} presents the experimental result on non-i.i.d data under time-invariant topology.

First, for the \textit{Average of Acc}, both FedAvg and three decentralized federated learning implementations have different levels of accuracy degradation on three datasets. 
This is because the non-i.i.d property would lead to users' local model divergence and early over-fitting.
Since D-PSGD additionally performs a net-work wide model average, it has higher accuracy than DACFL and CDSGD in MNIST and FMNIST.\footnote{Since all three DFL approaches do not converge in CIFAR-10, the result on CIFAR-10 is not counted here.} 
However, we should note that a network-wide model average usually becomes impractical when users are very scattered in a very large decentralized topology since an acceptable overhead would be caused by such a network-wide communication.
Therefore, in case that there is no network-wide model average, our DACFL outperforms CDSGD. 
Take the result under dense topology as an example, DACFL gets average accuracy of $86\%$, $70\%$, while CDSGD only reaches $58\%$, $40\%$ on MNIST and FMNIST, respectively.

Second, the numerical result of \textit{Var of Acc} in fig. \ref{s noniid4}, \ref{s noniid5} and \ref{s noniid6} also shows the superiority of DACFL compared to CDSGD. 
Actually, the accuracy variance is larger than that of i.i.d data due to the non-i.i.d property. 
However, we can still see that the accuracy variance of DACFL gradually decreases and stabilizes as the training progresses.

In summary, the DACFL is also feasible on non-i.i.d data (MNIST and FMNIST) under time-invariant topology.


\subsubsection{Time-varying Topology}
Fig. \ref{fig:dynamic non-i.i.d} presents the experimental result on non-i.i.d data under time-varying topology.

It can be seen from fig. \ref{d noniid1}, \ref{d noniid2} and \ref{d noniid3} that the overall average accuracy degradation due to non-i.i.d data still exists when compared with the result on i.i.d data. 
Because the randomness introduced by time-variant topology may alleviate the local model over-fitting caused by non-i.i.d data, our DACFL has a better performance than that of a time-invariant topology shown in Fig. \ref{fig:static non-i.i.d}, especially on FMNIST and CIFAR-10.
For the \textit{Var of Acc}, similar result to Fig. \ref{fig:static non-i.i.d} happens, i.e., the variance of DACFL gradually decreases and tends to zero, which confirms the viability of DACFL on tracking an ``average model".

To sum up, the DACFL is also viable on non-i.i.d data (MNIST and FMNIST) under time-varying topology.

\subsection{Convergence Speed vs Learning Rate and Topology Size}
To figure out how the learning rate and network topology size affect our solution, we also log the average test accuracy and average training loss on i.i.d MNIST with different learning rates and topology sizes. Fig. \ref{fig:n} shows the numerical result. Note that there are no decaying on learning rate and all topologies are dense in this part of experiments.
Except for the learning rate and topology size, other hyper-parameters in this experiment follow Table \ref{para-set}.

\subsubsection{Performance vs Learning Rate $\lambda$}
Fig. \ref{mean_acc}, \ref{mean_loss} and \ref{var_acc} show the result of average test accuracy, average training loss and variance over test accuracy with different learning rates, respectively. 
From Fig.\ref{mean_acc} and \ref{mean_loss}, we can see that a larger $\lambda$ could bring the benefit to faster convergence within the range $0.001 \le \lambda \le 0.01$.
This is because a larger learning rate makes the loss function decreases with a larger step size, which leads to a faster convergence.
However, this situation changes when $0.05 \le \lambda \le 0.1$, i.e, when $\lambda$ increases from $0.05$ to $0.1$, the convergence speed and convergence result become even worse. Smaller average test accuracy and larger average training loss with greater surge precisely reflect this phenomenon. 
It is because that an excessive learning rate $\lambda$ would lead to a larger upper bound of first-order difference of model parameter $ \theta$ and thus cause a larger relatively upper bound of first-order difference $\kappa$ in \eqref{eq10}. 
Consequently, a larger relatively bound would lead to a larger steady-stater error when using FODAC to track the average of users' models \cite{zhu2010discrete}. Hence, an excessive learning rate would be unfriendly to our DACFL solution.
From fig.\ref{var_acc}, it can be seen that a greater learning rate leads to a greater variance of accuracy. 
So, $\lambda=0.01$ should be the best choice in this experiment, which gets a higher average test accuracy and lower variance while ensuring fast convergence.

\subsubsection{Performance vs Topology Size $N$}
\begin{figure*}[htbp]
	
	\centering
	\subfigure[Average of Acc]{
		\label{mean_acc}
		\begin{minipage}[b]{0.31\textwidth}
			\includegraphics[width=1\textwidth]{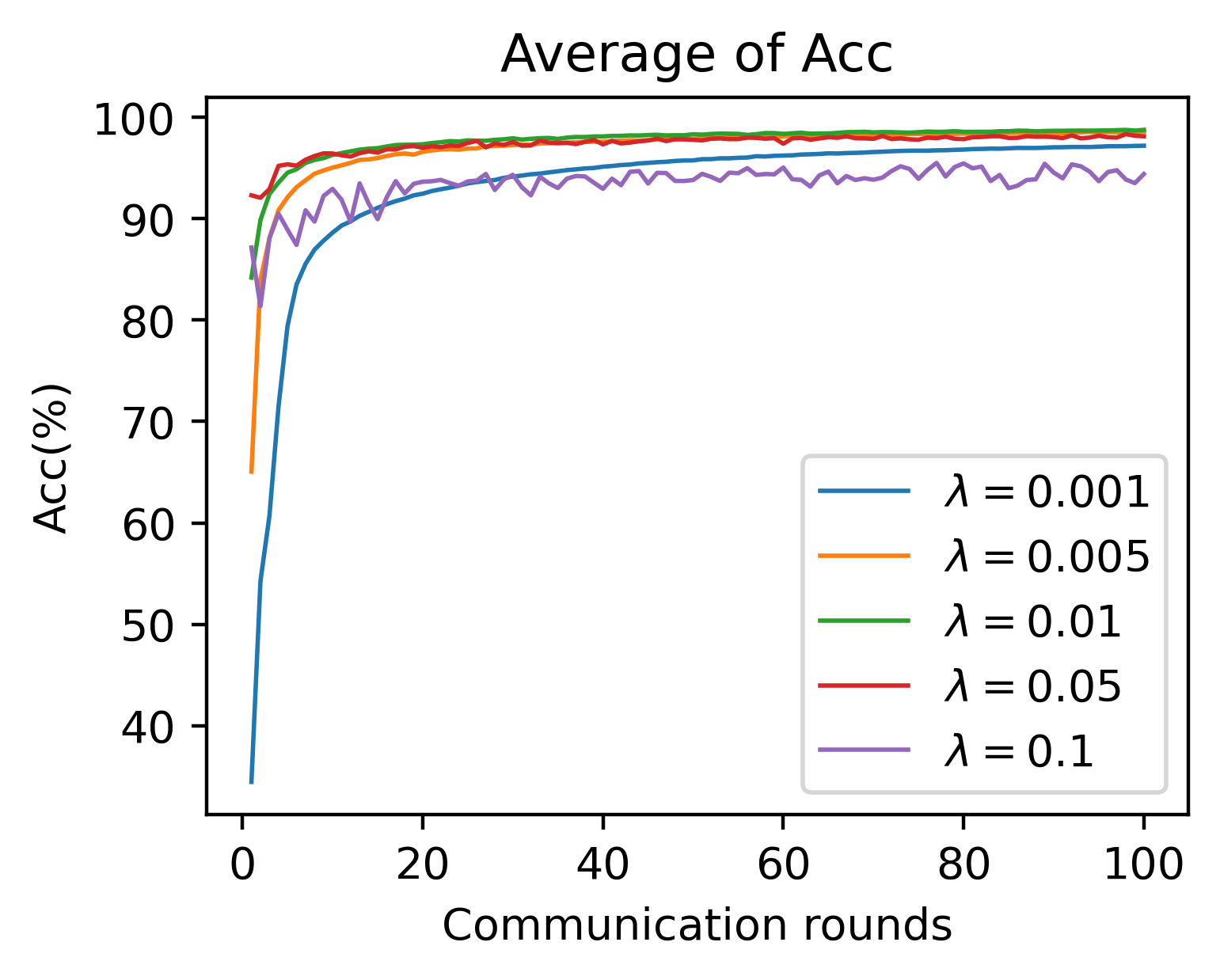}
		\end{minipage}
	}
	\subfigure[Average of Loss]{
		\label{mean_loss}
		\begin{minipage}[b]{0.31\textwidth}
			\includegraphics[width=1\textwidth]{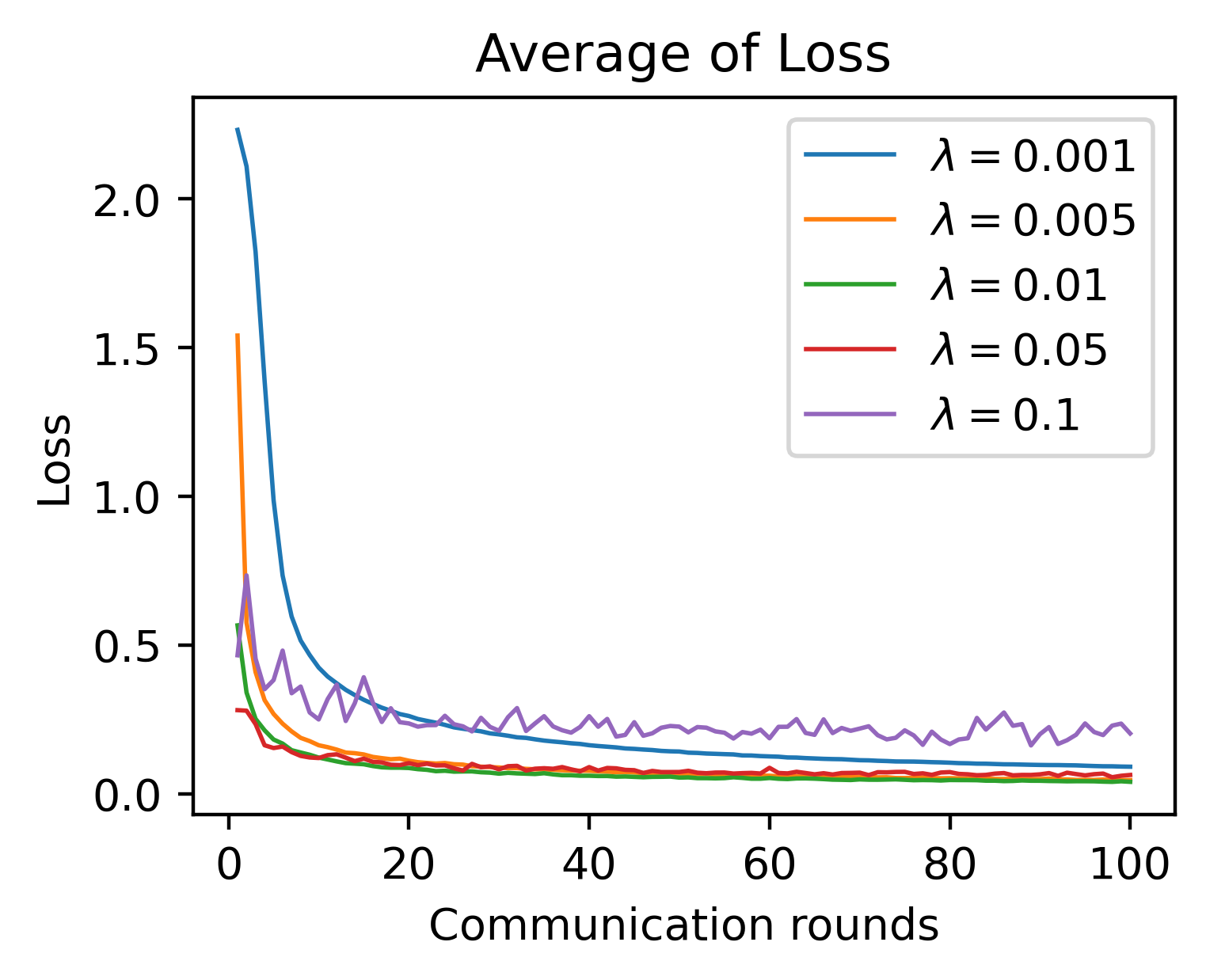}
		\end{minipage}
	}
	\subfigure[Var of Acc]{
		\label{var_acc}
		\begin{minipage}[b]{0.31\textwidth}
			\includegraphics[width=1\textwidth]{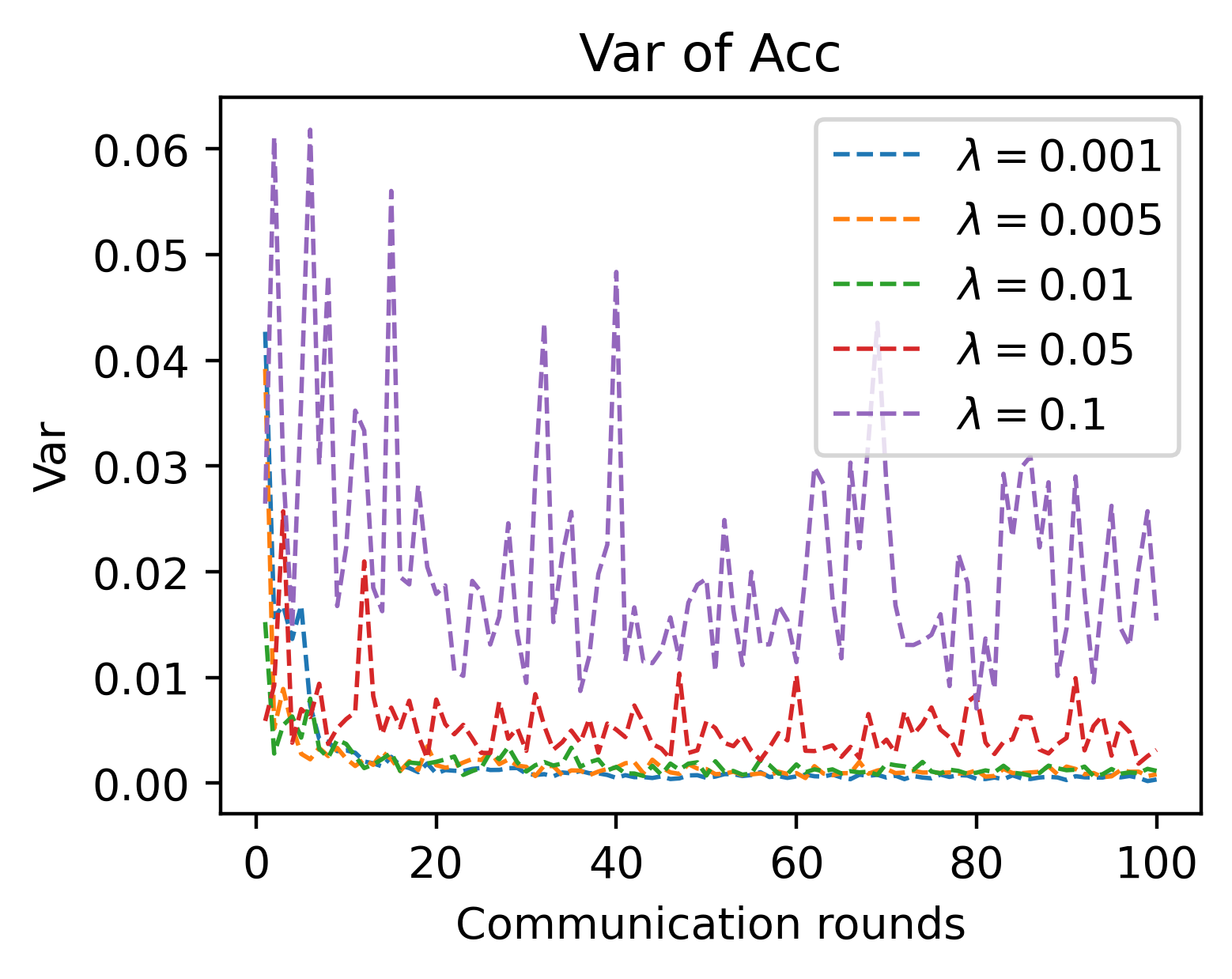}
		\end{minipage}
	}		
	\centering
	\subfigure[Average of Acc]{
		\label{n_mean_acc}
		\begin{minipage}[b]{0.31\textwidth}
			\includegraphics[width=1\textwidth]{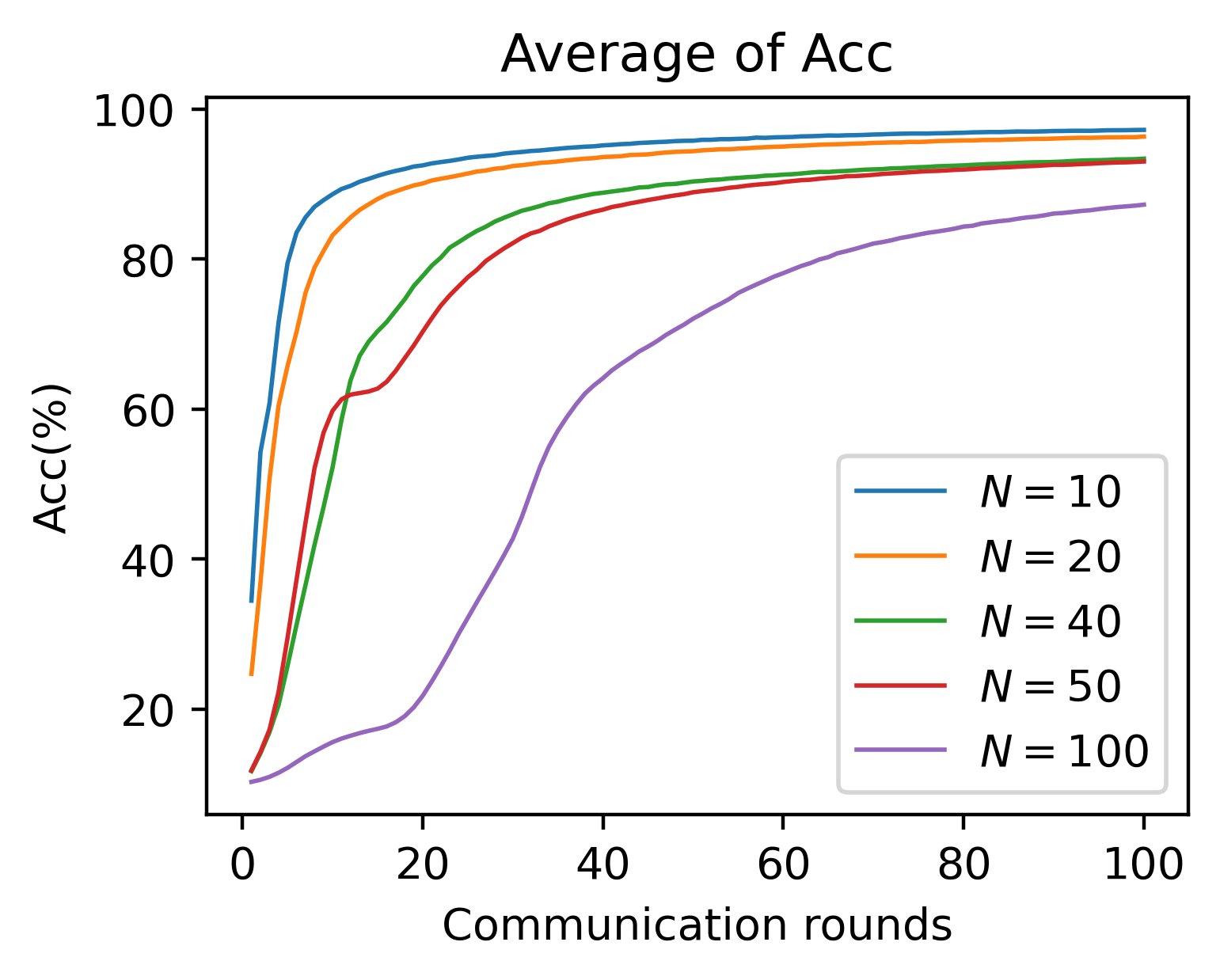}
		\end{minipage}
	}
	\subfigure[Average of Loss]{
		\label{n_mean_loss}
		\begin{minipage}[b]{0.31\textwidth}
			\includegraphics[width=1\textwidth]{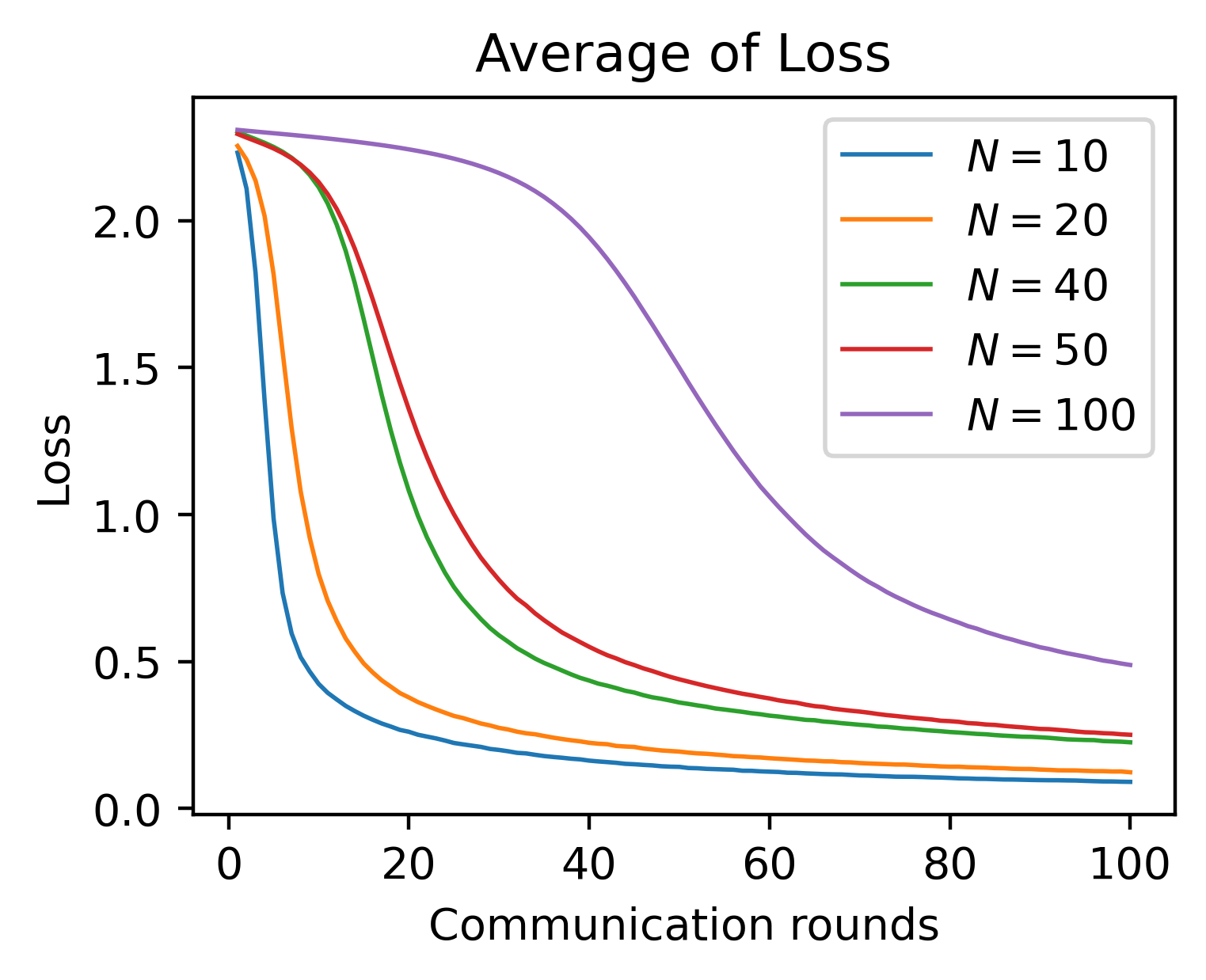}
		\end{minipage}
	}
	\subfigure[Var of Acc]{
		\label{n_var_acc}
		\begin{minipage}[b]{0.31\textwidth}
			\includegraphics[width=1\textwidth]{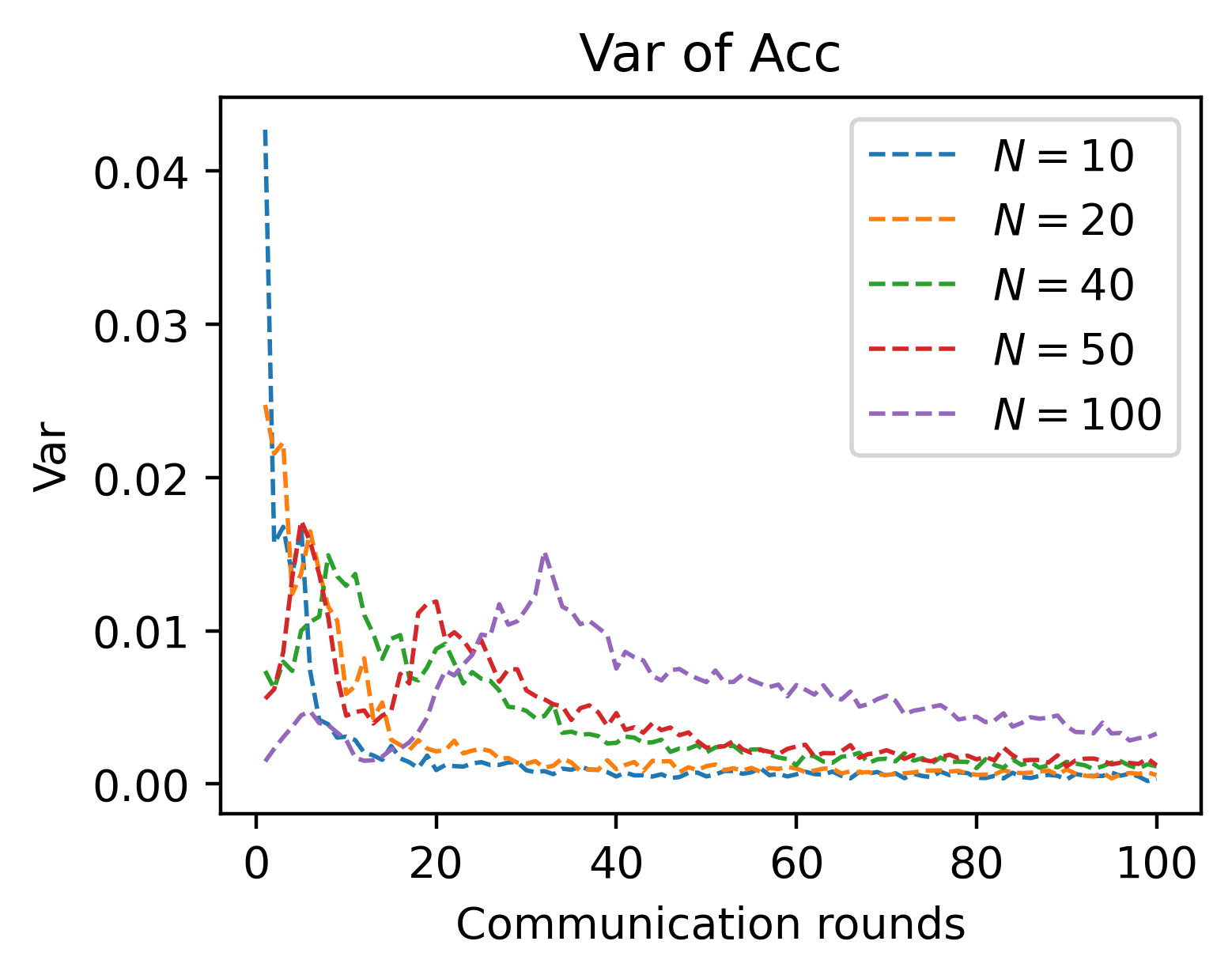}
		\end{minipage}
	}
	\caption{Performance with different learning rate and topology size} 	
	\label{fig:n}
\end{figure*}

Fig. \ref{n_mean_acc}, \ref{n_mean_loss} and \ref{n_var_acc} present the numerical result on different topology sizes. 
It is shown in fig. \ref{n_mean_acc} and \ref{n_mean_loss} that, as the size of topology $N$ grows, the convergence speed slows down. 
Also, the larger topology size it is, the lower final test accuracy it gets.
This is because that a larger size of topology would lead to a larger deviation among all users, which further cause a slower rate of our FODAC tracking for \textit{average model}.
Regarding the variance of accuracy, a larger topology usually results in a larger variance in the early training of our algorithm as is shown in fig. \ref{n_var_acc}. 
However, the accuracy variance on different topology gradually tend to $0$ after a certain number of iterations. 
This also declares the effectiveness of our solution by employing FODAC to track the \textit{average model}.

In a summary, a proper learning rate can accelerate the convergence of our DACFL training. 
Also, although DACFL is robust to different topology sizes, a smaller size is preferred to attain a better performance within a limit number of training rounds.

\section{Conclusions}
\label{sec7}

Over-reliance on the central PS makes the federated learning possibly paralyze when the server breaks down.
To alleviate this single point failure in conventional FL, existing researches have offered different DFL implementations including CDSGD and D-PSGD.
However, there exists significant variance between users' final models in CDSGD while D-PSGD necessitates a network-wide model average.
In this paper, we devise a new DFL method coined as DACFL to solve the deficiency in CDSGD and D-PSGD.
The DACFL treats the respective local training processes as discrete-time series and employ FODAC to track the \textit{average model} over all users.
To confirm the feasibility of DACFL, we also deliver a theoretical analysis on the premise of some assumptions, which offers a convergence guarantee of our solution.
Besides, we design specific experiments on MNIST, Fashion-MNIST and CIFAR-10 under i.i.d and non-i.i.d allocations, and compare the DACFL with D-PSGD and CDSGD.
The results verify the effectiveness of DACFL under different network topologies, and declare that DACFL outperforms D-PSGD and CDSGD in most cases.

There are several issues need further investigation.
First, DACFL solves the problems of CDSGD and D-PSGD at the expense of more communication overhead because each user exchanges both estimation states and local models during the training progress. Therefore, a more communication-efficient method for DACFL is deserved.
Second, since this work only considers DACFL with synchronized settings, an asynchronized decentralized federated learning deserves investigation for practical application.
Third, because DACFL works effectively only when mixing matrix is symmetric and doubly stochastic, users dropping out or joining in during the training process will change the nature of mixing matrix and thus yields negative effects on this method. Thus, designing an offline and join aware DACFL would be worthy of future research.

\appendices
\label{appendix}
\section{Proof of Theorem}
\subsection{Preliminaries}
In this section, we give an upper bound on the expected average squared gradient norms, which serves as a metric to measure the convergence rate for the non-convex objectives.

Before the detailed proof, here are some notations avoiding ambiguity. 
We denote the mixing matrix $\mathbf{W}=\left[\mathbf{w}_{ij}(t)\right] \in \mathbb{R}^{N \times N}$ a fully decentralized communication network topology with $N$ users, where $x_i^t, \omega_i^t$ represents for the estimation and local model of $i$-th user at round $t$, respectively.
We further have ${\bar \omega^t} = \frac{1}{N}\sum_{i = 1}^N {\omega_i^t} $ defined as the average model of all users at round $t$. 
Besides, we denote $\omega_i^{t'} = \sum_{j=1}^{t}\mathbf{w}_{ij}(t)\omega_j^t$ the neighborhood weighted average model, and use $\mathcal{g}_i^t = \nabla f_i\left(\omega_i^{t'} , \zeta_i^t\right)$ to denote the stochastic gradient of user $i$ at round $t$, where the $\zeta_i^t\subseteq \mathcal{D}_i$ is the uniformly sampled mini-batch from the $i$-th user's data shards at round $t$. With $\rm{T}$ denoting the transpose of a matrix, we also define the following notations 
\[{\boldsymbol{x}^t} = {\left[ {x_1^t, x_2^t, \cdots ,x_N^t} \right]^{\rm{T}}},\]
\[{\boldsymbol{\omega}^t} = {\left[ {\omega_1^t,\omega_2^t, \cdots ,\omega_N^t} \right]^{\rm{T}}},\]
to denote the sets of estimations and intermediate models at round $t$, respectively.
In the following proof, we consider a time-invariant topology where $\boldsymbol{\rm{W}}(t) = \boldsymbol{\rm{W}}$, such that $\mathbf{w}_{ij}(t) = \mathbf{w}_{ij}, t=0,1,\dots,T-1$.

\subsection{Proof of Theorem 1}
\begin{IEEEproof}
	According to the updating rules in \textbf{Algorithm 5} (line 4 to line 6), we have 
	\begin{align}
		\omega_i^{t'} &= \sum_{j=1}^{N}\mathbf{w}_{ij}(t)\omega_j^t\\
		\omega_i^{t+1} &= \omega_i^{t'} - \lambda \mathcal{g}_i^t
	\end{align}
	where $\mathbf{w}_{ij}(t)\in \mathbf{W}(t) $ denotes the $(i,j)$-th entry of the mixing matrix, thus
	\begin{equation}
		\label{eq15}
		\begin{split}
			\mathbb{E}\left[\left\| {\bar{\omega}^{t+1} - \bar{\omega}^t}\right\|^2\right]
			&=\mathbb{E}\left[{\frac{1}{N^2}}\left\|{\sum_{i=1}^{N}\left(\omega_i^{t+1}-\omega_i^t\right)}\right\|^2\right]\\
			& \le \mathbb{E}\left[\frac{1}{N}\sum_{i=1}^{N}\left\|\omega_i^{t+1} - \omega_i^{t}\right\|^2\right]\\
			&\le \theta^2
		\end{split}
	\end{equation}
	where the \eqref{eq15} follows from \textbf{Assumption 5}.
	
	Take the following recursive equations as examples,
	\begin{align*}
		\omega_i^{t+1} - \omega_i^t &= \sum_{j=1}^{N}\mathbf{w}_{ij}\omega_j^t - \lambda \mathcal{g}_i^t - \left(\sum_{j=1}^{N}\mathbf{w}_{ij}\omega_j^{t-1} - \lambda \mathcal{g}_i^{t-1}\right)\\
		&=\sum_{j=1}^{N}{\mathbf{w}_{ij}\left(\omega_j^t - \omega_j^{t-1}\right) - \lambda\left(\mathcal{g}_i^t-\mathcal{g}_i^{t-1}\right)}
	\end{align*}
	\begin{align*}
		&\omega_j^t - \omega_j^{t-1} =\sum_{i=1}^{N}\mathbf{w}_{ij}\left(\omega_i^{t-1}-\omega_i^{t-2}\right) - \lambda\left(\mathcal{g}_j^{t-1}-\mathcal{g}_j^{t-2}\right)\\
		&\omega_i^{t-1} - \omega_i^{t-2} = \sum_{j=1}^{N}\mathbf{w}_{ij}\left(\omega_j^{t-2} - \omega_j^{t-3}\right) - \lambda \left(\mathcal{g}_i^{t-2}-\mathcal{g}_i^{t-3}\right)\\
		&\omega_j^{t-2} - \omega_j^{t-3} = \sum_{i=1}^{N}\mathbf{w}_{ij}\left(\omega_i^{t-3} - \omega_i^{t-4}\right) - \lambda \left(\mathcal{g}_j^{t-3}-\mathcal{g}_j^{t-4}\right)\\
		&\vdots \\
		&\omega_i^{1} - \omega_i^{0} = \sum_{j=1}^{N}\mathbf{w}_{ij}\left(\omega_j^0 - \omega_j^{-1}\right) - \lambda\left(\mathcal{g}_i^0-\mathcal{g}_i^{-1}\right)
	\end{align*}
	
	Then we have
	\begin{align*}
		\omega_i^{t+1}-\omega_i^{t} &= \underbrace{\sum_{j=1}^{N}\mathbf{w}_{ij}\sum_{i=1}^{N}\mathbf{w}_{ij}\dots\sum_{j=1}^{N}\mathbf{w}_{ij}}_{\Sigma_i \,\text{or} \, \Sigma_j \,\text{total $t$ times}}\left(\omega_j^0-\omega_j^{-1}\right)\\
		& - \underbrace{\sum_{i=1}^{N}\mathbf{w}_{ij}\sum_{j=1}^{N}\mathbf{w}_{ij}\dots\sum_{j=1}^{N}\mathbf{w}_{ij}}_{\Sigma_i \,\text{or} \, \Sigma_j \,\text{total $t$-1 times}}\lambda\left(\mathcal{g}_i^0-\mathcal{g}_i^{-1}\right)\\
		&\vdots \\
		& - 
		\sum_{j=1}^{N}\mathbf{w}_{ij}\lambda\left(\mathcal{g}_j^{t-1}-\mathcal{g}_j^{t-2}\right) - \lambda\left(\mathcal{g}_i^t-\mathcal{g}_i^{t-1}\right)\\
		& \stackrel{(a)} = - \underbrace{\sum_{i=1}^{N}\mathbf{w}_{ij}\sum_{j=1}^{N}\mathbf{w}_{ij}\dots\sum_{j=1}^{N}\mathbf{w}_{ij}}_{\Sigma_i \,\text{or} \, \Sigma_j \,\text{total $t$-1 times}}\lambda\left(\mathcal{g}_i^0-\mathcal{g}_i^{-1}\right)\\
		& - \underbrace{\sum_{j=1}^{N}\mathbf{w}_{ij}\sum_{i=1}^{N}\mathbf{w}_{ij}\dots\sum_{j=1}^{N}\mathbf{w}_{ij}}_{\Sigma_i \,\text{or} \, \Sigma_j \,\text{total $t$-2 times}}\lambda\left(\mathcal{g}_i^1-\mathcal{g}_i^{0}\right)\\
		&\vdots \\
		& - 
		\sum_{j=1}^{N}\mathbf{w}_{ij}\lambda\left(\mathcal{g}_j^{t-1}-\mathcal{g}_j^{t-2}\right) - \lambda\left(\mathcal{g}_i^t-\mathcal{g}_i^{t-1}\right)\\
		& \stackrel{(b)} = - \lambda \left(\Delta \mathcal{g}_i^t + \Delta \mathcal{g}_j^{t-1} + \dots + \Delta \mathcal{g}_i^0\right)\\
		& \stackrel{(c)} = -\lambda \sum_{t=0}^{t}\Delta \mathcal{g}^t
	\end{align*}
	where $(a)$ follows from the initialization $\boldsymbol{\omega}^{0} = \boldsymbol{\omega}^{-1}$, $(b)$ follows from the \textbf{Assumption 4}, and $(c)$ follows from \textbf{Assumption 3}.
	
	Substituting the above equation into \eqref{eq15}, we have
	\begin{equation}
		\begin{aligned}
			\label{eq16}
			\mathbb{E}\left[\left\|\bar{\omega}^{t+1}-\bar{\omega}^t\right\|^2\right] &= \frac{\lambda^2}{N^2}\left\|\sum_{i=1}^{N}\sum_{t=0}^{t}\Delta \mathcal{g}^t\right\|^2 \\
			&\le \frac{\lambda^2}{N}\sum_{i=1}^{N}\underbrace{\left\|\sum_{t=0}^{t}\Delta \mathcal{g}^t\right\|^2}_{:\rm{T}_0}\\
			&\le \mathbb{E}\left[\frac{1}{N}\sum_{i=1}^{N}\left\|\omega_i^{t+1}-\omega_i^t\right\|^2\right]\\
			&\le \theta^2
		\end{aligned}
	\end{equation}
	So, we can bound $\rm{T}_0$ following \eqref{eq16},
	\begin{align}
		\label{T0}
		T0\le \frac{\theta^2}{\lambda^2}
	\end{align}
	
	Given the L-smooth assumption, the following inequality holds
	\begin{equation}
		\label{eq17}
		\begin{aligned}
			\mathbb{E}\left[ {f({{\bar \omega}^{t+1}})} \right] \le &\mathbb{E}\left[ {f({{\bar \omega}^{t }})} \right] + \mathbb{E}\underbrace{\left[ {\left\langle {\nabla f({{\bar \omega}^{t}}),{{\bar \omega}^{t+1}} - {{\bar \omega}^{t}}} \right\rangle } \right]}_{:\rm{T}_1} \\
			& + \frac{L}{2}\mathbb{E}\left[ {{{\left\| {{{\bar \omega}^{t+1}} - {{\bar \omega}^{t}}} \right\|}^2}} \right]
		\end{aligned}
	\end{equation}
	
	Now, let's look at $\rm{T}_1$,
	\begin{equation}
		\label{eq18}
		\begin{aligned}
			\rm{T}_1 &= \langle \nabla f(\bar{\omega}^{t}),\frac{1}{N}\sum_{i=1}^{N}\left(\omega_i^{t+1}-\omega_i^{t}\right) \rangle\\
			&=\langle \nabla f(\bar{\omega}^{t}), \frac{1}{N}\sum_{i=1}^{N}\left(-\lambda\sum_{t=0}^{t}\Delta \mathcal{g}^t\right)\rangle\\
			&=-\lambda \langle\nabla f(\bar{\omega}^{t}), \frac{1}{N}\sum_{i=1}^{N}\left(\sum_{t=0}^{t}\Delta \mathcal{g}^t\right)\rangle\\
			&\stackrel{(d)}= -\frac{\lambda}{2}\left[\left\|\nabla f(\bar{\omega}^{t})\right\|^2 + \left\|\frac{1}{N}\sum_{i=1}^{N}\left(\sum_{t=0}^{t}\Delta \mathcal{g}^t\right)\right\|^2\right] \\
			&+\frac{\lambda}{2}\left[\left\|\nabla f(\bar{\omega}^{t})-\frac{1}{N}\sum_{i=1}^{N}\left(\sum_{t=0}^{t}\Delta \mathcal{g}^t\right)\right\|^2\right]
		\end{aligned}
	\end{equation}
	where $(d)$ follows from the fundamental equation $\langle \mathbf{A},\mathbf{B}\rangle = \frac{1}{2}\left[\left\|\mathbf{A}\right\|^2 + \left\|\mathbf{B}\right\|^2 - \left\|\mathbf{A} - \mathbf{B}\right\|^2\right]$ for any vector $\mathbf{A},\mathbf{B}$.
	
	Substituting \eqref{eq16} and \eqref{eq18} into \eqref{eq17}, we have
	\begin{equation}
		\label{eq19}
		\begin{aligned}
			\mathbb{E}\left[f(\bar{\omega}^{t+1})\right] &\le \mathbb{E}\left[f(\bar{\omega}^t)\right]-\frac{\lambda}{2}\left\|\nabla f(\bar{\omega}^t)\right\|^2 \\
			&-\frac{\lambda}{2} \left\|\frac{1}{N}\sum_{i=1}^{N}\left(\sum_{t=0}^{t}\Delta \mathcal{g}^t\right)\right\|^2\\
			&+\frac{\lambda}{2} \underbrace{\left[\left\|\nabla f(\bar{\omega}^{t})-\frac{1}{N}\sum_{i=1}^{N}\left(\sum_{t=0}^{t}\Delta \mathcal{g}^t\right)\right\|^2\right]}_{:\rm{T}_2}\\
			&+\frac{L\lambda^2}{2N^2}\left\|\sum_{i=1}^{N}\sum_{t=0}^{t}\Delta \mathcal{g}^t \right\|^2
		\end{aligned}
	\end{equation}
	
	Now let's bound the $\rm{T}_2$,
	\begin{equation}
		\label{eq20}
		\begin{aligned}
			\rm{T}_2 &= \left\|\frac{1}{N}\sum_{i=1}^{N}\nabla f_i(\bar{\omega}^t) - \frac{1}{N}\sum_{i=1}^{N}\sum_{t=0}^{t}\Delta \mathcal{g}^t\right\|^2 \\
			&=\frac{1}{N^2}\left\| \sum_{i=1}^{N}\left(\nabla f_i(\bar{\omega}^t) - \sum_{t=0}^{t}\Delta \mathcal{g}^t\right)\right\|^2 \\
			&\stackrel{(e)} \le \frac{1}{N}\sum_{i=1}^{N}\left\|\nabla f_i(\bar{\omega}^t) - \sum_{t=0}^{t}\Delta \mathcal{g}^t\right\|^2 \\
			& \stackrel{(f)} \le \frac{1}{N}\sum_{i=1}^{N}\left(\left\|\nabla f_i(\bar{\omega}^t)\right\|^2 + \left\|\sum_{t=0}^{t}\Delta \mathcal{g}^t\right\|^2\right)\\
			&\stackrel{(g)} \le \frac{1}{N}\sum_{i=1}^{N}\left(G^2 + \frac{\theta^2}{\lambda^2}\right)\\
			&= G^2 + \frac{\theta^2}{\lambda^2}
		\end{aligned}
	\end{equation}
	where ($e$) follows by the inequality $\left\|\sum_{i=1}^{N}\mathbf{z}_i\right\|^2 \le N \sum_{i=1}^{N}\left\|\mathbf{z}_i\right\|^2$, ($f$) follows from the inequality $\left\|\mathbf{A}+\mathbf{B}\right\|^2 \le \left\|\mathbf{A}\right\|^2 + \left\|\mathbf{B}\right\|^2$ for any vector $\mathbf{A},\mathbf{B}$, and ($g$) follows from \textbf{Assumption 2} and \eqref{T0}.
	
	Substituting \eqref{eq20} into \eqref{eq19}, we have
	\begin{small}
		\begin{equation}
			\label{eq21}
			\begin{aligned}
				\mathbb{E}\left[f(\bar{\omega}^{t+1})\right] &\le \mathbb{E}\left[f(\bar{\omega}^t)\right] -\frac{\lambda}{2}\left\|\nabla f(\bar{\omega}^t)\right\|^2 \\
				&-\underbrace{\frac{\lambda}{2} \left\|\frac{1}{N}\sum_{i=1}^{N}\left(\sum_{t=0}^{t}\Delta \mathcal{g}^t\right)\right\|^2}_{:\rm{T}_3}\\
				& + \frac{\lambda}{2}\left(G^2+\frac{\theta^2}{\lambda^2}\right) \\
				& + \frac{L\lambda^2}{2N^2}\left\|\sum_{i=1}^{N}\sum_{t=0}^{t}\Delta \mathcal{g}^t \right\|^2\\
				&\stackrel{(h)}\le \mathbb{E}\left[f(\bar{\omega}^t)\right] -\frac{\lambda}{2}\left\|\nabla f(\bar{\omega}^t)\right\|^2  + \frac{\lambda}{2}\left(G^2+\frac{\theta^2}{\lambda^2}\right) \\
				& + \frac{L\lambda^2}{2N^2}\left\|\sum_{i=1}^{N}\sum_{t=0}^{t}\Delta \mathcal{g}^t \right\|^2\\
				& \stackrel{(i)} \le \mathbb{E}\left[f(\bar{\omega}^t)\right] -\frac{\lambda}{2}\left\|\nabla f(\bar{\omega}^t)\right\|^2  + \frac{\lambda}{2}\left(G^2+\frac{\theta^2}{\lambda^2}\right) \\
				& + \frac{L\lambda^2}{2N}\sum_{i=1}^{N}\left\|\sum_{t=0}^{t}\Delta \mathcal{g}^t\right\|^2 \\
				& \stackrel{(j)} \le \mathbb{E}\left[f(\bar{\omega}^t)\right] -\frac{\lambda}{2}\left\|\nabla f(\bar{\omega}^t)\right\|^2  + \frac{\lambda}{2}\left(G^2+\frac{\theta^2}{\lambda^2}\right) \\
				& + \frac{L\theta^2}{2}\\
				&=\mathbb{E}\left[f(\bar{\omega}^t)\right] -\frac{\lambda}{2}\left\|\nabla f(\bar{\omega}^t)\right\|^2  +\frac{\lambda}{2}\left(G^2+\frac{\theta^2}{\lambda^2}+\frac{L\theta^2}{\lambda}\right)
			\end{aligned}
		\end{equation}
	\end{small}
	where $(h)$ follows because $\rm{T}_3 \ge 0$, $(i)$ and $(j)$ follow from the inequality $\left\|\sum_{i=1}^{N}\mathbf{z}_i\right\|^2 \le N \sum_{i=1}^{N}\left\|\mathbf{z}_i\right\|^2$ aforementioned and \eqref{eq16}, respectively.
	
	Rearrange the \eqref{eq21}, we have
	\begin{equation}
		\label{eq22}
		\left\|\nabla f(\bar{\omega}^t)\right\|^2 \le \frac{2}{\lambda}\mathbb{E}\left[f(\bar{\omega}^t) - f(\bar{\omega}^{t+1})\right] + G^2 + \frac{\theta^2}{\lambda^2} + \frac{L\theta^2}{\lambda}
	\end{equation}
	
	For \eqref{eq22}, we sum it over $t \in\{0,1,2, \cdots, T-1\}$ first and then divide both sides by $\mathit{T}$,
	\begin{small}
		\begin{equation}
			\begin{aligned}
				\frac{1}{T}\sum_{t=0}^{T-1}\left\|\nabla f(\bar{\omega}^t)\right\|^2
				&\le \frac{2}{\lambda T}\mathbb{E}\left(f(\bar{\omega}^0) - f(\bar{\omega}^T)\right) + G^2 + \frac{\theta^2}{\lambda^2} + \frac{L\theta^2}{\lambda} \\
				&\le \frac{2}{\lambda T}\left(f(\bar{\omega}^0) - f^*\right) + G^2 + \frac{\theta^2}{\lambda^2} + \frac{L\theta^2}{\lambda}
			\end{aligned}
		\end{equation}
	\end{small}	
	where the $f^*$ is the optimum of loss function $f$, and this completes the proof.
\end{IEEEproof}

\section*{Acknowledgment}

This work was partially supported by Key Program of Natural Science Foundation of China under Grant(61631018), Huawei Technology Innovative Research.

\ifCLASSOPTIONcaptionsoff
  \newpage
\fi


\bibliographystyle{IEEEtran}

\begin{IEEEbiography}[{\includegraphics[width=1in,height=1.25in,clip,keepaspectratio]{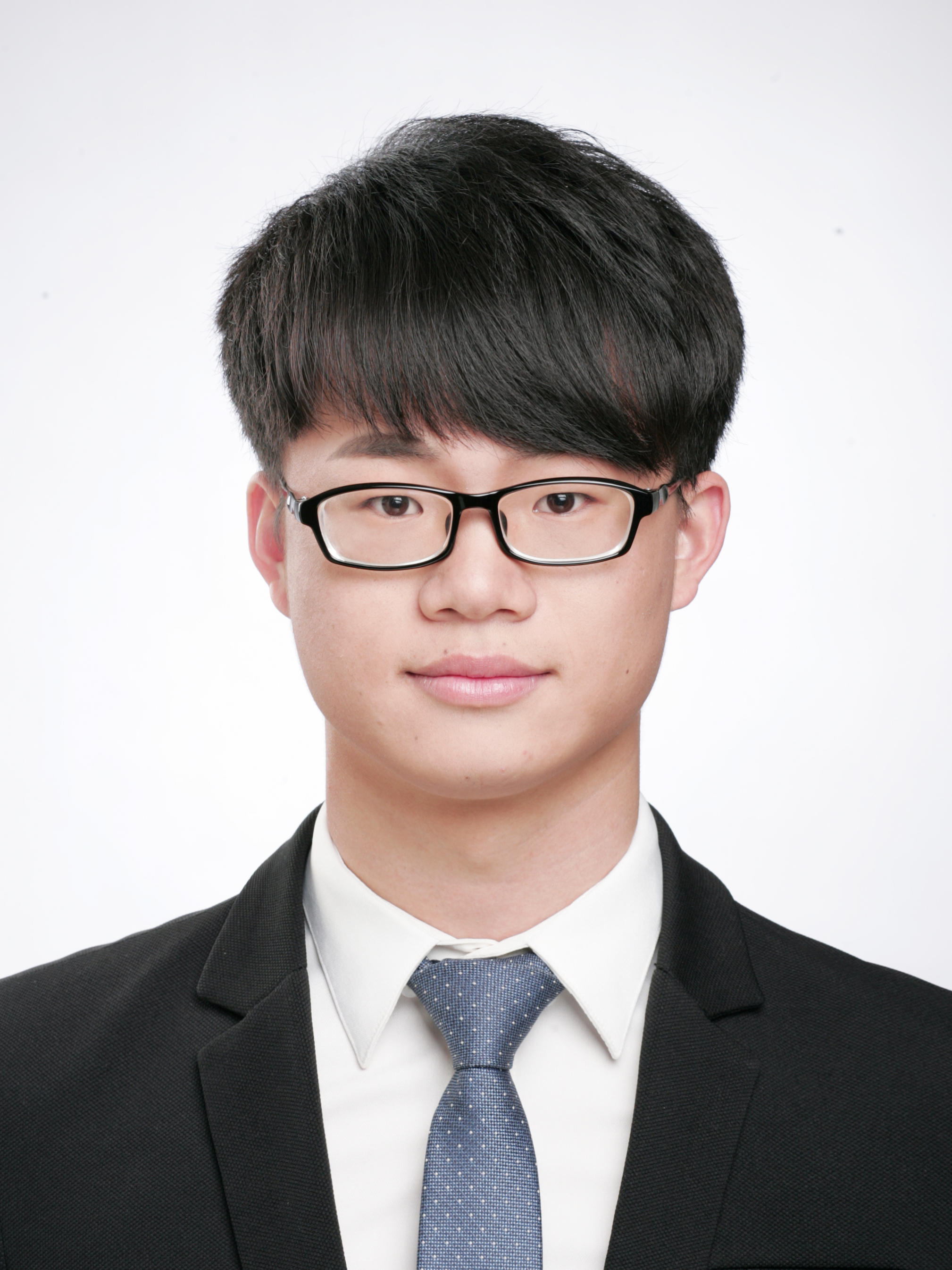}}]{Zhikun Chen}
	received the B.S. degree in Electronic Information Science and Technology from the Shandong University (SDU), Shandong, China, in 2018. He is currently pursuing his Ph.D. degree in the College of Information Science and Technology, University of Science and Technology of China (USTC), Hefei, China. His research interests are mainly in artificial intelligence, federated learning and wireless big data.
\end{IEEEbiography}

\begin{IEEEbiography}[{\includegraphics[width=1in,height=1.25in,clip,keepaspectratio]{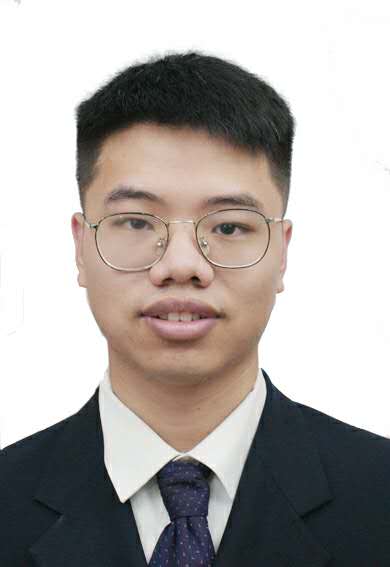}}]{Daofeng Li}
	received the B.S. degree in Electronic Engineering and Information Science from the University of Science and Technology of China (USTC), Hefei, China, in 2018. He is currently pursuing the M.E. degree in the College of Information Science and Technology, University of Science and Technology of China (USTC), Hefei, China. His research interests are mainly in artificial intelligence and wireless communication.
\end{IEEEbiography}

\begin{IEEEbiography}[{\includegraphics[width=1in,height=1.25in,clip,keepaspectratio]{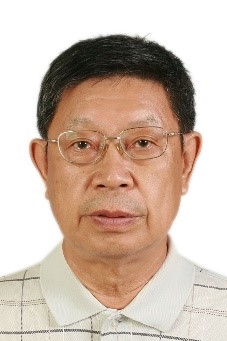}}]{Jinkang Zhu} (M'92-LM'18) received his B.S. degree in Electrical Engineering from Sichuan University, China, in 1966. He joined University of Science and Technology of China (USTC) in 1966, and has been a professor of USTC since 1992. He has been committed to research on wireless mobile communications and networks, signal processing for communications and the future wireless technologies.
\end{IEEEbiography}

\begin{IEEEbiography}[{\includegraphics[width=1in,height=1.25in,clip,keepaspectratio]{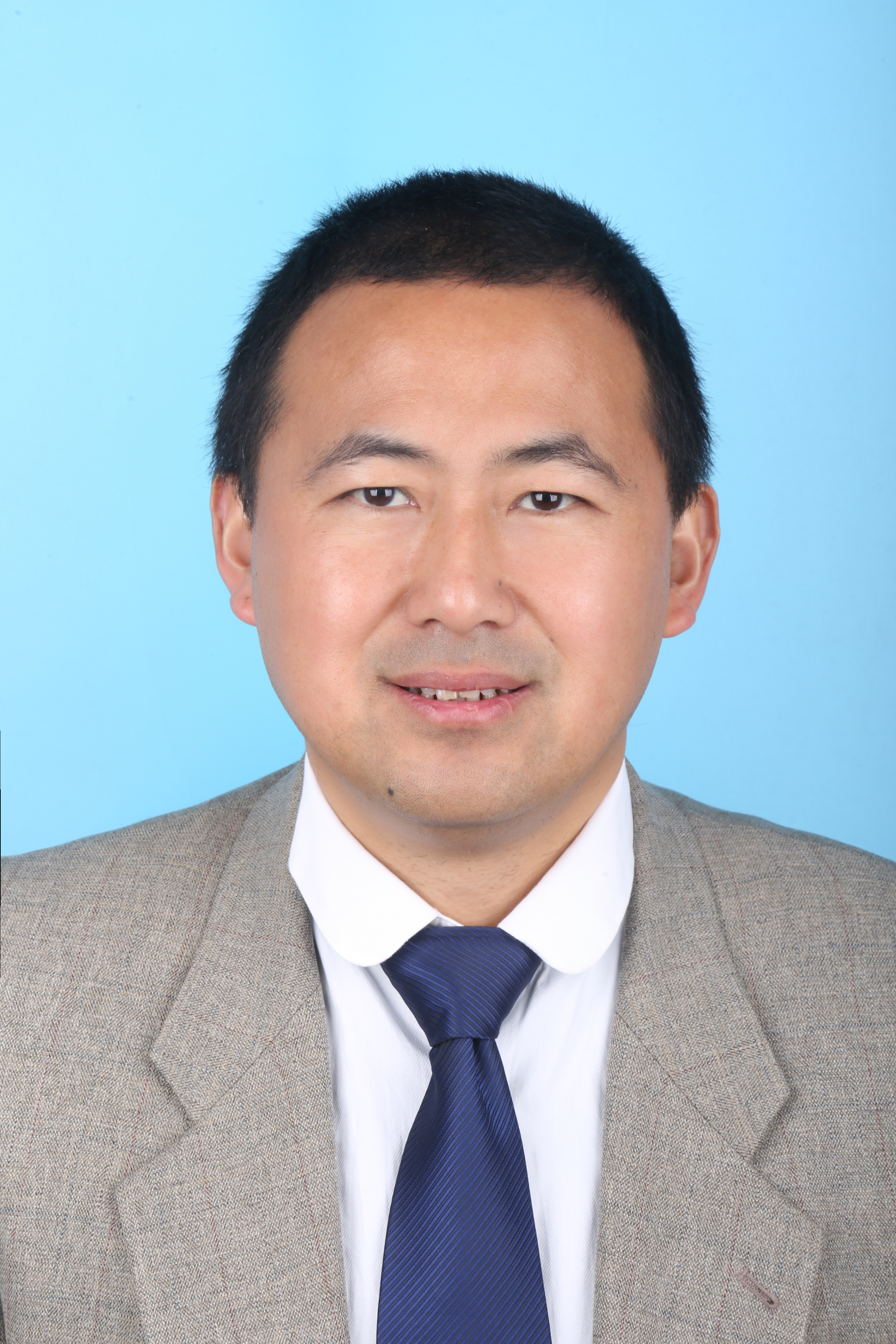}}]{Sihai Zhang}
	(M'09-SM'18) received the Ph.D. degree from the Department of Computer Science and Technology, University of Science and Technology of China (USTC), Hefei, China, in 2006. He has been with the PCNSS Laboratory, Department of Electronic Engineering and Information Science, USTC, since 2009. His research interests include wireless networks, big data analysis and intelligent algorithms.
	
	He is currently an associate professor of Electronic Engineering with the Department of Electronic Engineering and Information Science, USTC. He has authored or co-authored over 60 technical papers, such as the IEEE TETC, the IEEE TVT, MONET, and WPC. He initiated the research field of wireless big data in 2014. He has participated in projects, including the National Science Foundation of China for Machine Type Communications, Key Program of the National Natural Science Foundation of China for Wireless Big Data. He has served over 15 international conferences as a member of organizing committee, TPC member or a reviewer, such as publication chair for WCSP 2014. In 2016, he has co-chaired special sessions on Wireless Big Data in WCSP 2016, Machine Type Communications in WPMC2016 and guest editor of special issue on Wireless Big Data for JCIN.
\end{IEEEbiography}
\balance
\end{document}